\makeatletter\def\input@path{{arxiv/}}\makeatother
\makeatletter\providecommand\@LN[2]{}\makeatother
\documentclass[11pt, a4paper, logo, copyright, nonumbering]{mll_style}

\usepackage{amsmath,amsfonts,bm}

\def\eqref#1{equation~\ref{#1}}

\def\1{\bm{1}}

\DeclareMathAlphabet{\mathsfit}{\encodingdefault}{\sfdefault}{m}{sl}
\SetMathAlphabet{\mathsfit}{bold}{\encodingdefault}{\sfdefault}{bx}{n}

\PassOptionsToPackage{numbers,square,sort&compress}{natbib}
\usepackage{natbib}
\usepackage[most]{tcolorbox}
\usepackage{nicefrac}
\usepackage{multirow}
\usepackage{subcaption}
\usepackage{wrapfig}
\usepackage{placeins}
\renewcommand{\topfraction}{0.85}
\renewcommand{\textfraction}{0.15}
\usepackage{enumitem}
\usepackage{algorithm}
\usepackage{algpseudocode}
\usepackage{booktabs}
\usepackage{tabularx}
\usepackage{makecell}
\usepackage[capitalise,noabbrev]{cleveref}

\usepackage[toc,page,header]{appendix}
\hypersetup{linktoc=all}

\makeatletter
\newif\ifarcappendixentries
\newcommand{\arcappendixcontents}{%
  \begingroup
  \small
  \setlength{\parskip}{0pt}
  \hypersetup{linkcolor=citecolor}
  \renewcommand{\contentsname}{\textcolor{citecolor}{Contents}}%
  \arcappendixentriesfalse
  \def\arcfirstappendix{appendix.A}%
  \let\arccontentsline\contentsline
  \renewcommand*\contentsline[4]{%
    \def\arcentryanchor{##4}%
    \ifx\arcentryanchor\arcfirstappendix
      \arcappendixentriestrue
    \fi
    \ifarcappendixentries
      \arccontentsline{##1}{##2}{##3}{page.##3}%
    \fi}%
  \let\l@part\@gobbletwo
  \let\l@xpart\@gobbletwo
  \renewcommand*\l@section[2]{%
    \addpenalty{-\@highpenalty}%
    \vskip 1.5em
    \@dottedtocline{1}{0pt}{1.8em}{\bfseries ##1}{\bfseries ##2}}%
  \let\arcsubsectionentry\l@subsection
  \renewcommand*\l@subsection[2]{%
    \vskip 0.3em
    \arcsubsectionentry{##1}{##2}}%
  \tableofcontents
  \endgroup
}
\makeatother

\usepackage{fontawesome5}
\usepackage{simpleicons}
\definecolor{findingaccent}{HTML}{5E77D6}
\definecolor{findingback}{HTML}{F1F5FE}
\definecolor{findingframe}{HTML}{CFD9F3}
\newtcolorbox{researchquestion}{enhanced,
  width=0.94\linewidth, center,
  colback=abstractbgcolor, colframe=abstractbgcolor, boxrule=0pt, arc=5pt,
  left=9pt, right=9pt, top=7pt, bottom=7pt,
  before skip=9pt, after skip=10pt, fontupper=\normalsize}
\newtcolorbox{findingbox}[1]{enhanced, colback=findingback, colframe=findingframe, boxrule=0.4pt, arc=5pt,
  sharp corners=west, borderline west={3.5pt}{0pt}{findingaccent},
  title={\faLightbulb\enspace Key Takeaways: #1}, fonttitle=\bfseries\small, coltitle=white,
  attach boxed title to top left={xshift=10pt, yshift*=-\tcboxedtitleheight/2},
  boxed title style={colback=findingaccent, colframe=findingaccent, boxrule=0pt, arc=3pt, left=5pt, right=6pt, top=2pt, bottom=2pt},
  left=8pt, right=8pt, top=5pt, bottom=5pt, before skip=10pt, after skip=10pt, fontupper=\small,
  before upper={\setlist[itemize]{leftmargin=1.1em, topsep=0pt, partopsep=0pt, itemsep=2pt, parsep=0pt}}}

\usepackage{pifont}
\usepackage{xspace}
\usepackage{tikz}
\usetikzlibrary{positioning,calc,arrows.meta,fit,backgrounds}
\newcommand{\name}{RoboARC\xspace}
\definecolor{markgreen}{HTML}{2E9A5A}\definecolor{markred}{HTML}{D2462A}
\newcommand{\cmark}{\textcolor{markgreen}{\ding{51}}}
\newcommand{\xmark}{\textcolor{markred}{\ding{55}}}
\DeclareRobustCommand{\pmark}{\tikz[baseline=-0.55ex]{\fill[black!70] (0,0) -- (90:0.62ex) arc (90:270:0.62ex) -- cycle;
  \draw[black!70,line width=0.35pt] (0,0) circle (0.62ex);}}
\newcommand{\na}{\textcolor{black!35}{--}}
\newcommand{\priv}{\textit{Privileged}\xspace}
\newcommand{\std}{\textit{Standard}\xspace}
\definecolor{ink}{HTML}{1F2329}
\definecolor{ink2}{HTML}{4B525C}
\definecolor{ink3}{HTML}{7C848E}
\definecolor{privc}{HTML}{8F98A3}
\definecolor{privink}{HTML}{6B737D}
\definecolor{stdc}{HTML}{5A9EA2}
\definecolor{stdink}{HTML}{3F8589}
\definecolor{panel}{HTML}{F3F4F6}
\definecolor{panel2}{HTML}{EAEEF3}
\definecolor{cell0}{HTML}{E4E9EF}
\definecolor{rule}{HTML}{DDE1E5}
\usepackage{colortbl}
\usepackage{float}
\newsavebox{\fittabbox}
\newenvironment{fittab}{\begin{lrbox}{\fittabbox}}{\end{lrbox}%
  \ifdim\wd\fittabbox>0.9\linewidth \resizebox{\linewidth}{!}{\usebox{\fittabbox}}\else\usebox{\fittabbox}\fi}
\definecolor{markorange}{HTML}{D2683F}\definecolor{markteal}{HTML}{2F93A6}
\newcommand{\roboarcmark}[1]{\tikz[x={(#1*0.05,0pt)},y={(0pt,#1*0.05)},baseline=(current bounding box.south)]{%
  \useasboundingbox (2,6) rectangle (30,26);
  \draw[markorange,line width=#1*0.16,line cap=round] (6,22) -- (18,22) arc[start angle=90,end angle=-90,radius=5] -- (11,12);
  \draw[markteal,line width=#1*0.16,line cap=round] (26,10) -- (14,10) arc[start angle=270,end angle=90,radius=5] -- (21,20);}}

\fancypagestyle{firststyle}{
    \fancyhead[L]{%
        \raisebox{-4pt}{\roboarcmark{30pt}}%
        \hspace{6pt}{\sffamily\bfseries\LARGE RoboARC}%
    }
    \fancyhead[C]{}
    \fancyhead[R]{%
        \includegraphics[height=30pt]{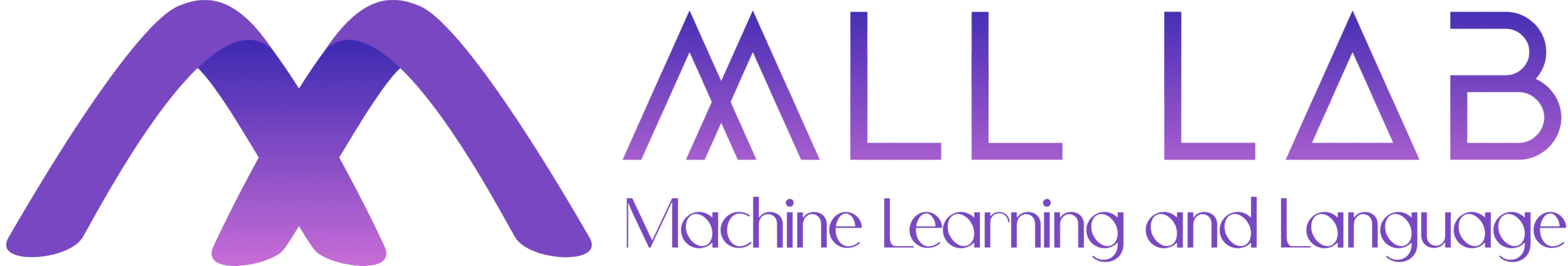}%
    }
}

\renewcommand{\absfont}{\fontsize{10.3}{12.0}\selectfont}

\title{\centering RoboARC: Physical Puzzles for Robotic Manipulation under Spatial Constraints}
\date{}

\author{
{\fontsize{11}{15}\selectfont\parbox{\linewidth}{\centering
Qineng Wang*$^{1}$, Jianwen Lyu*$^{1}$, Pingyue Zhang*$^{1}$, Reuben Tan$^{2\dag}$, Xiaoyu Huang$^{3\dag}$,\newline
Tonghe Zhang$^{4\dag}$, Kangrui Wang$^{1}$, Yelong Shen$^{2}$, Jianfeng Gao$^{2}$,\newline
Guanya Shi$^{4}$, Ruohan Zhang$^{1}$, Manling Li$^{1}$}}
\\
\small $^1$Northwestern University~~~$^2$Microsoft~~~$^3$UC Berkeley~~~$^4$Carnegie Mellon University
\\
\footnotesize \emph{* Equal contribution; $\dag$ Core contributor.}
\\
{
    \footnotesize
    \href{https://roboarc.embodied-first-exam.ai}{\faGlobe~Website}
    \quad
    \href{https://github.com/mll-lab-nu/RoboARC}{\faGithub~Code}
    \quad
    \href{https://huggingface.co/datasets/RoboARC/RoboARC-runs}{\faDatabase~Agent runs}
}
\vspace{-10pt}
}

\begin{abstract}
Many hard manipulation tasks come down to a spatial constraint that only a few motions satisfy, such as passing a part
through a gap barely wider than itself or separating two interlocked parts. These motions depend on the exact geometry of
each scene, so demonstrations and skills recorded elsewhere rarely transfer. We
therefore ask: when manipulation hinges on spatial constraints, can robots reason out the right motions? We introduce \textbf{\name}, a benchmark of $90$ tasks in six constraint types over rigid and soft materials, on three robots in MuJoCo. Every task is a frozen scene with a physics-based grader and a verified reference solution that replays
bit-identically. Each task comes in a \emph{Privileged} regime, with the full simulator state and resets, and a
\emph{Standard} regime with real-sensor inputs (camera images and proprioception) and a single attempt without resets.
We evaluate seven coding agents and four VLA
policies, none trained on these tasks. The strongest agent, GPT-6 Astra in Codex, solves $86\,\%$ of the tasks in the \priv regime and $43\,\%$ in the \std regime,
and the four VLA policies, run zero-shot, solve none of their $18$ one-arm puzzles. On $15$ puzzles that GPT-6 Astra solved in the \priv regime and failed in the \std regime, object poses add $4$ points and part geometry $13$, but both together add $36$, from $24\,\%$ to $60\,\%$. Only $7\,\%$ of the coding agents' failed runs pass more than half of a task, and $53\,\%$ of the failed runs in the \std regime are stuck on a single step. Frontier coding agents already reason out
the right motions for many of these tasks once they know where the parts are and what shape they have. From their own sensors,
in a single attempt, the strongest solves fewer than half of the tasks.

\end{abstract}

\begin{document}
\maketitle
\noindent\begin{minipage}{\linewidth}
  \centering
  \includegraphics[width=\linewidth]{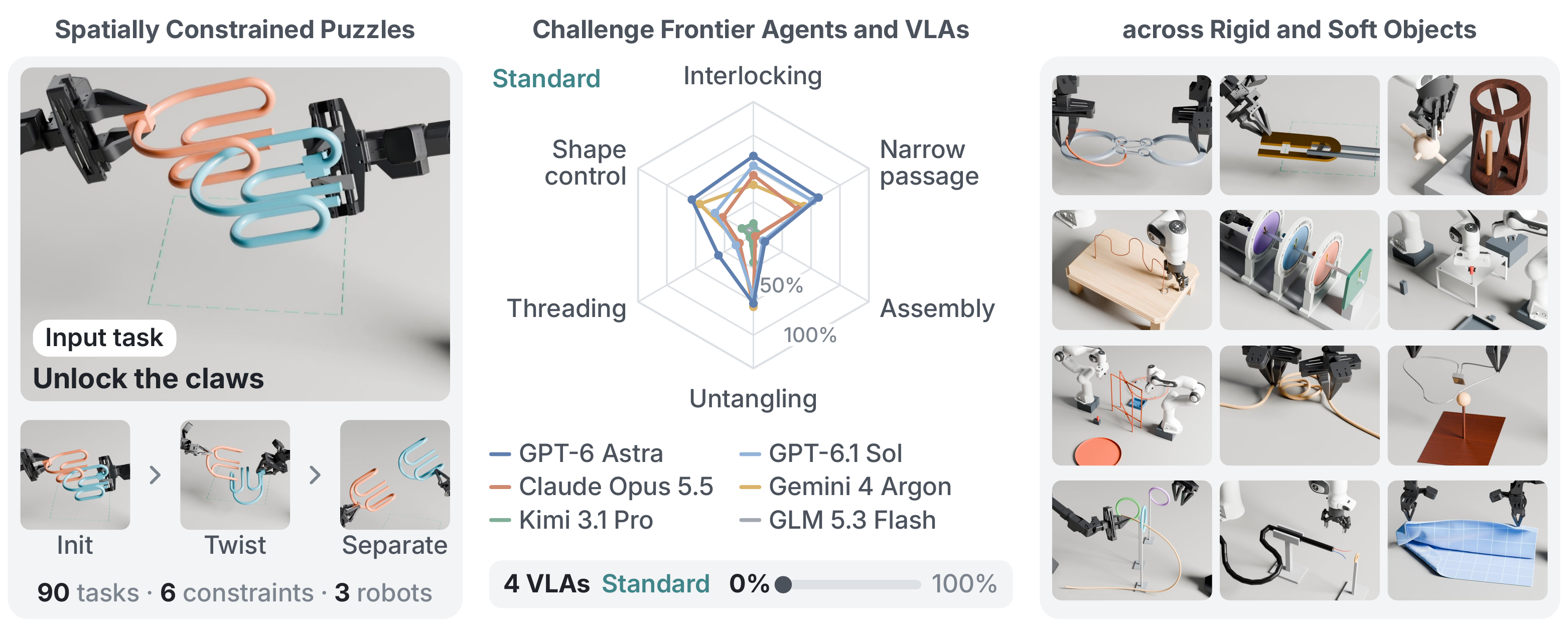}
  \captionof{figure}{\textbf{\name} asks whether robots can reason out the right motions when manipulation hinges on
  spatial constraints. Its $90$ tasks span six constraint types over rigid and soft materials (left and right). In the
  \std regime the strongest coding agent solves $43\%$ of the tasks, and four zero-shot VLA policies solve none of their $18$ one-arm puzzles (middle).}
  \label{fig:teaser}
\end{minipage}
\clearpage

\section{Introduction}
\label{sec:intro}

\begin{figure}[t]
  \centering
  \includegraphics[width=\textwidth]{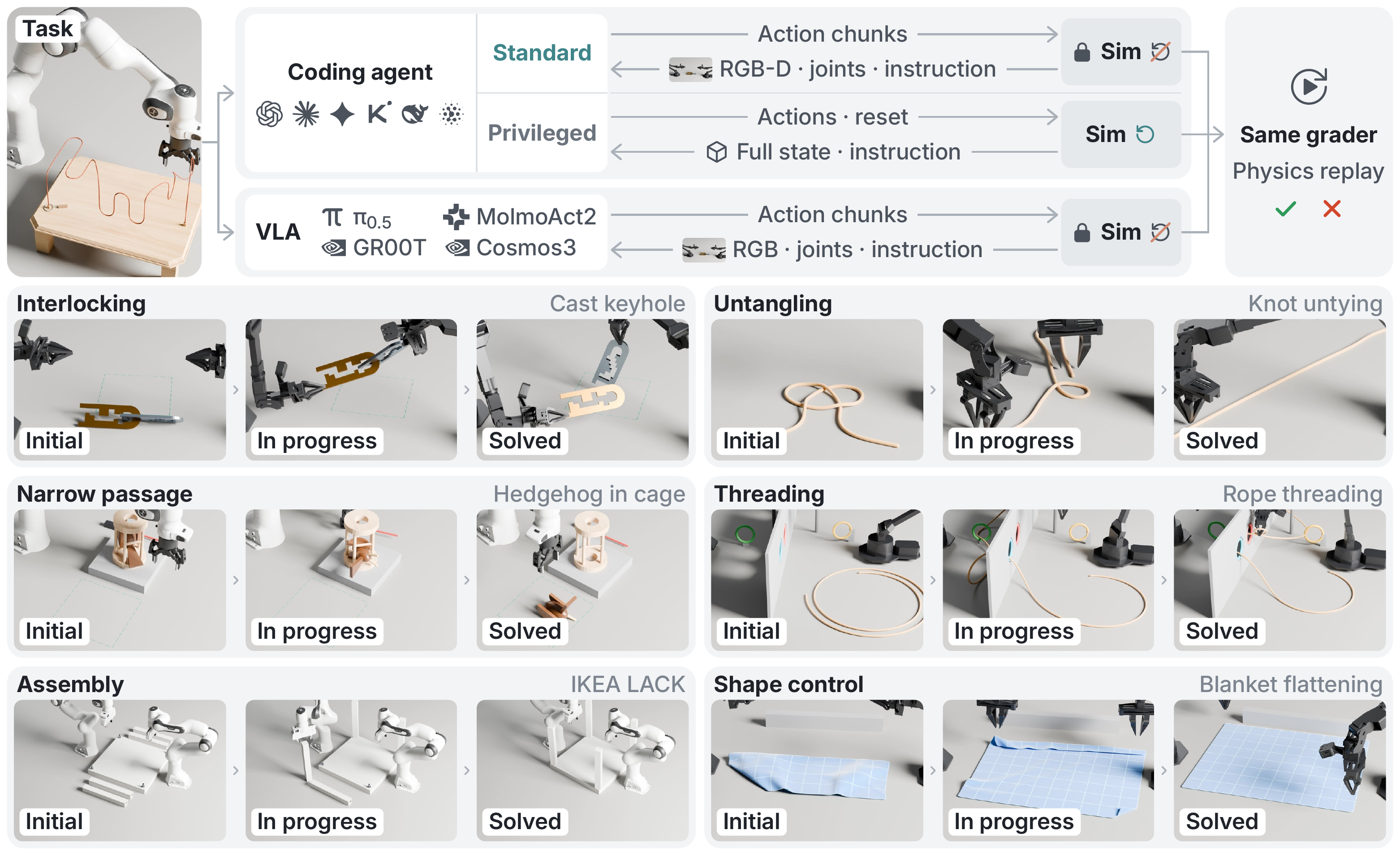}
  \caption{\textbf{Overview of \name.} \emph{Top:} the two regimes share the scene, physics and grader. In the \std regime a coding agent sends action chunks and receives RGB-D images, joint readings
and the instruction, without resets. In the \priv regime it also reads the full simulator state and
may reset. VLA policies run under the \std rules with RGB images. One grader replays the physics of every run. \emph{Bottom:} one task per constraint type, from the initial to the solved state.}
  \label{fig:gallery}\label{fig:modes}
\end{figure}

In some manipulation tasks only a few motions succeed. Untangling a cable, threading a rope through a fixture or sliding
a board into a frame~\cite{grannen2020untangling,seita2021deformableravens,lee2021ikea} succeeds only along the few
motions that a \emph{spatial constraint} leaves open: a gap barely wider than the part, two parts that interlock, or a
rope that must not pass through itself. Which motions remain is set by the exact geometry of the scene and changes from
scene to scene. Policies learned from demonstrations, as in vision-language-action (VLA)
models~\cite{zitkovich2023rt2,kim2024openvla,black2024pi0}, and programs composed from known
skills~\cite{liang2023codeaspolicies,huang2023voxposer} succeed when the right motion resembles one seen before, such as
a grasp, a push or a placement. Coding agents built on models that reason well in symbolic domains, from abstract
puzzles~\cite{chollet2019measure} to computer use~\cite{xie2024osworld,terminalbench}, already control robots through
code~\cite{fu2026capx} and could work out a motion for the scene at hand. This raises the question:
\begin{researchquestion}
\textbf{When manipulation hinges on spatial constraints, can robots reason out the right motions?}
\end{researchquestion}

Existing benchmarks do not address this question. Manipulation benchmarks draw their difficulty from the number of
steps and the variety of objects~\cite{james2020rlbench,mees2022calvin,liu2023libero,nasiriany2024robocasa,li2024behavior1k,tao2024maniskill3},
rarely from the geometry a motion must clear (\cref{tab:compare}). Reasoning benchmarks pose puzzles in text, images or
abstract environments~\cite{chollet2019measure,mindtopo,chain2026}, where no motion has to satisfy a contact. Motion
planning has long used physical puzzles such as the alpha puzzle as tests for planners that are given the exact geometry~\cite{amato1998alphapuzzle,zhang2020disentangle}.
\name poses them to complete systems and grades the motion the robot executes.

To answer this question, we introduce \textbf{\name}, a benchmark of physical puzzles under spatial constraints
(\cref{fig:teaser}), built on three principles.

\begin{table}[t]
\centering
\caption{\textbf{Comparison with related benchmarks.} A check mark means that at least some tasks have the
property (\pmark\ under Topological: a single task). Physics: success is computed from the simulated state (\pmark: partly by a VLM rubric). Verified solution:
every task ships a solution that passes its own check. Exact replay: the simulation is stated to be deterministic, so the same actions reproduce the same states. State hidden: the coding agent is never given the simulator's state (object poses,
shapes). \Cref{tab:compare_full} extends the comparison to $24$ benchmarks, with all definitions in \cref{sec:suppl_compare}.}
\label{tab:compare}
\scriptsize
\setlength{\tabcolsep}{1.6pt}
\setlength{\tabcolsep}{4pt}
\begin{fittab}\begin{tabular}{@{}l r *{9}{c}@{}}
\toprule
 &  & \multicolumn{3}{c}{Tasks} & \multicolumn{3}{c}{Grading} & \multicolumn{3}{c}{Interfaces} \\
\cmidrule(lr){3-5} \cmidrule(lr){6-8} \cmidrule(lr){9-11}
Benchmark & \#Tasks & Long-horizon & Deformable & Topological & Physics & Verified solution & Exact replay & Coding agent & State hidden & Learned policy \\
\midrule
\multicolumn{11}{@{}l}{\rule{0pt}{2.1ex}\textsc{\scriptsize robot manipulation}} \\
BEHAVIOR-1K~\cite{li2024behavior1k} & $1{,}000$ & \cmark & \cmark & \xmark & \cmark & \xmark & \xmark & \xmark & \na & \cmark \\
VLABench~\cite{zhang2024vlabench} & $100$ & \cmark & \xmark & \xmark & \cmark & \xmark & \xmark & \cmark & \xmark & \cmark \\
KinDER~\cite{huang2026kinder} & $25$ & \cmark & \xmark & \xmark & \cmark & \xmark & \xmark & \xmark & \na & \cmark \\
CaP-X~\cite{fu2026capx} & $187$ & \cmark & \xmark & \xmark & \cmark & \xmark & \xmark & \cmark & \cmark & \cmark \\
BiGym 2.0~\cite{zhang2026bigym2} & $20$ & \cmark & \xmark & \xmark & \cmark & \cmark & \cmark & \cmark & \cmark & \cmark \\
EmbodiedSWE~\cite{shen2026embodiedswe} & $28$ & \cmark & \cmark & \pmark & \pmark & \xmark & \xmark & \cmark & \xmark & \cmark \\
\addlinespace[2pt]
\multicolumn{11}{@{}l}{\rule{0pt}{2.1ex}\textsc{\scriptsize computer-use agents}} \\
SWE-bench~\cite{jimenez2024swebench} & $2{,}294$ & \xmark & \na & \na & \na & \cmark & \na & \cmark & \na & \na \\
Terminal-Bench~\cite{terminalbench} & $89$ & \cmark & \na & \na & \na & \cmark & \na & \cmark & \na & \na \\
OSWorld~\cite{xie2024osworld} & $369$ & \cmark & \na & \na & \na & \xmark & \na & \cmark & \na & \na \\
\midrule
\textbf{\name (ours)} & $90$ & \cmark & \cmark & \cmark & \cmark & \cmark & \cmark & \cmark & \cmark & \cmark \\
\bottomrule
\end{tabular}\end{fittab}
\end{table}

\noindent\textbf{First, difficulty comes from spatial constraints.} Like ARC~\cite{chollet2019measure}, each task is easy to
state, but its solution is specific to its geometry, and the agent is given no demonstration of it. Unlike
ARC, the answer is a motion. \name has $90$ tasks, $78$ from $19$ puzzle families and $12$ IKEA scenes, in six constraint types: interlocking, narrow passage, assembly,
untangling, threading and shape control (\cref{fig:gallery,fig:stats}). They span
rigid and soft materials and three robots in MuJoCo~\cite{todorov2012mujoco}.

\noindent\textbf{Second, each task comes in two regimes.} A policy can fail because it cannot see the constraint, because it cannot reason through it, or because it cannot carry out the motion. Each task therefore comes in two regimes that share the scene, the physics and the grader
(\cref{fig:modes}). The \priv regime gives the simulator itself, with its full state and resets. The \std regime gives real-sensor inputs, calibrated cameras and
proprioception, and one live episode. The gap between the regimes measures what access to the simulator is worth, and \cref{sec:ladder} gives it back to GPT-6 Astra one kind at a time.

\noindent\textbf{Third, the tasks are open to any policy.} Coding agents reach
the tasks through Harbor~\cite{harbor}, and general policies through a Gymnasium
environment~\cite{towers2024gymnasium} that VLA policies join as openpi policy servers~\cite{black2024pi0}. Policies range from VLAs that map images to actions to coding agents that write planners. All are
judged by the same physics-based grader on the motion they execute.

We evaluate seven coding agents, each a complete system of harness, model and reasoning effort, and four VLA
policies, none trained on these tasks (\cref{sec:experiments}). To find which information an agent lacks in the \std regime, we give it back to
GPT-6 Astra at high effort, one kind at a time (\cref{sec:ladder}). We then replay the runs of all seven agents and read their transcripts to find where the failed runs stop, what form the failed attempts take and whether the agents plan before they act
(\cref{sec:endings}). Our main findings are as follows.
\begin{itemize}[leftmargin=1.5em,itemsep=2pt,topsep=4pt]
  \item \textbf{The strongest coding agent solves $\mathbf{43\%}$ of the tasks in the \std regime, and zero-shot VLA policies solve none (\cref{tab:main,tab:vla}).}
  GPT-6 Astra in Codex solves $86\%$ in the \priv regime and $43\%$ in the \std regime. On their $18$ one-arm puzzles, the four VLA policies often touch
  the task object but solve none.
  \item \textbf{Proprietary coding agents lose more than half of their successes in the \std regime (\cref{sec:main_results}).} The four
  proprietary agents solve $73$--$86\%$ of the tasks in the \priv regime and $29$--$43\%$ in the \std regime.
  \item \textbf{For GPT-6 Astra, object poses and part geometry are synergistic (\cref{sec:ladder}).} On $15$ puzzles that GPT-6 Astra solved in the \priv regime and
  failed in the \std regime, object poses alone add $4$ points and part geometry $13$, but both together add $36$,
  twice what the two add separately, from $24\,\%$ to $60\,\%$.
  \item \textbf{Failed runs of the coding agents make little progress and many are stuck on one step (\cref{sec:endings}).}
  Only $7\,\%$ pass more than half of a task, and $53\,\%$ (\std) and $38\,\%$ (\priv) are stuck on the step where they stop.
  A third stop at the key step, the one that passes the task's constraint.
  
\end{itemize}

\section{Related work}
\label{sec:related}

\begin{figure}[t]
  \centering
  \includegraphics[width=\textwidth]{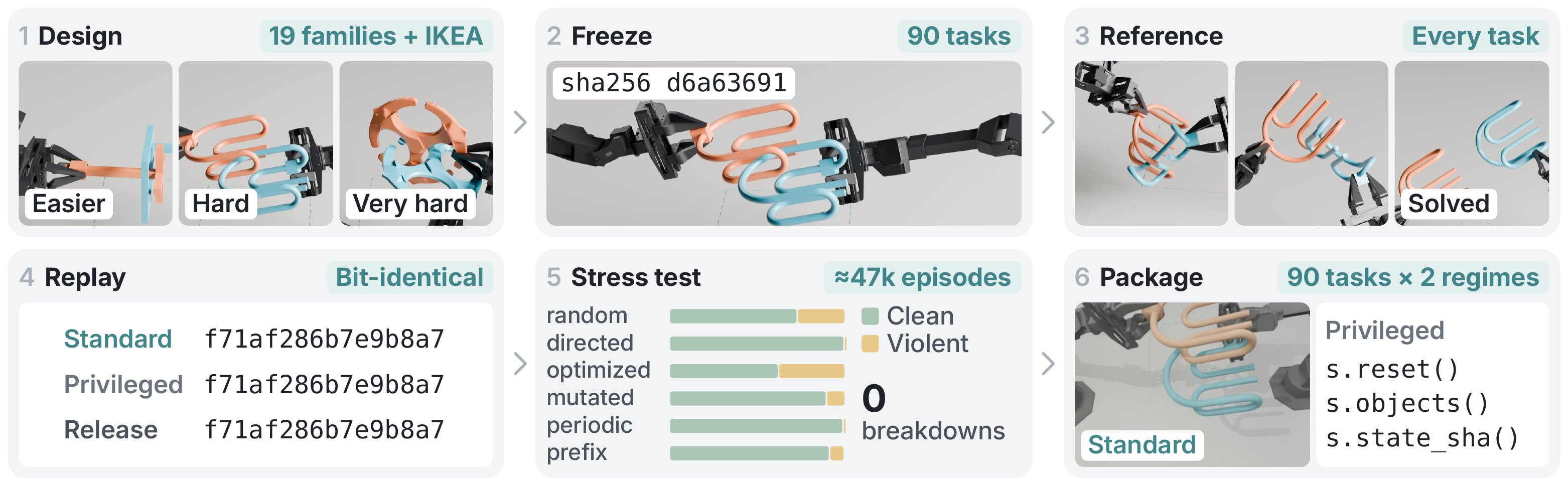}
  \caption{\textbf{Task curation pipeline}, shown for interlocked claws 05. Each task is designed at a difficulty
rating, frozen as one scene, solved by a verified reference, replayed bit-identically, stress-tested and packaged in both
regimes.}
  \label{fig:pipeline}
\end{figure}

\paragraph{Robot manipulation benchmarks and policies.}
Benchmarks for learned policies span single-arm, language-conditioned, household and dual-arm
tasks~\cite{james2020rlbench,yu2020metaworld,zhu2020robosuite,gu2023maniskill2,tao2024maniskill3,mees2022calvin,
liu2023libero,gong2023arnold,zhang2024vlabench,nasiriany2024robocasa,li2024behavior1k,pumacay2024colosseum,
garcia2025gembench,li2024simplerenv,mu2025robotwin,chen2026robotwin2,lin2026robowits}, mostly addressed with
vision-language-action models~\cite{brohan2023rt1,zitkovich2023rt2,team2024octo,kim2024openvla,black2024pi0,
black2025pi05,nvidia2025gr00t,fang2026molmoact2,team2025gemini15} or other imitation
policies~\cite{chi2023diffusion,zhao2023aloha,kim2026cosmospolicy} trained on
demonstrations~\cite{oxe2024,khazatsky2024droid}. Assembly~\cite{narang2022factory,tang2023industreal,tang2024automate,
zhang2026alphanist,tian2022assemble,tian2024asap,tian2025fabrica,lee2021ikea,heo2023furniturebench} and deformable
objects~\cite{lin2020softgym,seita2021deformableravens,huang2021plasticinelab,chen2023daxbench,lu2024garmentlab,
garciacamacho2020cloth,avigal2022speedfolding,nair2017rope,peng2024tiebot,grannen2020untangling,
sundaresan2021untangling,viswanath2022untangling,luo2024cablerouting} are studied one skill at a time. These tasks grow
harder with more objects or steps, while \name's puzzles are hard because of the little space a constraint leaves.

\paragraph{Spatial reasoning and physical puzzles.}
Motion planning treats narrow passages and disentanglement puzzles such as the alpha puzzle as hard
cases~\cite{kavraki1996prm,hsu2003bridge,amato1998alphapuzzle,zhang2020disentangle,sucan2012ompl,
chamzas2022motionbenchmaker}, task and motion planning adds discrete choices~\cite{garrett2021tamp}, and KinDER isolates
kinematic and dynamic reasoning~\cite{huang2026kinder}, for which coding agents now write
planners~\cite{merler2026coding}. A hand trained in simulation has solved a
Rubik's cube~\cite{akkaya2019rubik}. Model benchmarks test physical outcome
prediction~\cite{bakhtin2019phyre,bear2021physion,chow2025physbench}, spatial questions about images, video and
exploration~\cite{chen2024spatialvlm,yang2025thinking,wang2026mindcube,zhang2026tos,chen2026behaviorspatial}, egocentric
world modeling~\cite{wang2026enact,wang2025vagen,fan2026woven}, topological relations~\cite{mindtopo}, interlocking
puzzles as action sequences~\cite{chain2026} and skill acquisition~\cite{chollet2019measure,chollet2025arcagi2}, and
grade an answer or a symbolic plan. \name grades the executed motion under contact.

\paragraph{Coding agents from screens to robots.}
Coding agents are evaluated on operating systems~\cite{xie2024osworld,liu2023agentbench},
the web~\cite{zhou2023webarena}, terminals~\cite{terminalbench}, code
repositories~\cite{jimenez2024swebench,chan2024mlebench}, assistant tasks~\cite{mialon2024gaia} and task
length~\cite{kwa2025measuring}, where audits show that flawed graders distort
scores~\cite{zhu2025establishing,zhong2026impossiblebench}. On robots, language models write programs over perception
and motion APIs~\cite{liang2023codeaspolicies,singh2023progprompt,mu2024robocodex,ahn2022saycan,huang2022inner,
huang2023voxposer,huang2024rekep,ma2024eureka}, and recent agents run the whole loop~\cite{fu2026capx,xiao2026enpire,
tan2026forca,zhang2026unexpected,yang2026robotworld,ma2026rlebench,li2024eai,yang2025embodiedbench}. EmbodiedSWE~\cite{shen2026embodiedswe} and BiGym~2.0~\cite{zhang2026bigym2} draw their difficulty from long, dexterous
everyday tasks. \name also records where agents fail under spatial constraints, on the same
scenes with full access and with real-sensor inputs (\cref{tab:compare_full}).

\section{RoboARC}
\label{sec:roboarc}

A \name task is a frozen simulated scene, a robot, a goal stated in words and a grader that checks the simulated state.
A policy of any class must reach the goal within a wall-clock budget (about $3$\,h per task, \cref{sec:suite}).

\subsection{Problem formulation}
\label{sec:formulation}

We cast each task as a partially observable Markov decision process (POMDP)~\cite{kaelbling1998pomdp}
$\mathcal{M}=(\mathcal{S},\mathcal{A},T,\Omega,O,s_0,G,F)$. A state $s\in\mathcal{S}$ is the full configuration of the
scene: the robot's joints, the poses of rigid parts, the shape of deformable ones, and their contacts. An action
$a\in\mathcal{A}$ is a chunk of joint or end-effector targets, and the transition $T$ is the physics simulation,
deterministic from the frozen start state $s_0$. The grader defines a goal set $G\subset\mathcal{S}$ (a ring free of its
post, a rope without crossings) and a failure set $F\subset\mathcal{S}$, possibly empty (a forbidden contact, a part dropped). A policy
$\pi$ chooses $a_t\sim\pi(\cdot\mid \ell,h_t)$ from the task instruction $\ell$ and its history $h_t=(o_0,a_0,\dots,o_t)$, with observations
$o_t\sim O(\cdot\mid s_t)$, and succeeds if
\begin{equation}
  \exists\, t\le H:\; s_t\in G \quad\text{and}\quad s_k\notin F \;\;\forall k\le t,
  \label{eq:success}
\end{equation}
where $H$ is the last step the wall-clock budget allows. What makes the tasks hard is the shape of the feasible set:
the state trajectories that reach $G$ without touching $F$ form a thin tube around one path through a millimetre
clearance, an interlock or a knot, so a plan made from slightly wrong geometry falls outside it.

The two regimes share $\mathcal{S},\mathcal{A},T,G$ and $F$ and differ in what the policy can access. In the
\priv regime the policy observes the state itself, $O(s)=s$, and may reset to $s_0$, so it can try and check motions in
the simulator before it delivers one. In the \std regime it observes camera images, proprioception and the camera calibration, and must infer
$s_t$ from them within a single episode.

\subsection{Task suite}
\label{sec:suite}

\begin{wrapfigure}{r}{0.47\linewidth}
  \vspace{-10pt}\centering
  \includegraphics[width=\linewidth]{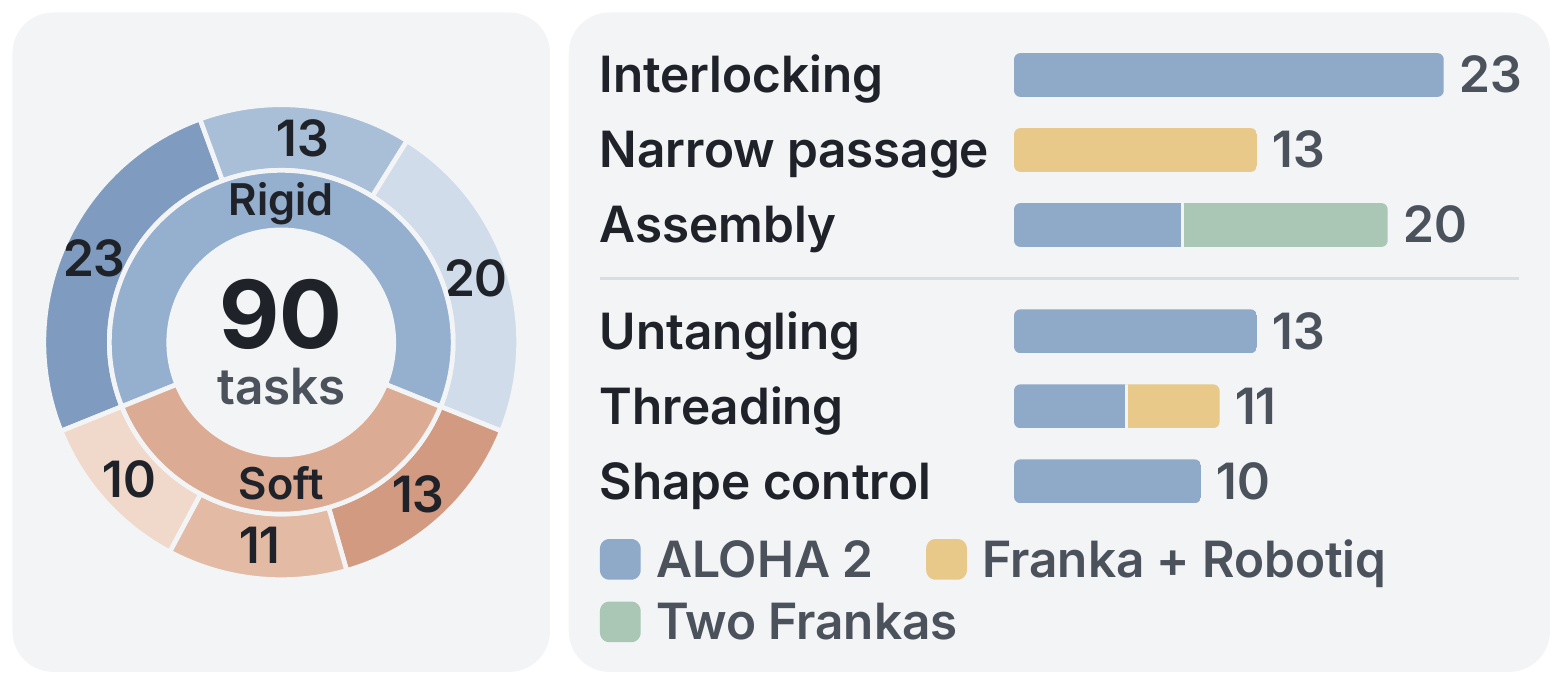}
  \captionsetup{font=small}
  \caption{\textbf{Six constraint types over rigid and soft materials, on three robots.} The ring splits the $90$ tasks into rigid and soft (inner) and into the six types (outer). Each bar counts the tasks of
one type, split by the robot they run on. Reference solution lengths are in the appendix (\cref{fig:reflen}).}
  \label{fig:stats}
  \vspace{-6pt}
\end{wrapfigure}

\paragraph{Constraint types.}
We organize the tasks by the spatial constraint that decides success and by material (\cref{fig:stats}). The rigid
types are \emph{interlocking} (separating or joining hooked or keyed parts), \emph{narrow passage} (guiding a held part
through millimetre clearances) and \emph{assembly} (fitting parts together, up to IKEA furniture). The soft types are
\emph{untangling} (untying knots, freeing rings caught on cords), \emph{threading} (passing a rope or fibers through
holes or a ring in a given order) and \emph{shape control} (unfolding, flattening or folding a blanket). Success is defined by the final state, often by its topology, and the same
rope is threaded in one family and untied in another.

\paragraph{Scale and simulation.}
\name has $90$ tasks: $78$ in $19$ puzzle families and $12$ IKEA scenes, most families at several difficulty ratings. They
run in MuJoCo~\cite{todorov2012mujoco} on three robots, ALOHA\,2~\cite{aldaco2024aloha2}, a Franka Panda with a Robotiq
gripper, or two Franka arms, with assisted grasping for rigid puzzle parts and patch grasps for cloth. Each puzzle has three domain-randomization seeds (\cref{fig:seeds}), and an agent's session is limited to
$190$\,min per task ($180$\,min for DeepSeek v4.1 Flash and GLM 5.3 Flash, $6$\,h for wire splicing). \Cref{sec:suppl_suite} gives the control rates, grasping rules,
simulation and rendering.

\subsection{Two observation regimes and three interfaces}
\label{sec:modes}

Each task ships as two Harbor~\cite{harbor} task directories, one per regime, which share the frozen scene, the physics and the
success rule (\cref{fig:modes}). In the \priv regime the simulator runs inside the agent's container, so the agent has
full access to it: the agent reads the simulator code and the full state, resets and replays freely, and delivers one
trajectory that the verifier replays twice from the frozen state. In the \std regime the simulator runs in a separate
container and the agent gets no object state: it sees RGB-D images, proprioception, calibrated cameras and the contact forces and penetration depths that the grader checks, submits
action chunks in a single episode without resets, and the live episode is graded. Coding agents reach it through a
fixed observation and action API, and VLA policies through a Gymnasium environment~\cite{towers2024gymnasium}
with the same physics and grader (\cref{sec:suppl_modes}).

\subsection{Task curation}
\label{sec:validity}

Every task passes the pipeline of
\cref{fig:pipeline} before release. It is designed at a difficulty rating, frozen as one scene, and ships a reference
solution that its own grader accepts in both regimes, while a do-nothing agent fails. The simulation is deterministic,
so every reference replay ends on the same state hash on every CPU class we tested. Before release, $46{,}662$ stress
episodes drove the puzzle scenes, and no command earned a success the task does not grant (\cref{sec:suppl_validity}). A replay of $250$ solved puzzle trials of the coding agents records every contact at every simulator step: the deepest contact of a task part has a median of $0.29$\,mm and a maximum of $2.13$\,mm, against $0.22$ and $2.18$\,mm in the reference solutions (\cref{sec:suppl_audit}).

\section{Experiments}
\label{sec:experiments}

\subsection{Setup}
\label{sec:setup}

\paragraph{Policy classes.}
We evaluate two policy classes, each as released, the way a robot meets a problem it has never seen. Coding agents receive the instruction and write their own code. In the \priv regime they plan and check their plans in simulation, and in the \std regime they write their own
perception. VLA policies run in the \std regime. We evaluate four public VLA policies fine-tuned on DROID~\cite{khazatsky2024droid} on the $18$
puzzles that use their robot, one Franka arm with a Robotiq 2F-85 gripper ($13$ in narrow passage and $5$ in threading). Each is called as its authors' DROID client calls it, with DROID's camera poses, gripper mounting and start pose (\cref{sec:suppl_vla}).

\paragraph{Agents.}
An agent is a harness, a model and a reasoning effort, evaluated as one system. We evaluate seven: GPT-6 Astra and
GPT-6.1 Sol (xhigh) in Codex, Claude Opus 5.5 (xhigh) in Claude Code, Gemini 4 Argon (high) in the Antigravity SDK, and
three open-weight models, Kimi 3.1 Pro (max) in Kimi Code, DeepSeek v4.1 Flash (high) in DeepSeek Harness (DSH) and
GLM 5.3 Flash (high) in Claude Code. Each receives only the task's instruction, with no task-specific primitives or
demonstrations, the time limit of \cref{sec:suite}, and network access limited to package indices and the model API.

\paragraph{Protocol.}
Each trial runs one task in one regime in Harbor with one GPU and the task's declared CPU and memory, under the time limit of \cref{sec:suite}, and a separate verifier grades it (\cref{sec:modes}). In the \priv regime the agent shares the GPU with the simulator. In the \std regime only the simulator uses it.  Trials that end in an infrastructure or grader error are rerun. We report the success rate over all $90$ tasks, overall and per constraint type. The
overall rate weighs every task equally, and \cref{sec:suppl_family} gives the ranking with every family weighted equally. Each task runs once with each of three
domain-randomization seeds, and we report the mean success rate and its standard deviation over the seeds
(\cref{tab:main,tab:main_sd}). The per-family results are in the appendix (\cref{tab:family}).

\definecolor{privg}{HTML}{8F98A3}
\DeclareRobustCommand{\swatch}[2]{{\setlength{\fboxsep}{1.2pt}\colorbox{#1}{#2}}}
\begin{table}[t!]
\centering
\caption{\textbf{Main results.} Success rate (\%) of coding agents per constraint type in the \std and \priv regimes,
averaged over three runs, with the standard deviation in the All column. Darker shading means a higher rate,
\swatch{stdc!50}{teal} for \std and \swatch{privg!45}{grey} for \priv, and bold marks the best per column. Columns: interlocking, narrow passage, assembly, untangling, threading, shape control.}
\label{tab:main}
\scriptsize
\setlength{\tabcolsep}{2.3pt}
\renewcommand{\arraystretch}{1.15}

\begin{fittab}\begin{tabular}{@{}l l *{6}{c} c @{\hspace{7pt}} *{6}{c} c@{}}
\toprule
 & & \multicolumn{7}{c}{\textcolor{stdink}{\textbf{Standard}}} & \multicolumn{7}{c}{\textcolor{privink}{\textbf{Privileged}}} \\
\cmidrule(lr){3-9} \cmidrule(l){10-16}
Agent & Harness (effort) & Interl. & Passage & Assem. & Untang. & Thread. & Shape & \textbf{All} & Interl. & Passage & Assem. & Untang. & Thread. & Shape & \textbf{All} \\
\midrule
\multicolumn{16}{@{}l}{\textsc{\tiny proprietary}} \\
GPT-6 Astra & Codex \textcolor{black!50}{(xhigh)} & \cellcolor{stdc!33}\textbf{$\mathbf{59.4}$} & \cellcolor{stdc!31}\textbf{$\mathbf{56.4}$} & \cellcolor{stdc!6}\textbf{$\mathbf{10.0}$} & \cellcolor{stdc!28}$51.3$ & \cellcolor{stdc!17}\textbf{$\mathbf{30.3}$} & \cellcolor{stdc!29}\textbf{$\mathbf{53.3}$} & \cellcolor{stdc!23}\textbf{$\mathbf{42.6}$}{\tiny\textcolor{black!50}{$\pm 5.1$}} & \cellcolor{privg!34}$94.2$ & \cellcolor{privg!30}$84.6$ & \cellcolor{privg!31}\textbf{$\mathbf{85.0}$} & \cellcolor{privg!22}$61.5$ & \cellcolor{privg!32}\textbf{$\mathbf{87.9}$} & \cellcolor{privg!35}$96.7$ & \cellcolor{privg!31}\textbf{$\mathbf{85.6}$}{\tiny\textcolor{black!50}{$\pm 4.4$}} \\
GPT-6.1 Sol & Codex \textcolor{black!50}{(xhigh)} & \cellcolor{stdc!29}$52.2$ & \cellcolor{stdc!28}$51.3$ & \cellcolor{stdc!5}$8.3$ & \cellcolor{stdc!25}$46.2$ & \cellcolor{stdc!8}$15.2$ & \cellcolor{stdc!18}$33.3$ & \cellcolor{stdc!19}$34.8${\tiny\textcolor{black!50}{$\pm 5.0$}} & \cellcolor{privg!33}$91.3$ & \cellcolor{privg!34}\textbf{$\mathbf{94.9}$} & \cellcolor{privg!25}$68.3$ & \cellcolor{privg!23}$64.1$ & \cellcolor{privg!27}$75.8$ & \cellcolor{privg!34}$93.3$ & \cellcolor{privg!29}$81.1${\tiny\textcolor{black!50}{$\pm 1.1$}} \\
Claude Opus 5.5 & Claude Code \textcolor{black!50}{(xhigh)} & \cellcolor{stdc!25}$44.9$ & \cellcolor{stdc!21}$38.5$ & $1.7$ & \cellcolor{stdc!28}$51.3$ & \cellcolor{stdc!7}$12.1$ & \cellcolor{stdc!15}$26.7$ & \cellcolor{stdc!16}$29.3${\tiny\textcolor{black!50}{$\pm 1.7$}} & \cellcolor{privg!34}\textbf{$\mathbf{95.7}$} & \cellcolor{privg!30}$84.6$ & \cellcolor{privg!22}$61.7$ & \cellcolor{privg!24}\textbf{$\mathbf{66.7}$} & \cellcolor{privg!29}$81.8$ & \cellcolor{privg!35}$96.7$ & \cellcolor{privg!29}$80.7${\tiny\textcolor{black!50}{$\pm 0.6$}} \\
Gemini 4 Argon & Antigravity \textcolor{black!50}{(high)} & \cellcolor{stdc!21}$37.7$ & \cellcolor{stdc!24}$43.6$ & $1.7$ & \cellcolor{stdc!30}\textbf{$\mathbf{53.8}$} & \cellcolor{stdc!7}$12.1$ & \cellcolor{stdc!26}$46.7$ & \cellcolor{stdc!17}$30.7${\tiny\textcolor{black!50}{$\pm 4.6$}} & \cellcolor{privg!32}$88.4$ & \cellcolor{privg!22}$61.5$ & \cellcolor{privg!18}$50.0$ & \cellcolor{privg!23}$64.1$ & \cellcolor{privg!31}$84.8$ & \cellcolor{privg!36}\textbf{$\mathbf{100.0}$} & \cellcolor{privg!26}$73.3${\tiny\textcolor{black!50}{$\pm 4.0$}} \\
\addlinespace[2pt]
\multicolumn{16}{@{}l}{\textsc{\tiny open-weight}} \\
Kimi 3.1 Pro & Kimi Code \textcolor{black!50}{(max)} & \cellcolor{stdc!5}$8.7$ & $2.6$ & $0$ & \cellcolor{stdc!11}$20.5$ & \cellcolor{stdc!2}$3.0$ & \cellcolor{stdc!6}$10.0$ & \cellcolor{stdc!4}$7.0${\tiny\textcolor{black!50}{$\pm 1.7$}} & \cellcolor{privg!15}$42.0$ & \cellcolor{privg!9}$25.6$ & \cellcolor{privg!6}$16.7$ & \cellcolor{privg!14}$38.5$ & $3.0$ & \cellcolor{privg!24}$66.7$ & \cellcolor{privg!11}$31.5${\tiny\textcolor{black!50}{$\pm 3.6$}} \\
DeepSeek v4.1 Flash & DSH \textcolor{black!50}{(high)} & \cellcolor{stdc!6}$11.6$ & $0$ & $0$ & \cellcolor{stdc!8}$15.4$ & $0$ & \cellcolor{stdc!2}$3.3$ & \cellcolor{stdc!3}$5.6${\tiny\textcolor{black!50}{$\pm 1.9$}} & \cellcolor{privg!6}$15.9$ & \cellcolor{privg!3}$7.7$ & $1.7$ & \cellcolor{privg!6}$17.9$ & $0$ & \cellcolor{privg!8}$23.3$ & \cellcolor{privg!4}$10.7${\tiny\textcolor{black!50}{$\pm 1.3$}} \\
GLM 5.3 Flash & Claude Code \textcolor{black!50}{(high)} & \cellcolor{stdc!3}$5.8$ & $2.6$ & $0$ & \cellcolor{stdc!4}$7.7$ & $0$ & \cellcolor{stdc!6}$10.0$ & \cellcolor{stdc!2}$4.1${\tiny\textcolor{black!50}{$\pm 2.8$}} & \cellcolor{privg!2}$5.8$ & $0$ & \cellcolor{privg!2}$5.0$ & \cellcolor{privg!2}$5.1$ & $0$ & \cellcolor{privg!4}$10.0$ & \cellcolor{privg!2}$4.4${\tiny\textcolor{black!50}{$\pm 2.2$}} \\
\bottomrule
\end{tabular}\end{fittab}
\end{table}

\subsection{Main results}
\label{sec:main_results}

\Cref{tab:main} gives the coding agents' success rates by constraint type. Two results stand out.

\paragraph{In the \priv regime, the proprietary agents solve most tasks.}
In the \priv regime the four proprietary agents solve $73\%$ to $86\%$ of the tasks, and each solves at least $85\%$ of the interlocking and shape-control tasks. The open-weight agents solve at most $31\%$. 

\paragraph{In the \std regime, the same agents keep at most half of these successes.}
The strongest agent, GPT-6 Astra, solves $43\%$ of the tasks, and the four proprietary agents keep $36\%$ to $50\%$ of their \priv success rate. Every family except wire splicing has a task that some agent solves in this regime,
$62$ of the $90$ tasks in all. The rates per difficulty rating are in the appendix (\cref{tab:tiers}).  The loss is largest in assembly and threading, $63$ to $65$ points on average for GPT-6 Astra, GPT-6.1 Sol and Claude Opus 5.5
(\cref{fig:gap}). In assembly no agent exceeds $10\%$, and in threading the strongest solves $30\%$. The open-weight agents solve at most $7\%$ of the tasks in this regime.

\begin{wrapfigure}{r}{0.44\linewidth}
  \vspace{-10pt}\centering
  \includegraphics[width=\linewidth]{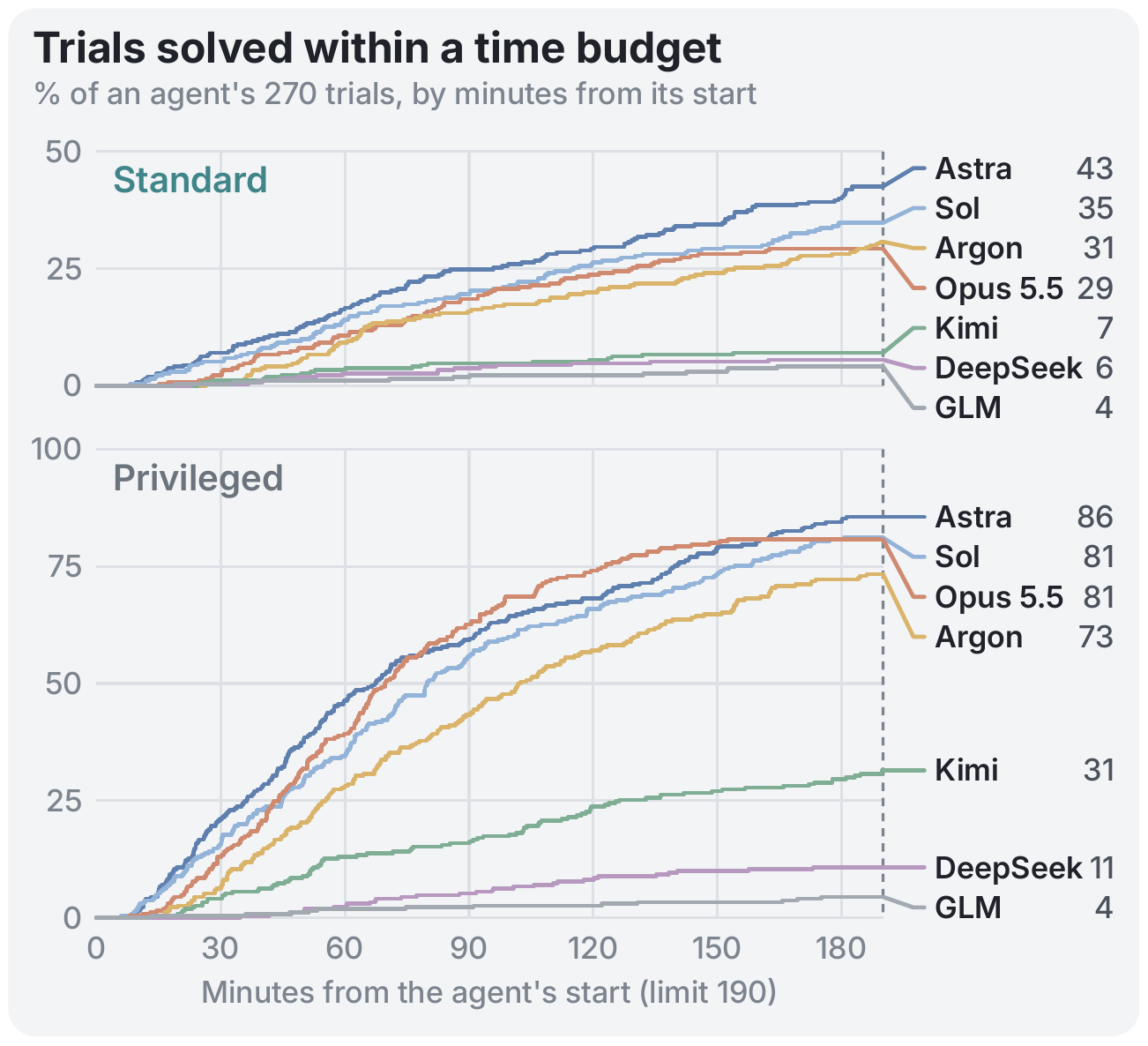}
  \captionsetup{font=small}
  \caption{\textbf{Trials solved within a time budget.} Share of an agent's $270$ trials per regime that are solved within $t$ minutes of its start, read from runs made under the full limit. The dashed line marks $180$\,min. Labels give the rate at the limit.}
  \label{fig:budget}
  \vspace{-6pt}
\end{wrapfigure}
\paragraph{Most trials are solved well inside the time limit.}
\Cref{fig:budget} reads every solved trial at the minute at which it was solved, counted from the agent's start
(\cref{sec:suppl_budget}). No agent solves a task in the first five minutes. GPT-6 Astra solves $46\%$ of its \priv trials and $17\%$ of its \std
trials within one hour, and $68\%$ and $29\%$ within two. Over all seven agents a solved trial takes a median of
$65$\,min in the \priv regime and $77$\,min in the \std regime. The curves of GPT-6 Astra, GPT-6.1 Sol and Gemini 4
Argon still rise at the limit, where each solves between $11$ and $15$ trials per regime after minute $160$, and the
curve of Claude Opus 5.5 is flat by then. In the \priv regime Claude Opus 5.5 has solved the most trials between
minutes $80$ and $160$, and GPT-6 Astra before and after.

\begin{table}[t!]
\centering
\caption{\textbf{VLA policies solve none of the $\mathbf{18}$ one-arm puzzles.} Success rate (\%) of four public VLA
policies fine-tuned on DROID in the \std regime, three runs per puzzle. The grey row gives GPT-6 Astra on the same puzzles for reference (\cref{tab:vla_agents}). Passage: narrow passage, Thread.: threading. Numbers in parentheses count
the puzzles. Touched: share (\%) of episodes in which the gripper touches the task object. Setup, inputs and an example episode: \cref{sec:suppl_vla}.}
\label{tab:vla}
\footnotesize
\setlength{\tabcolsep}{3pt}
\begin{fittab}\begin{tabular}{@{}l c c c c@{}}
\toprule
Policy & Passage ($13$) & Thread. ($5$) & All ($18$) & Touched \\
\midrule
$\pi_{0.5}$ & $0$ & $0$ & $0$ & $33.3$ \\
MolmoAct2 & $0$ & $0$ & $0$ & $35.2$ \\
GR00T N1.7 & $0$ & $0$ & $0$ & $31.5$ \\
Cosmos3-Nano-Policy & $0$ & $0$ & $0$ & $61.1$ \\
\midrule
\textcolor{black!60}{GPT-6 Astra (Codex)} & \textcolor{black!60}{$56.4$} & \textcolor{black!60}{$6.7$} & \textcolor{black!60}{$42.6$} & \textcolor{black!60}{--} \\
\bottomrule
\end{tabular}\end{fittab}
\end{table}

\paragraph{Zero-shot VLA policies solve none of their $18$ puzzles.}
\Cref{tab:vla} gives the zero-shot result of the VLA policies. The four policies touch the task object in $31\%$ to $61\%$ of the episodes but solve
none of the $216$, which sets the zero-shot reference point for learned policies on \name. \Cref{tab:vla_agents} lists the seven coding agents on the same puzzles, where GPT-6 Astra solves $43\%$ in the \std regime.

\subsection{Transfer to a real robot}
\label{sec:real}

The \std regime gives an agent camera images and proprioception, as a real robot does, so the same kind of interface can drive hardware.

\begin{wrapfigure}{r}{0.40\linewidth}
  \vspace{-10pt}\centering
  \includegraphics[width=\linewidth]{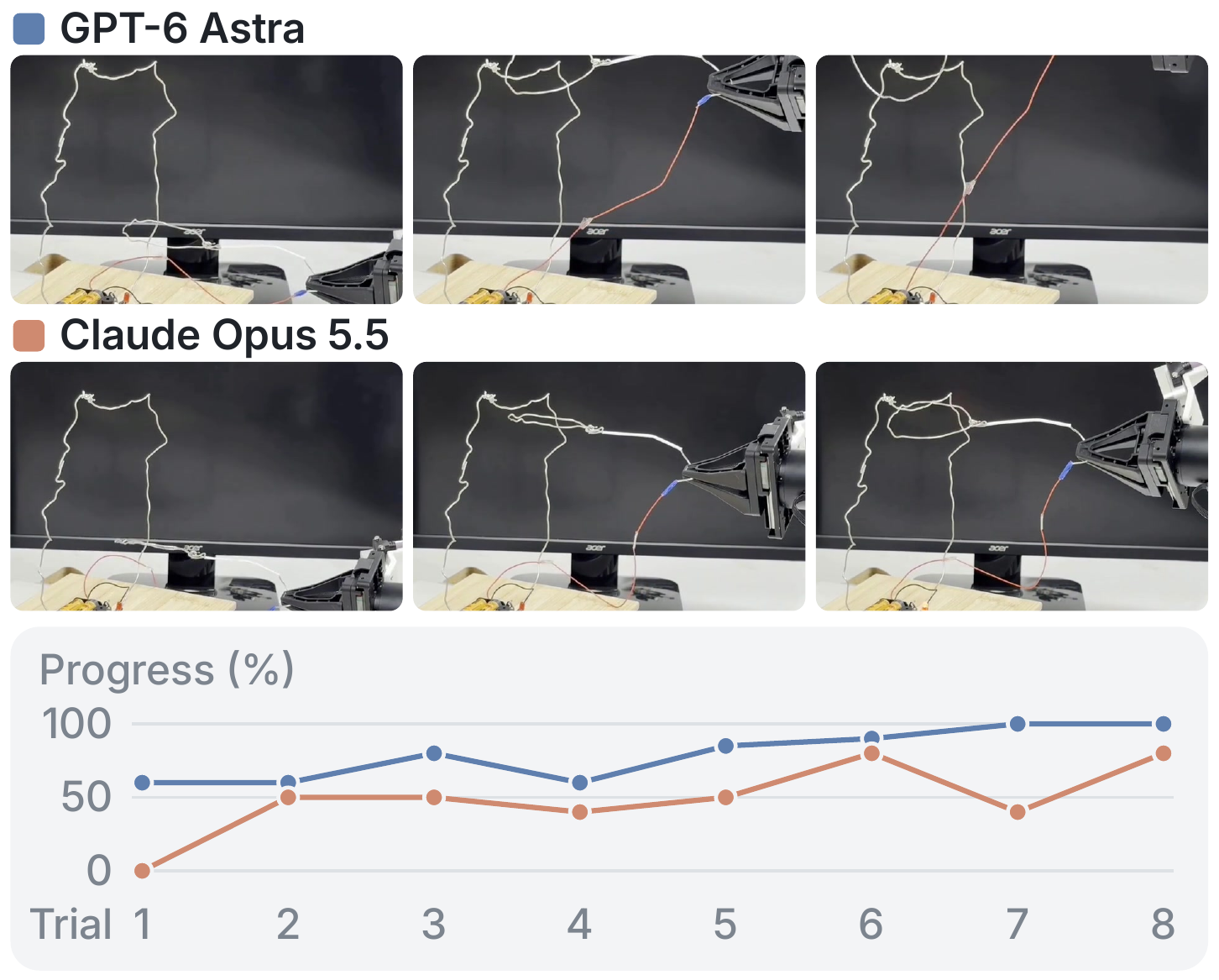}
  \captionsetup{font=small}
  \caption{\textbf{Two coding agents play a real buzz wire through the same kind of interface.} \emph{Top:}
frames from one trial per agent. \emph{Bottom:} progress, the share of the wire covered before the loop touches it, over eight consecutive trials per agent.}
  \label{fig:real}
  \vspace{-6pt}
\end{wrapfigure}
We connected two coding agents to a single YAM arm through the same kind of
interface, with absolute end-effector commands and three RealSense cameras (\cref{sec:suppl_real}), and let each play eight consecutive trials
of a real buzz wire game (\cref{fig:real}). GPT-6 Astra covers more
of the wire than Claude Opus 5.5 in every trial: $65\%$ against $35\%$ over the first four trials and $94\%$
against $63\%$ over the last four, and it passes the whole wire in trials $7$ and $8$. This order agrees with their overall ranking in the \std regime
($43\%$ against $29\%$, \cref{tab:main}). On the five simulated buzz-wire tasks, GPT-6 Astra also covers
more of the wire than Claude Opus 5.5 in the \std regime, $59\%$ against $41\%$ on average over all runs.
\par\WFclear

\section{Analysis}
\label{sec:analysis}

The strongest agent loses half of its successes between the two regimes (\cref{tab:main}). \Cref{sec:ladder} gives
information back to GPT-6 Astra inside the \std regime and measures what each kind is worth. \Cref{sec:endings} replays the runs of the seven coding agents and
records where failed runs stop and what form failed attempts take.

\subsection{Poses and geometry are synergistic for GPT-6 Astra}
\label{sec:ladder}

\begin{wrapfigure}{r}{0.47\linewidth}
  \vspace{-10pt}\centering
  \includegraphics[width=\linewidth]{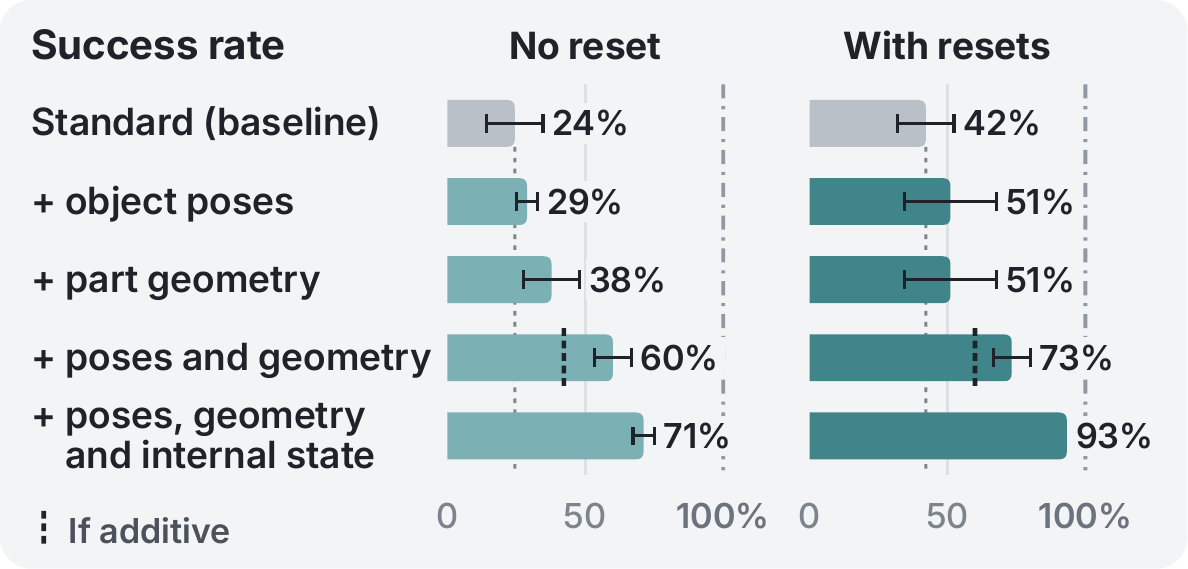}
  \captionsetup{font=small}
  \caption{\textbf{Poses and geometry are synergistic for GPT-6 Astra.} Success of GPT-6 Astra (high effort) on the $15$ puzzles in the \std regime, with one kind of \priv information given back per row (mean of three runs, whisker $\pm1$ standard deviation). Black tick: the gains of poses and of geometry added.}
  \label{fig:ladder}
  \vspace{-6pt}
\end{wrapfigure}

To find which information GPT-6 Astra is missing, we take the $15$ rigid puzzles that GPT-6 Astra at high effort solved in the \priv regime and
failed in the \std regime in earlier runs (\cref{sec:suppl_ablation}), and give back one kind of \priv information at a time inside the \std regime (\cref{fig:ladder}).

\needspace{14\baselineskip}
Each arm, starting with the unchanged \std regime, runs three times on every task, and we report the mean and the standard
deviation over the three runs.

Object poses alone add little ($+4.4$ points). A pick or a placement can be planned from a pose, but a constrained
motion also needs the shape it must clear. The parts' geometry alone (shapes and grasp regions, each in its part's frame) adds $+13.3$. Given both, success rises from $24.4\,\%$ to $60.0\,\%$, a gain of $+35.6$ points, $17.8$ (standard error $9$) more than the two added. Resets alone add $+17.8$ points. With resets, poses or geometry alone add $+8.9$ each and both add $+31.1$. Adding the internal state (joint positions and velocities, poses of moving links, grasp phase and mechanism readings) brings the agent to
$71.1\,\%$ without resets and $93.3\,\%$ with them. In the main-table runs at xhigh effort, GPT-6 Astra solves $15.6\,\%$ of these puzzles in the \std regime and $91.1\,\%$ in the \priv regime.

\par\WFclear
\subsection{How the agents solve tasks in the \priv regime}
\label{sec:privmethods}

\begin{figure}[t!]
  \centering
  \includegraphics[width=\textwidth]{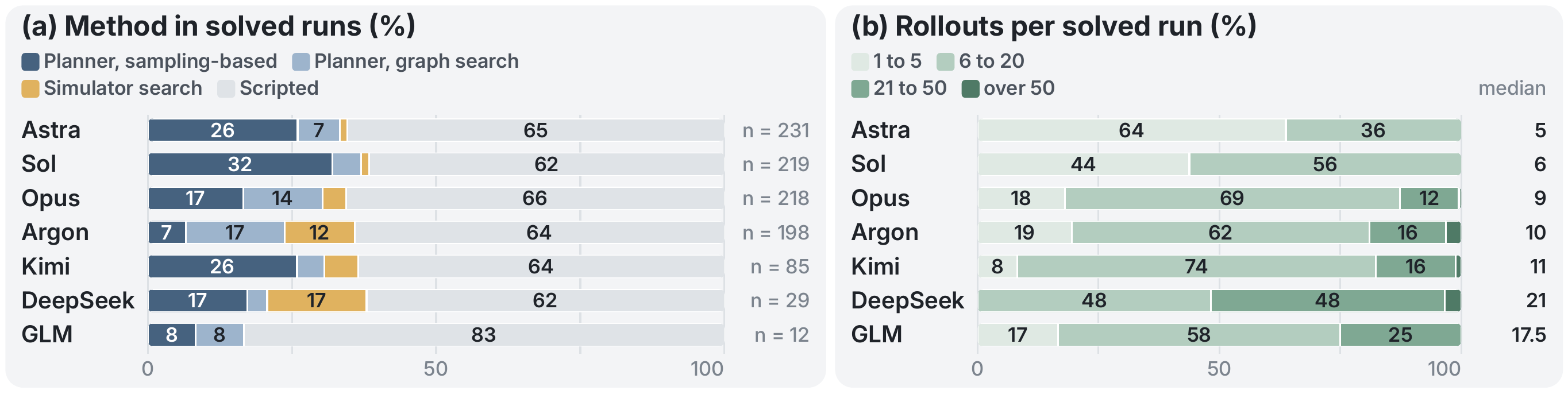}
  \caption{\textbf{Methods and rollouts of the solved \priv trials.} (a) Share of each agent's solved trials whose code
  contains a sampling-based planner, a graph-search planner, a search with simulator rollouts, or scripted waypoints
  only. (b) Rollouts per solved trial, with the median on the right.}
  \label{fig:privmethods}
\end{figure}

The \priv regime gives an agent the simulator, so a solved trial can come from a planner, from a search over
rollouts, or from a motion the agent derives and corrects. We classify all $1{,}890$ \priv trials by the code and the
commands of the agent, with rules on what the code contains (\cref{sec:suppl_privmethods}). Of the $992$ solved trials,
$64\,\%$ script waypoints computed from the state and the geometry and correct them over a few trials, $31\,\%$ contain
a motion planner, and $5\,\%$ search with simulator rollouts or a global optimiser (\cref{fig:privmethods}a). The
planners belong to the two rigid constraint types. They appear in $65\,\%$ of the solved interlocking trials and
$69\,\%$ of the solved narrow-passage trials, and in $3\,\%$ of the solved trials of the other four types. The agents
write these planners themselves: $259$ of the $305$ trials with a planner use only code the agent wrote, and $93\,\%$
plan on a geometric model that the agent builds from the scene.

A rollout is a stretch of simulation between two resets. A solved trial uses a median of $7$ rollouts, and $69\,\%$ of
the solved trials use at most $10$ (\cref{fig:privmethods}b). No solved trial of GPT-6 Astra or GPT-6.1 Sol uses more
than $20$. Failed trials search more, with a median of $18$ rollouts. Among the
four proprietary agents, $79\,\%$ of the trials with a planner and $86\,\%$ of the scripted trials are solved.

The spatial constraints of \name are defined for the robot that executes the motion with its grasp, and a classical
planner baseline on the free piece shows what the robot adds (\cref{sec:suppl_planner}). Given the exact
geometry and the reference solution's goal pose, RRT-Connect~\cite{kuffner2000rrtconnect} or a probabilistic
roadmap~\cite{kavraki1996prm} from OMPL~\cite{sucan2012ompl} finds a collision-free route for the free piece on $17$ of
the $36$ interlocking and narrow-passage puzzles in at least one run, $9$ of them within a minute. On the $9$ of these
puzzles where a scripted execution reproduces the reference solution, RRT-Connect plans $31$ routes from the reference
solution's grasp, and none is executable by the robot: $24$ leave the arm's reach or joint limits, and $7$ drive the
robot into the puzzle or the table. The hard part is turning a feasible route of the piece into a motion the robot can
carry out.

\subsection{Failure modes and agent strategies}
\label{sec:endings}

\begin{figure}[t!]
  \centering
  \includegraphics[width=\textwidth]{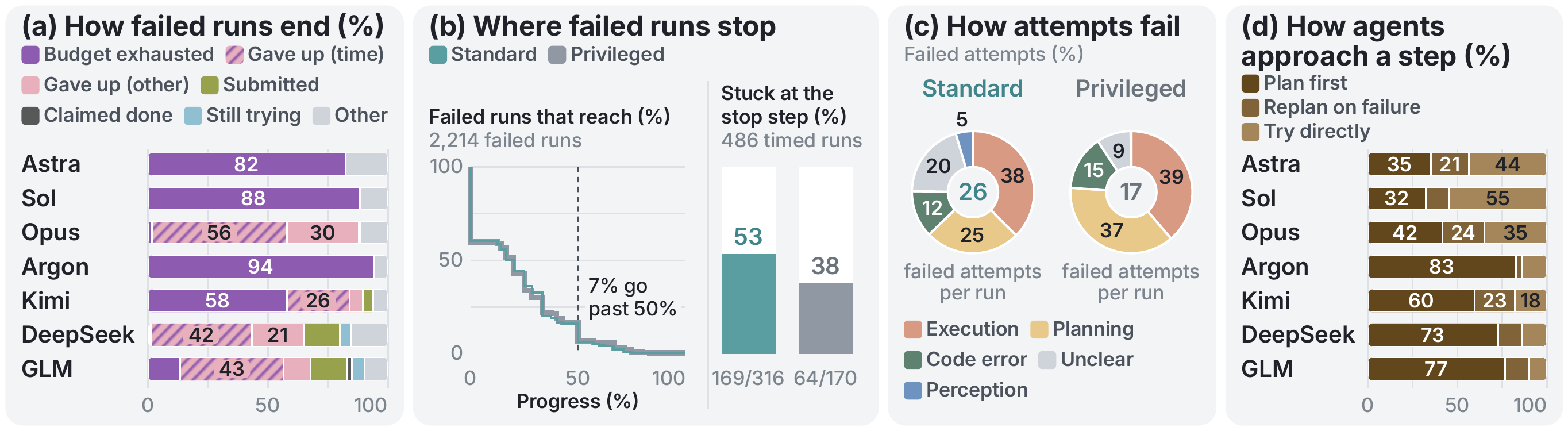}
  \caption{\textbf{How the coding agents' failed runs end, where they stop, what form failed attempts take and how agents approach a step.}
(a) $69\,\%$ of failed runs end with the budget used up or with the agent giving up for lack of time (hatched).
(b) Left: $7\,\%$ of the $2{,}214$ failed runs pass more than half of the task's stages. Right: share of timed failed runs stuck on their stop step. (c) Planning and Execution are
the most common forms of failed attempts (ring centre: mean failed attempts per run). (d) Astra, Sol and Opus plan first less often than the
other agents.}
  \label{fig:endings}
  \end{figure}
We replay $3{,}609$ runs of the seven agents, $2{,}214$ of them failed, and label the transcripts of a sample of $1{,}210$. Each failed attempt (a try at a step that misses the step's goal, in failed or solved runs) gets one form, and one step in each of $301$ sampled runs ($43$ per agent) gets one approach. The rules and checks are in \cref{sec:suppl_endings}.

\paragraph{Failed runs of the coding agents make little progress and many are stuck on one step.}
In $69\,\%$ of the $2{,}214$ failed runs, the agent uses up its budget ($40.4\,\%$) or gives up because it says time is running out ($28.5\,\%$, \cref{fig:endings}a). Only $7\,\%$ of the failed runs pass more than half of the task's stages (\cref{fig:endings}b, left). In a sample of $486$ failed runs with step times, $53\,\%$ in the \std and $38\,\%$ in the \priv regime are stuck on the step where they stop: it takes at least a third of the run and three or more failed attempts (right, \cref{sec:suppl_time}). The stop step is often the task's key step, the one that
passes its constraint: $34\,\%$ of the failed runs stop there, against $17\,\%$ for a random step. In the \priv regime a
reached key step is solved as often as the other steps ($82\,\%$ against $80\,\%$). In the \std regime it is solved in
$52\,\%$ of cases and the other steps in $63\,\%$ (\cref{fig:suppl_keystep}).

\begin{figure}[t!]
  \centering
  \includegraphics[width=\textwidth]{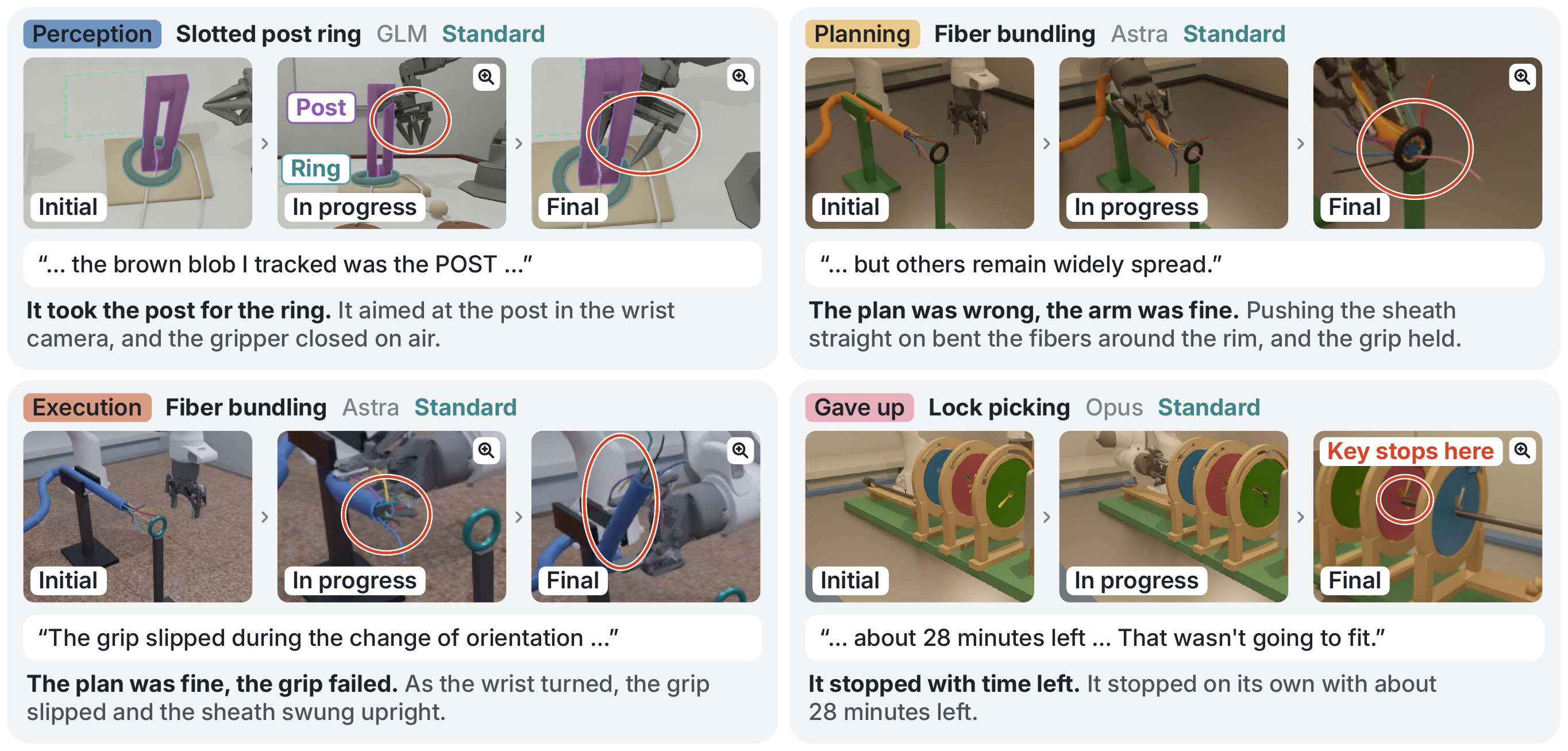}
  \caption{\textbf{Four failed runs in the \std regime: three forms of failed attempts and one way a run ends.} Each panel gives the label, the task and the agent, three frames of the run, a quote from the agent's log and what went wrong. Red rings mark the error, and a magnifier marks a zoomed frame. In the Perception panel the post and the ring are coloured. The Planning and Execution panels are two runs of GPT-6 Astra on the same step, so one constraint can fail in either form.}
  \label{fig:cases}
  \end{figure}

\paragraph{Most of the coding agents' failed attempts are Planning or Execution errors.}
Each label names what happens in a failed attempt (\cref{tab:suppl_labels}), so a part's shape estimated wrong is recorded where it surfaces, as a plan that collides or a grip that slips. Perception: the agent's estimate of an object's position or size is wrong, as when it takes the post for the ring (\cref{fig:cases}). Planning: the robot runs the code as written, but the commands cannot reach the step's goal, for example a wrong target, an unreachable pose or a collision. Execution: the commands would reach the goal, but the robot or the object does not follow them. In the \std regime Execution makes up $38\,\%$ of the failed attempts, Planning $25\,\%$ and Perception $5\,\%$, and in the \priv regime Planning and Execution make up three quarters (\cref{fig:endings}c).

\needspace{14\baselineskip}
\begin{wrapfigure}{r}{0.47\linewidth}
  \vspace{-10pt}\centering
  \includegraphics[width=\linewidth]{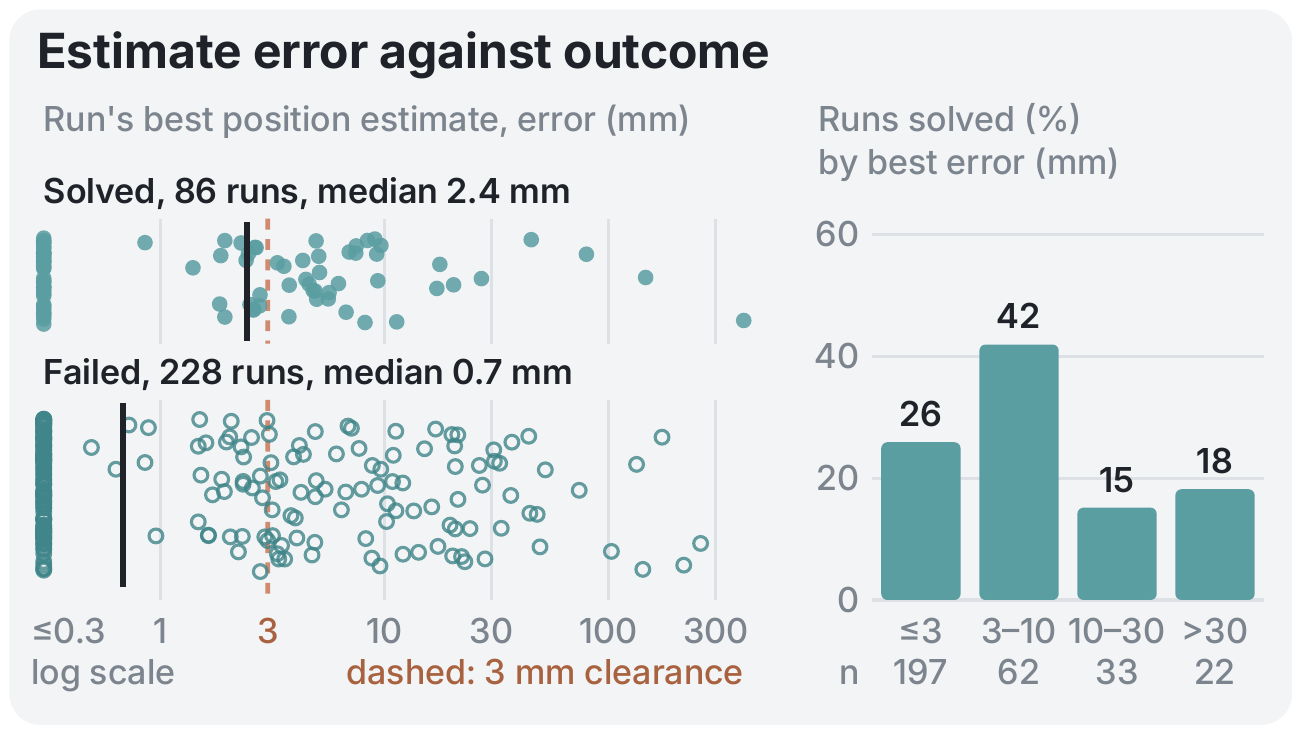}
  \captionsetup{font=small}
  \caption{\textbf{The agents' position estimates against the outcome.} Left: error of a trial's best position estimate of a task part, for solved and failed \std trials on the rigid puzzles. Right: share of trials solved, by that error.}
  \label{fig:estimates}
  \vspace{-6pt}
\end{wrapfigure}
\paragraph{The agents locate parts to millimetres and misjudge their extent.}
For the \std trials on the rigid puzzles, we compare the positions and boxes that an agent writes down for a task part
with the simulator's values at that moment (\cref{sec:suppl_estimates}). In the $314$ trials with a usable estimate,
the best position estimate of a trial lies a median of $1.5$\,mm from the part. Solved and failed trials do not differ
in this error, which exceeds $3$\,mm in $41\,\%$ of the solved and $36\,\%$ of the failed trials
(\cref{fig:estimates}). The extent of a part is estimated less well. In the $208$ trials that state a box for a part,
the best box is off by a median of $13.9$\,mm on its worst face, in solved and failed trials alike. This agrees with
\cref{sec:ladder}, where object poses add little and part geometry adds more.

\paragraph{Astra, Sol and Opus act first, while the other agents think first.}
For one sampled step in each run, we read the log up to the third attempt at that step and label how the agent approaches it. Plan first: before its first attempt, the agent computes a clearance, path or pose from numbers or explicit geometry. Replan on failure: it computes nothing first, but after a miss it measures again or changes the path or pose. Try directly: after a miss it only changes a step size, a force or a threshold. Astra, Sol and Opus, the three agents with the highest \priv rates, plan first in only $35\,\%$, $32\,\%$ and $42\,\%$ of their sampled steps, the other four agents in $60$ to $83\,\%$ (\cref{fig:endings}d, \cref{sec:suppl_endings}).

\subsection{What the failures look like}
\label{sec:qualitative}

\Cref{fig:cases} shows four failed runs of the \std regime. In the Perception case (GLM, slotted post ring), the agent takes the post for the ring in the wrist camera, so the gripper closes on air. In the Planning case (Astra, fiber bundling), the agent pushes the sheath straight on as planned. The grip holds, but the fibers bend around the ring's rim. In the Execution case (Astra, the same step in another run), the plan is sound, but the grip slips as the wrist turns and the sheath swings upright. In the Gave up case (Opus, lock picking), the agent stops on its own with about $28$ minutes of its time left, citing time. Closing these gaps calls for estimating poses and part geometry from the cameras, planning the key step before acting, and correcting each motion from feedback as it runs.

\begin{findingbox}{Where Coding Agents Fail}
\begin{itemize}[leftmargin=1.1em, topsep=0pt, partopsep=0pt, itemsep=2pt, parsep=0pt]
  \item \textbf{Poses and geometry are synergistic for GPT-6 Astra.} Alone they add $4$ and $13$ points, together $36$ ($24\,\%$ to $60\,\%$).
  \item \textbf{Failed runs make little progress and many are stuck on one step.} Only $7\,\%$ pass more than half of a task, and a third stop at the key step.
\end{itemize}
\end{findingbox}

\section{Conclusion}
\label{sec:conclusion}

We introduced \name, a benchmark that asks whether robots can reason out the right motions under spatial constraints. Its $90$ tasks in six constraint types each have a physics-based grader and a verified reference solution. The strongest coding agent solves $86\%$ of the tasks in the \priv regime and $43\%$ in the \std regime, and the four zero-shot VLA policies solve none of their $18$ one-arm puzzles. On $15$ puzzles that GPT-6 Astra solved in the \priv regime and failed in the \std regime, object poses and part geometry add $4$ and $13$ points alone and $36$ together, which raises GPT-6 Astra's success from $24\,\%$ to $60\,\%$. Only $7\,\%$ of the coding agents' failed runs pass more than half of a task, $53\,\%$ of the failed runs in the \std regime are stuck on one step, and a third stop at the step that passes the constraint. Most of their failed attempts are Planning or Execution errors. Frontier coding agents already reason out the right motions for many of these tasks once they know where the parts are and what shape they have. From their own sensors, in a single attempt, the strongest solves fewer than half of the tasks. Coding agents and VLA policies fail in different places. Coding agents often reason out a route through the constraint and lose it in contact-rich execution, while the zero-shot VLA policies reach the task object in $31\%$ to $61\%$ of the episodes but solve none of the puzzles. A reasoning agent that plans the route, paired with a visuomotor policy that executes it in closed loop, is a natural combination to test on \name. The same scenes, graders and interfaces are open to any policy. We will release all tasks, graders, reference solutions and agent runs.

\pdfbookmark[0]{References}{arc-references}
{\small
\bibliographystyle{plainnat}
\bibliography{arxiv/main}
}

\clearpage
\appendix
\pdfbookmark[0]{Appendix}{arc-appendix}
\pdfbookmark[1]{Contents}{arc-appendix-contents}
\arcappendixcontents
\clearpage
\renewcommand{\topfraction}{0.9}\renewcommand{\bottomfraction}{0.7}
\renewcommand{\textfraction}{0.08}\renewcommand{\floatpagefraction}{0.75}
\setcounter{topnumber}{3}\setcounter{bottomnumber}{2}\setcounter{totalnumber}{5}

\section{Comparison with related benchmarks}
\label{sec:suppl_compare}
\Cref{tab:compare,tab:compare_full} mark a property when at least some tasks of a benchmark have it, as its paper or official documentation describes them. The
columns mean the following.
\begin{table}[!htbp]
\centering
\caption{\textbf{Extended comparison with related benchmarks.} The columns of \cref{tab:compare} and four more, for
$24$ benchmarks. A check mark means that at least some tasks have the property, as the benchmark's paper or
official documentation describes them. \pmark: part of the score comes from a VLM rubric (Physics), or a single task has the property (Topological). \na: does not apply, or no
fixed task set (ManiSkill3). \#Tasks as each paper counts them. \textsuperscript{a}A real-world benchmark, where
Physics and Paired regimes refer to its simulated counterpart. All definitions: \cref{sec:suppl_compare}.}
\label{tab:compare_full}
\scriptsize
\setlength{\tabcolsep}{2pt}
\renewcommand{\arraystretch}{1.08}
\begin{fittab}\begin{tabular}{@{}l r *{13}{c}@{}}
\toprule
 &  & \multicolumn{4}{c}{Tasks} & \multicolumn{3}{c}{Grading} & \multicolumn{4}{c}{Interfaces} & \multicolumn{2}{c}{Evaluation} \\
\cmidrule(lr){3-6} \cmidrule(lr){7-9} \cmidrule(lr){10-13} \cmidrule(lr){14-15}
Benchmark & \#Tasks & \shortstack{Long-\\horizon} & \shortstack{Deform-\\able} & \shortstack{Topo-\\logical} & \shortstack{Bi-\\manual} & \shortstack{Physics} & \shortstack{Verified\\solution} & \shortstack{Exact\\replay} & \shortstack{Coding\\agent} & \shortstack{State\\hidden} & \shortstack{Learned\\policy} & \shortstack{Paired\\regimes} & \shortstack{Zero-\\shot} & \shortstack{Real\\robot} \\
\midrule
\multicolumn{15}{@{}l}{\rule{0pt}{2.1ex}\textsc{\scriptsize robot manipulation}} \\
RLBench~\cite{james2020rlbench} & $100$ & \cmark & \xmark & \xmark & \xmark & \cmark & \cmark & \xmark & \xmark & \na & \cmark & \cmark & \xmark & \xmark \\
CALVIN~\cite{mees2022calvin} & $34$ & \cmark & \xmark & \xmark & \xmark & \cmark & \cmark & \xmark & \xmark & \na & \cmark & \cmark & \xmark & \xmark \\
LIBERO~\cite{liu2023libero} & $130$ & \cmark & \xmark & \xmark & \xmark & \cmark & \cmark & \xmark & \xmark & \na & \cmark & \xmark & \xmark & \xmark \\
ManiSkill3~\cite{tao2024maniskill3} & \na & \cmark & \cmark & \xmark & \cmark & \cmark & \xmark & \xmark & \xmark & \na & \cmark & \cmark & \cmark & \cmark \\
RoboCasa~\cite{nasiriany2024robocasa} & $100$ & \cmark & \xmark & \xmark & \xmark & \cmark & \cmark & \xmark & \xmark & \na & \cmark & \xmark & \xmark & \cmark \\
BEHAVIOR-1K~\cite{li2024behavior1k} & $1{,}000$ & \cmark & \cmark & \xmark & \cmark & \cmark & \xmark & \xmark & \xmark & \na & \cmark & \xmark & \xmark & \cmark \\
VLABench~\cite{zhang2024vlabench} & $100$ & \cmark & \xmark & \xmark & \xmark & \cmark & \xmark & \xmark & \cmark & \xmark & \cmark & \xmark & \cmark & \xmark \\
RoboTwin 2.0~\cite{chen2026robotwin2} & $50$ & \cmark & \xmark & \xmark & \cmark & \cmark & \cmark & \xmark & \cmark & \xmark & \cmark & \xmark & \cmark & \cmark \\
KinDER~\cite{huang2026kinder} & $25$ & \cmark & \xmark & \xmark & \xmark & \cmark & \xmark & \xmark & \xmark & \na & \cmark & \cmark & \cmark & \cmark \\
RoboWits~\cite{lin2026robowits} & $200$ & \cmark & \cmark & \xmark & \cmark & \cmark & \xmark & \xmark & \xmark & \na & \cmark & \cmark & \cmark & \cmark \\
\addlinespace[2pt]
\multicolumn{15}{@{}l}{\rule{0pt}{2.1ex}\textsc{\scriptsize deformable objects and assembly}} \\
SoftGym~\cite{lin2020softgym} & $10$ & \xmark & \cmark & \xmark & \cmark & \cmark & \xmark & \xmark & \xmark & \na & \cmark & \cmark & \xmark & \cmark \\
Deformable Ravens~\cite{seita2021deformableravens} & $12$ & \cmark & \cmark & \cmark & \xmark & \cmark & \cmark & \xmark & \xmark & \na & \cmark & \cmark & \xmark & \cmark \\
FurnitureBench\textsuperscript{a}~\cite{heo2023furniturebench} & $8$ & \cmark & \xmark & \xmark & \xmark & \cmark & \cmark & \xmark & \xmark & \na & \cmark & \cmark & \xmark & \cmark \\
AutoMate~\cite{tang2024automate} & $100$ & \xmark & \xmark & \xmark & \xmark & \cmark & \xmark & \xmark & \xmark & \na & \cmark & \xmark & \xmark & \cmark \\
\addlinespace[2pt]
\multicolumn{15}{@{}l}{\rule{0pt}{2.1ex}\textsc{\scriptsize coding and multimodal agents on robots}} \\
EmbodiedBench~\cite{yang2025embodiedbench} & $1{,}128$ & \cmark & \xmark & \xmark & \xmark & \cmark & \xmark & \xmark & \xmark & \na & \xmark & \xmark & \cmark & \xmark \\
CaP-X~\cite{fu2026capx} & $187$ & \cmark & \xmark & \xmark & \cmark & \cmark & \xmark & \xmark & \cmark & \cmark & \cmark & \cmark & \cmark & \cmark \\
EmbodiedSWE~\cite{shen2026embodiedswe} & $28$ & \cmark & \cmark & \pmark & \cmark & \pmark & \xmark & \xmark & \cmark & \xmark & \cmark & \cmark & \cmark & \cmark \\
Merler et al.~\cite{merler2026coding} & $28$ & \cmark & \xmark & \xmark & \xmark & \cmark & \xmark & \xmark & \cmark & \xmark & \xmark & \xmark & \cmark & \xmark \\
RLE-Bench~\cite{ma2026rlebench} & $51$ & \cmark & \xmark & \xmark & \cmark & \cmark & \xmark & \xmark & \cmark & \cmark & \cmark & \cmark & \cmark & \xmark \\
RobotWorld~\cite{yang2026robotworld} & $84$ & \cmark & \xmark & \xmark & \cmark & \cmark & \xmark & \xmark & \cmark & \cmark & \xmark & \xmark & \cmark & \xmark \\
BiGym 2.0~\cite{zhang2026bigym2} & $20$ & \cmark & \xmark & \xmark & \cmark & \cmark & \cmark & \cmark & \cmark & \cmark & \cmark & \xmark & \xmark & \xmark \\
\addlinespace[2pt]
\multicolumn{15}{@{}l}{\rule{0pt}{2.1ex}\textsc{\scriptsize computer-use agents}} \\
SWE-bench~\cite{jimenez2024swebench} & $2{,}294$ & \xmark & \na & \na & \na & \na & \cmark & \na & \cmark & \na & \na & \na & \cmark & \na \\
Terminal-Bench~\cite{terminalbench} & $89$ & \cmark & \na & \na & \na & \na & \cmark & \na & \cmark & \na & \na & \na & \cmark & \na \\
OSWorld~\cite{xie2024osworld} & $369$ & \cmark & \na & \na & \na & \na & \xmark & \na & \cmark & \na & \na & \na & \cmark & \na \\
\midrule
\textbf{\name (ours)} & $90$ & \cmark & \cmark & \cmark & \cmark & \cmark & \cmark & \cmark & \cmark & \cmark & \cmark & \cmark & \cmark & \cmark \\
\bottomrule
\end{tabular}\end{fittab}
\end{table}

\begin{itemize}[leftmargin=1.2em,itemsep=1pt,topsep=2pt]
  \item \emph{\#Tasks}: tasks as the benchmark's paper counts them. Each \name task comes in two regimes and counts once. ManiSkill3 is a
  framework without a fixed task count.
  \item \emph{Long-horizon}: a task needs several dependent sub-goals in a row, not one pick, push or placement.
  \item \emph{Deformable}: a task moves cloth, rope, cable or another deformable body.
  \item \emph{Topological}: success depends on how parts are linked, threaded, knotted or enclosed, not only on where they are.
  A threaded rope, a freed ring or an untied knot counts, a stacked or inserted part does not. \pmark: a single task.

  \item \emph{Physics}: a program computes success from the simulated state. \pmark: part of the score comes from a VLM rubric.
  \item \emph{Verified solution}: every task ships a solution that passes the task's own success check, so every task is
  known to be solvable.
  \item \emph{Exact replay}: the benchmark states that its simulation is deterministic, so that the same actions from
  the same start reproduce the same states.
  \item \emph{Coding agent}: the benchmark evaluates agents that write and run programs against the task.
  \item \emph{State hidden}: the coding agent is never given the simulator's state (object poses, shapes) and works
  from its cameras and proprioception.
  \item \emph{Learned policy}: the benchmark evaluates trained policies, such as VLA, imitation or reinforcement-learning policies.
  \item \na: the property does not apply, for example a robot property for a software benchmark, or State hidden for a
  benchmark without coding agents.
\end{itemize}
\paragraph{Extended comparison.}
\Cref{tab:compare_full} extends \cref{tab:compare} to $24$ benchmarks. It adds the manipulation benchmarks on which
learned policies are most often evaluated, benchmarks for deformable objects and assembly, and the benchmarks that put
coding and multimodal agents on robots. It keeps the columns of \cref{tab:compare} and adds four columns, which mean the
following.
\begin{itemize}[leftmargin=1.2em,itemsep=1pt,topsep=2pt]
  \item \emph{Bimanual}: a task is done with two arms.
  \item \emph{Paired regimes}: the benchmark offers the same tasks, under the same success rule, with the simulator's
  state as input and with sensor inputs only.
  \item \emph{Zero-shot}: the benchmark evaluates pretrained models or agents as released, with no demonstration of
  its tasks and no training on them.
  \item \emph{Real robot}: the benchmark's paper reports an experiment on a physical robot with its tasks, its
  policies or its interface.
\end{itemize}

\section{The task suite}
\label{sec:suppl_tasksuite}

\subsection{Task gallery}
\label{sec:suppl_gallery}

\Cref{fig:gallery_all} shows every task of \name in five frames of its reference solution, from the start through three intermediate steps to the solved state, one row per task and grouped by constraint type.

\providecommand{\galpart}[3]{\hbox{\includegraphics[width=\linewidth,trim={0 #2bp 0 #3bp},clip]{arxiv/figures/fig_gallery_p#1.pdf}}}
\providecommand{\galstack}[1]{\noindent\vbox{\offinterlineskip #1}\par}
\renewcommand{\floatpagefraction}{0.7}

\begin{figure}[p]
  \centering
  \galstack{\galpart{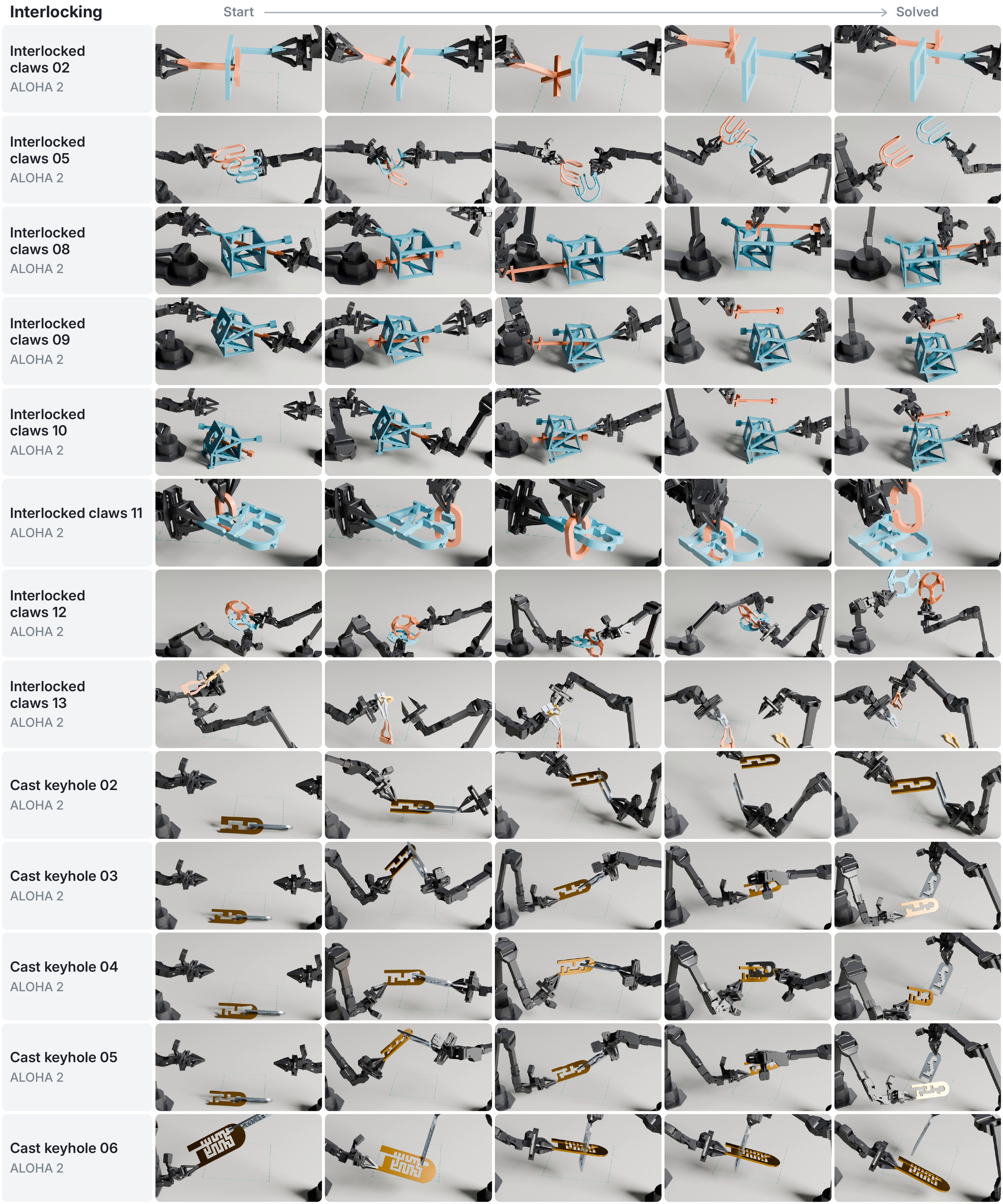}{0}{0}}
  \caption{\textbf{Every task of \name, from start to solved.} Five frames of each task's reference solution, grouped by constraint type.}
  \label{fig:gallery_all}
\end{figure}

\begin{figure}[p]
  \ContinuedFloat
  \centering
  \galstack{\galpart{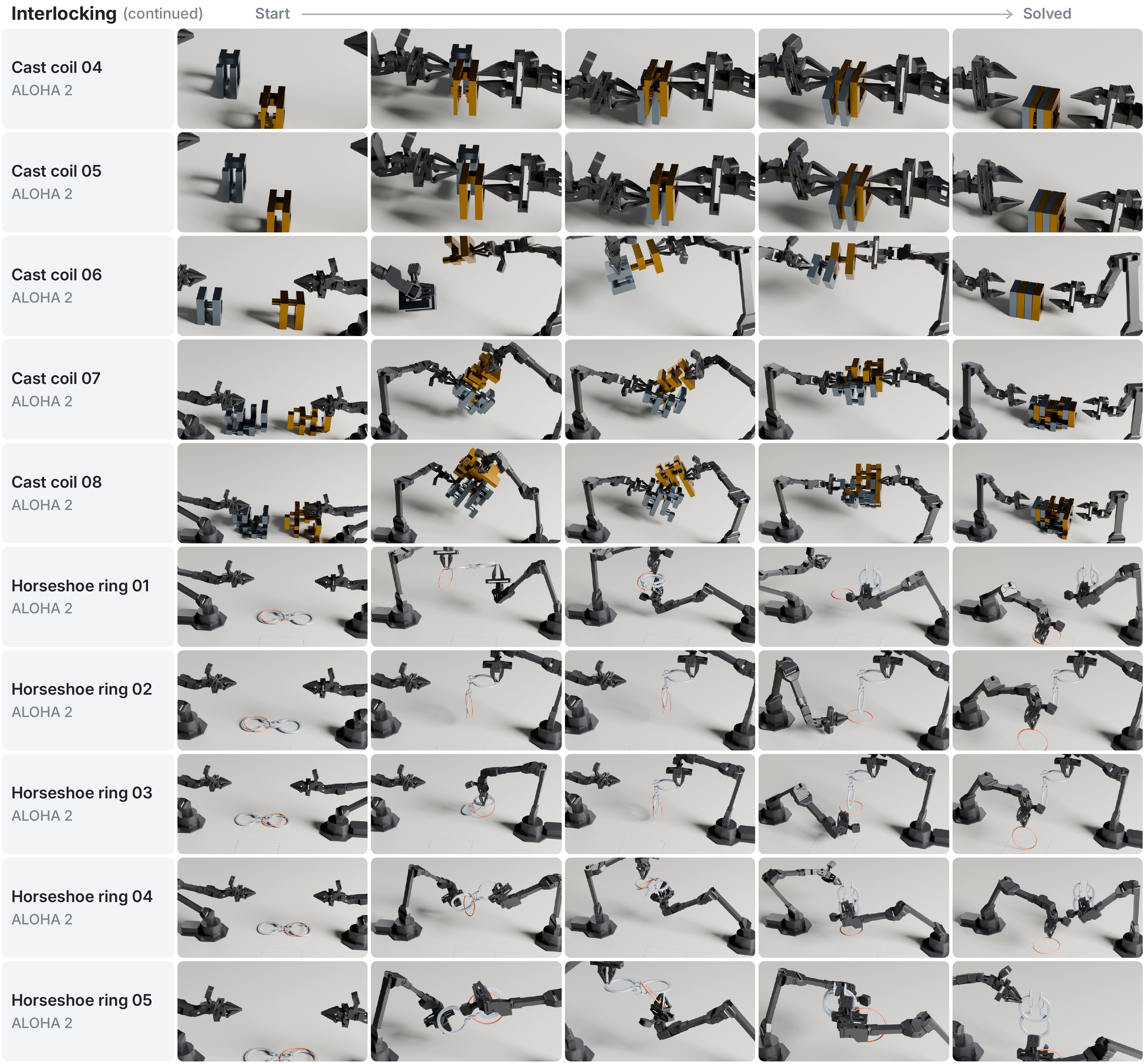}{0}{0}\vskip 8pt\galpart{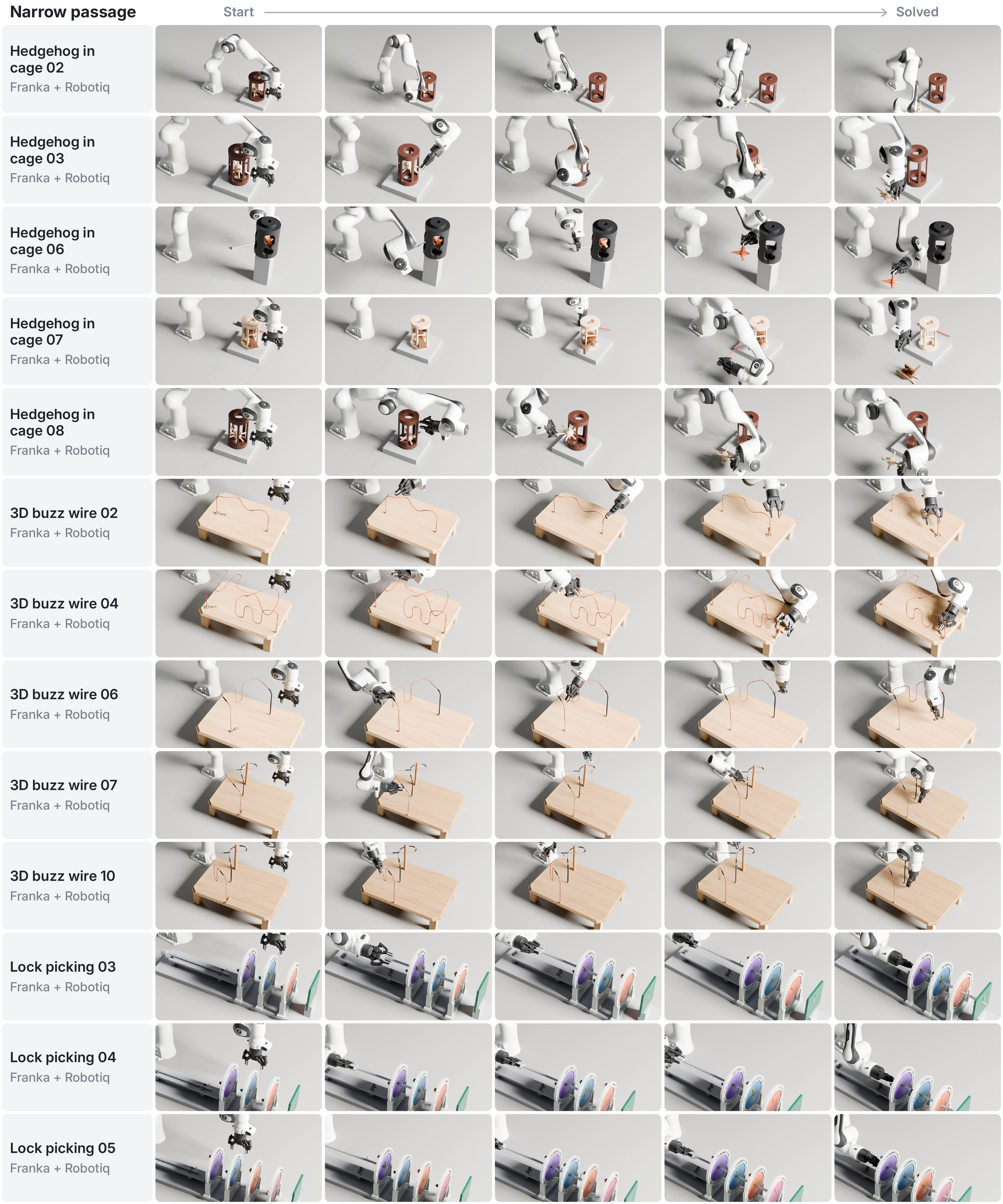}{1222.96}{0}}
  \caption[]{(continued)}
\end{figure}

\begin{figure}[p]
  \ContinuedFloat
  \centering
  \galstack{\galpart{3}{1766.96}{0}\galpart{3}{0}{578}\vskip 8pt\galpart{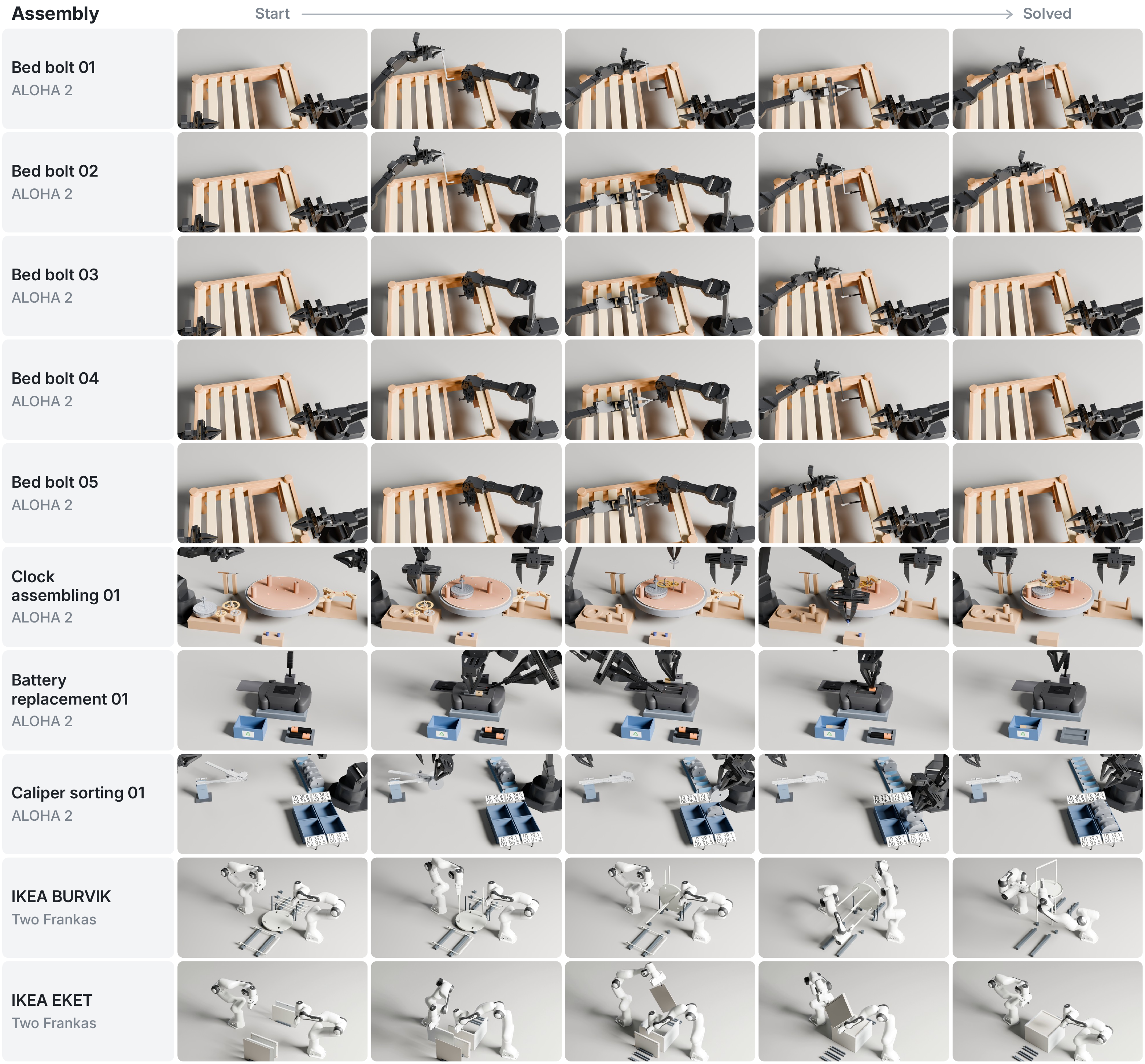}{679.92}{0}}
  \caption[]{(continued)}
\end{figure}

\begin{figure}[p]
  \ContinuedFloat
  \centering
  \galstack{\galpart{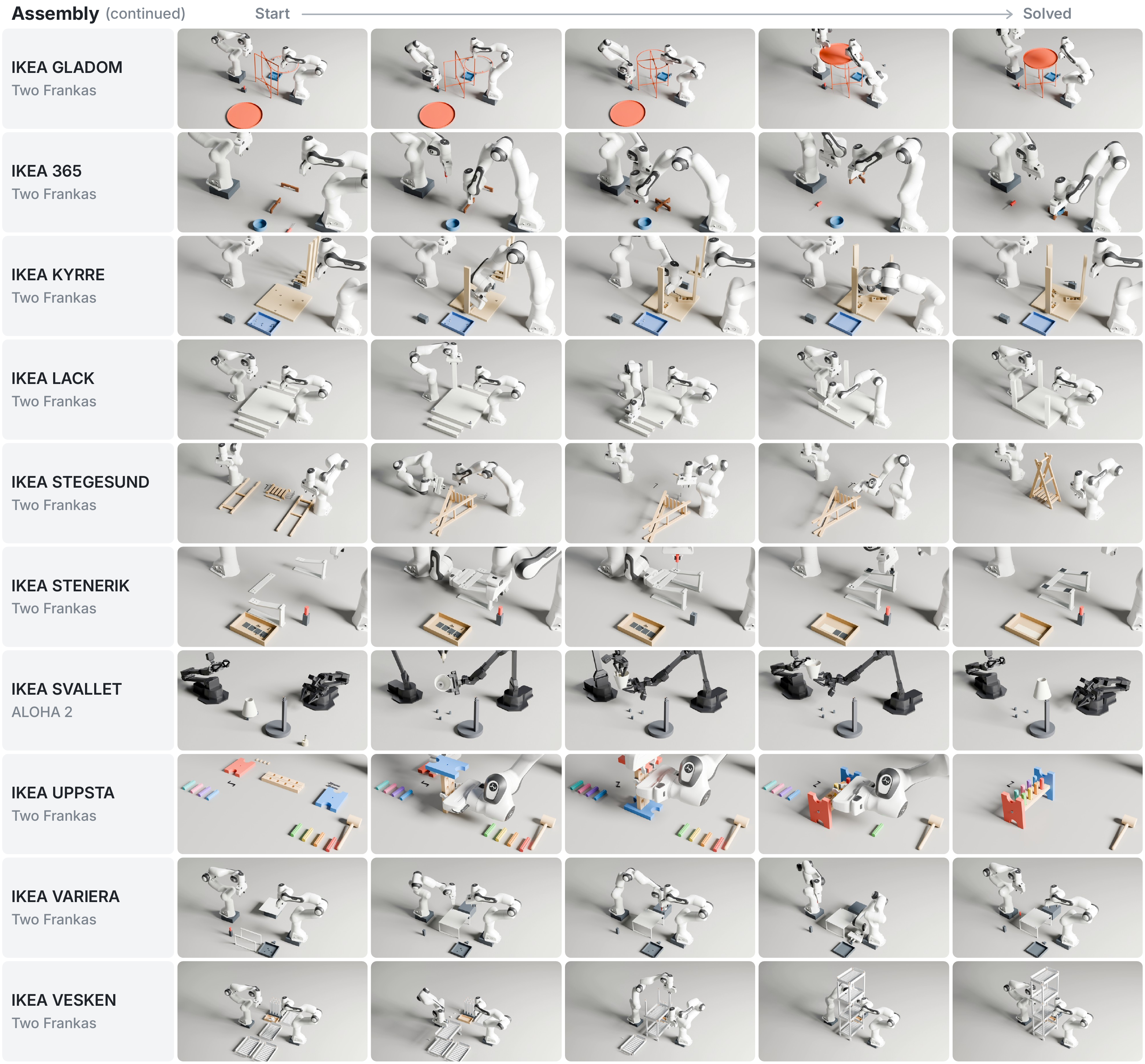}{1359.92}{0}\galpart{4}{0}{714}\galpart{5}{0}{36.4}}
  \caption[]{(continued)}
\end{figure}

\begin{figure}[p]
  \ContinuedFloat
  \centering
  \galstack{\galpart{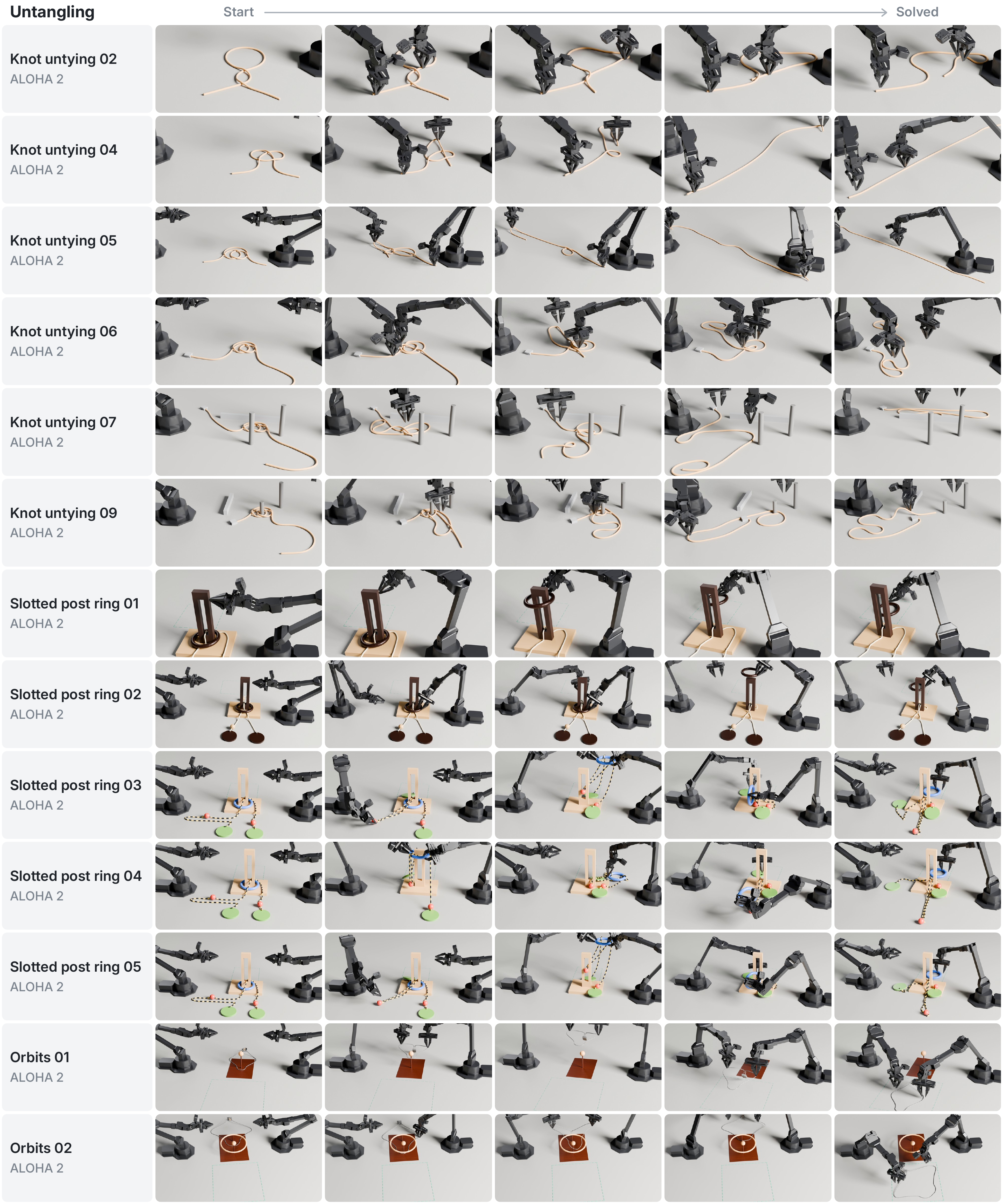}{0}{0}}
  \caption[]{(continued)}
\end{figure}

\begin{figure}[p]
  \ContinuedFloat
  \centering
  \galstack{\galpart{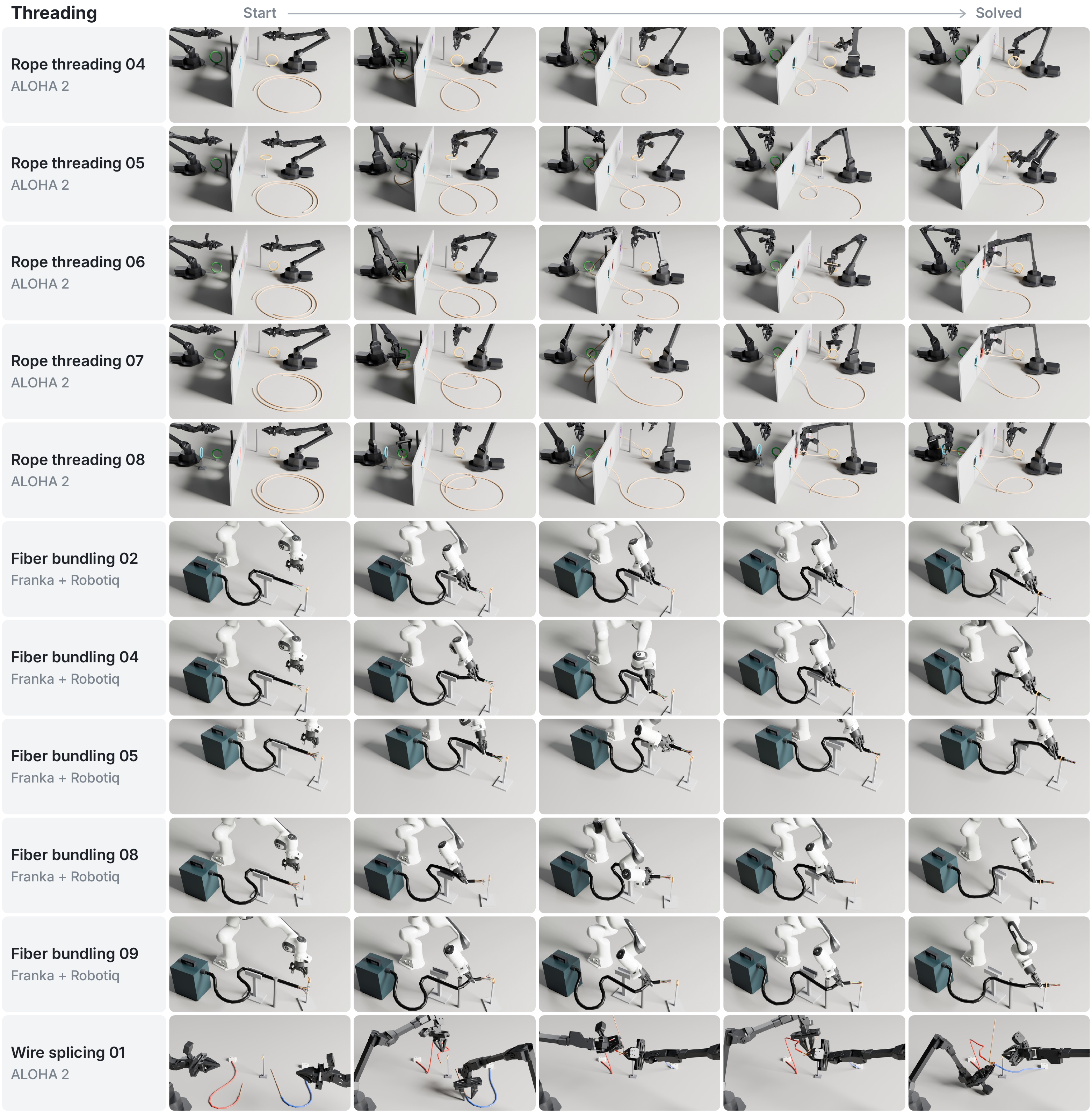}{0}{0}}
  \caption[]{(continued)}
\end{figure}

\begin{figure}[!t]
  \ContinuedFloat
  \centering
  \galstack{\galpart{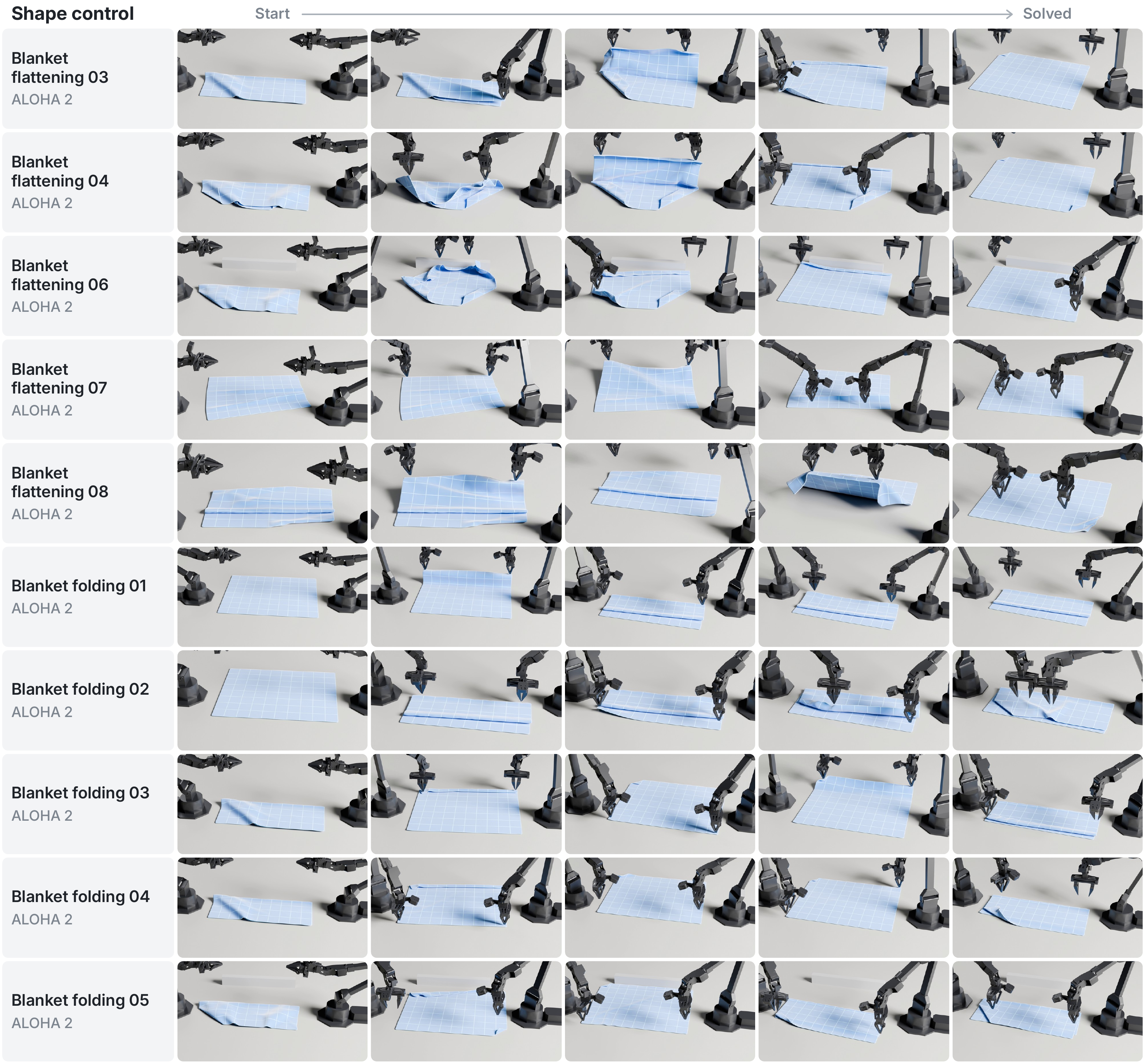}{0}{0}}
  \caption[]{(continued)}
\end{figure}

\subsection{Constraint types, scale and simulation}
\label{sec:suppl_suite}

\paragraph{Constraint types.}
We organize the tasks by the spatial constraint that decides success and by material (\cref{fig:stats}). Rigid tasks are of three types. In
\emph{interlocking}, hooked or keyed parts must be separated or assembled along a narrow release path. In
\emph{narrow passage}, a held tool or part is guided through millimetre clearances, as in a 3D buzz wire, a
hedgehog in a cage or a key through rotor plates. \emph{Assembly} ranges from driving a bolt with a key and
seating the gear train of a watch movement to assembling IKEA furniture.

Soft tasks are \emph{untangling} (untying knots and
freeing rings and loops caught on cords), \emph{threading} (passing a rope through holes in a given order and
direction, bundling fibers through a ring, splicing two wires) and \emph{shape control} (unfolding, flattening and
folding a blanket). Success is defined by the final state, often by its topology (is the ring free, does
the rope cross itself, did every fiber pass the ring). A task tests its constraint type: the same rope is threaded in
one family and untied in another.
\paragraph{Scale and robots.}
\name has $90$ tasks: $78$ puzzles in $19$ families and $12$ IKEA furniture scenes. A family groups tasks of one design at several difficulty
ratings. Puzzles run on ALOHA\,2~\cite{aldaco2024aloha2}, two $6$-DoF arms with parallel grippers ($14$-D joint targets at
$40$\,Hz, $25$\,Hz for rope threading), or on a $7$-DoF Franka Panda with a Robotiq 2F-85 gripper ($8$-D joint targets at $40$\,Hz). IKEA scenes are assembled by two Franka Panda arms (one scene by ALOHA\,2) at a $400$\,Hz control tick with
physical grasps.

Rigid puzzle parts use assisted grasping: a closing gripper whose fingers both touch a part inside its
declared grasp region holds the part rigidly until it opens, or until the hold would need more than $60$--$65$\,N. Cloth is held by local patches, and some knot tasks allow physical grasping only. The instruction states these rules, the
success rule and the time one control step takes. For each region, the \std instruction gives the radius but not the centre. The \priv regime and the part-geometry arm of the ablation (\cref{sec:suppl_ablation}) give the regions.

\paragraph{Simulation.}
All tasks run on MuJoCo~3.14~\cite{todorov2012mujoco} on the CPU in double precision, and the cloth of the blanket families
runs on a deterministic CUDA step. Ropes and cords are chains of capsule links, and the spliced wires chains of
elastoplastic ball joints, so that a bend stays bent. Images are path traced by the Nyx renderer~\cite{nyx}. Each puzzle task
has three domain-randomization seeds, the same scene and physics in other colours and materials, and each of the three runs of
an evaluation uses one seed (\cref{sec:suppl_seeds}).

\begin{figure}[!htbp]
  \centering
  \includegraphics[width=0.6\linewidth]{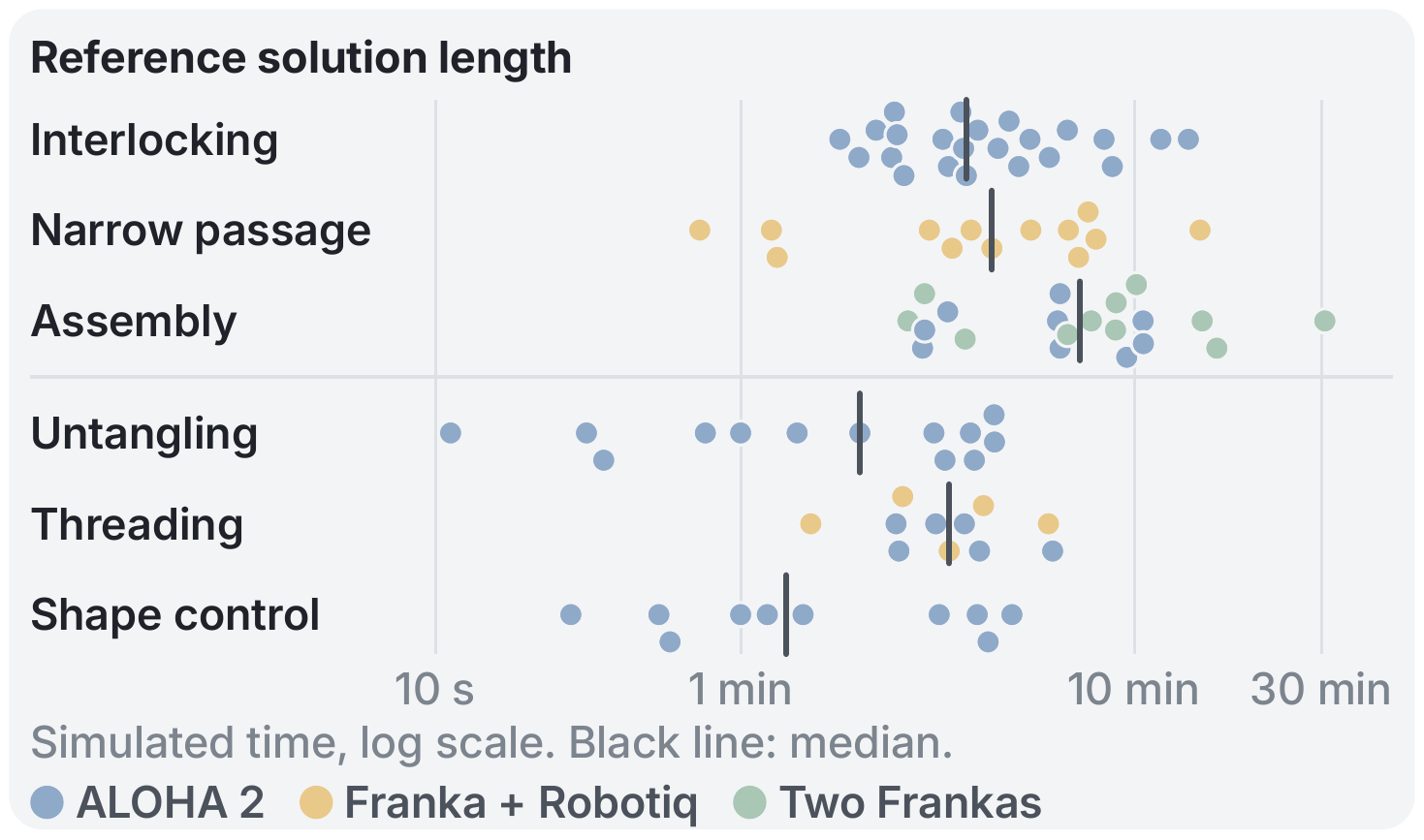}
  \caption{\textbf{Reference solution length per constraint type.} Simulated time of each task's reference solution,
one dot per task coloured by robot, with the median of each type as a black line.}
  \label{fig:reflen}
\end{figure}

\paragraph{Budget.}
An agent's session is limited to $11{,}400$\,s of wall clock per task, $190$ minutes, and to $21{,}600$\,s ($6$\,h) for
the one wire-splicing task. DeepSeek v4.1 Flash and GLM 5.3 Flash ran under $10{,}800$\,s (\cref{sec:suppl_agents}).
The reference solutions take from under a minute to $85$ minutes of wall clock (two hours for wire splicing,
\cref{tab:tasks}). \Cref{fig:reflen} gives
their simulated time, and \cref{sec:suppl_budget} the share of the trials that are solved within a shorter time.

\renewcommand{\floatpagefraction}{0.75}

\subsection{Task list}
\label{sec:suppl_tasks}

\Cref{tab:tasks} lists every task with its type, difficulty rating, and the length and wall-clock time of its
reference solution. \Cref{tab:families} describes each family and the success rule its grader checks, with the thresholds the task texts state.

\begin{table}[!t]
\centering
\caption{\textbf{The task families.} Task counts in brackets. ALOHA\,2: two $6$-DoF arms with parallel grippers. Franka: one Franka Panda with a Robotiq 2F-85 gripper. Every success rule must hold continuously for $0.5$\,s of simulated time unless its row states another time. The families of the soft constraint types follow in the second part.}
\label{tab:families}
\scriptsize
\setlength{\tabcolsep}{3pt}
\renewcommand{\arraystretch}{1.05}
\begin{fittab}\begin{tabular}{@{}>{\raggedright\arraybackslash}p{1.85cm} >{\raggedright\arraybackslash}p{1.52cm} >{\raggedright\arraybackslash}p{1.25cm} >{\raggedright\arraybackslash}p{4.0cm} >{\raggedright\arraybackslash}p{6.48cm}@{}}
\toprule
Family & Type & Robot & Task & Success rule \\
\midrule
Interlocked claws ($8$) & Interlocking & ALOHA\,2 & Free one hooked piece and place it in the goal square, or separate two or three interlocked pieces (Cast ABC, Rotor, Trinity) & Placement: every point of the piece at least $1$\,mm inside the goal square, the rest on the puzzle side of a fixed plane. Separation: the pieces at least $10$\,mm apart along one of ten directions. Pieces at most $1$\,mm/s and $0.02$\,rad/s \\
Cast keyhole ($5$) & Interlocking & ALOHA\,2 & Free the keyhole plate from the fork plate and place it in the goal square & Every point of the keyhole plate at least $1$\,mm inside the goal square, the fork plate on the puzzle side, the plate at rest \\
Cast coil ($5$) & Interlocking & ALOHA\,2 & Assemble two pieces into their interlocking configuration and release both & Assembly frames within $1$\,mm and $0.02$\,rad, one piece within $3$\,mm of the tabletop, both released with $5$\,mm clearance and at rest \\
Horseshoe ring ($5$) & Interlocking & ALOHA\,2 & Free a ring caught on two horseshoes joined by chain links and place it in the goal square & The ring at least $1$\,mm inside the goal square, horseshoes and chains on the puzzle side with all four links intact, the ring at rest. A broken link fails the episode \\
\midrule
Hedgehog in cage ($5$) & Narrow passage & Franka & Take a spiked ball, star or owl out of a fixed cage and into the goal square & Every point of the object at least $1$\,mm inside the goal square, the cage on the puzzle side, the object at rest, for $2$\,s \\
3D buzz wire ($5$) & Narrow passage & Franka & Guide a wand's loop along a bent wire from start to finish, then set the wand down & One continuous pass from within $10$\,mm of the start to within $10$\,mm of the finish with the wire inside the loop, the wand on its support within $5$\,mm and $10^\circ$, released, for $1$\,s. Any metal contact with the wire fails \\
Lock picking ($3$) & Narrow passage & Franka & Insert a two-tab key through a row of rotor plates and an end plate & The shaft through every plate, the leading tab at least $1$\,mm past the end plate, the tabs never in solid plate, every rotor within its slot ($0.5^\circ$) \\
\midrule
Bed bolt ($5$) & Assembly & ALOHA\,2 & Insert a bolt into a threaded hole and drive it home with an L-shaped key (hex, Torx, square, Phillips or slotted) & The head within $1.5$\,mm of the block, the tip at least $83.5$\,mm deep with the thread caught, the axis within $2^\circ$, bolt and key at rest \\
Clock assembling ($1$) & Assembly & ALOHA\,2 & Assemble a watch gear train at $8\times$ scale, screw down the bridge, wind the mainspring & Four wheels seated and meshed, the bridge seated and its two screws turned $2$ turns, the barrel wound at least $120^\circ$, then the fourth wheel turns at least $3.5$ times in an untouched $5$\,s run turning $450$ times as far as the barrel (within $2\,\%$) \\
Battery replacement ($1$) & Assembly & ALOHA\,2 & Open a controller's cover, pry out two old AA cells into a recycling box, seat two new cells, latch the cover & New cells seated (axis within $5^\circ$, spring compressed at least $1$\,mm) with the right polarity, the cover latched within $0.5$\,mm and $1^\circ$, both old cells in the box, all at rest \\
Caliper sorting ($1$) & Assembly & ALOHA\,2 & Read four look-alike sleeves on a vernier caliper and sort each into the bin labelled with its readings & Each sleeve within $40$\,mm of the centre of the bin whose readings match its own (within $0.1$), released, at rest. A sleeve left in a wrong bin fails \\
IKEA furniture ($12$) & Assembly & $2$ Franka\textsuperscript{a} & Assemble a piece of IKEA furniture with physical grasps (screws, bolts, dowels, snaps) & The scene's own assembly check, every part released and at rest ($0.01$\,m/s, $0.05$\,rad/s), at the last row and again after a $1$\,s hold \\
\bottomrule
\end{tabular}\end{fittab}

\vspace{2pt}
{\scriptsize \textsuperscript{a}Two Franka Panda arms with Franka Hands, ALOHA\,2 in one scene (SVALLET).}
\end{table}

\begin{table}[!t]
\ContinuedFloat
\centering
\caption[]{\textbf{The task families} (continued).}
\scriptsize
\setlength{\tabcolsep}{3pt}
\renewcommand{\arraystretch}{1.05}
\begin{fittab}\begin{tabular}{@{}>{\raggedright\arraybackslash}p{1.85cm} >{\raggedright\arraybackslash}p{1.52cm} >{\raggedright\arraybackslash}p{1.25cm} >{\raggedright\arraybackslash}p{4.0cm} >{\raggedright\arraybackslash}p{6.48cm}@{}}
\toprule
Family & Type & Robot & Task & Success rule \\
\midrule
Knot untying ($6$) & Untangling & ALOHA\,2 & Untie a knotted rope and lay it out flat, in two tasks inside a goal region & No crossing or overlap in the top view, non-neighbouring segments at least one rope diameter apart, the rope on the table, released with $1$\,mm clearance, at rest \\
Slotted post ring ($5$) & Untangling & ALOHA\,2 & Free a ring caught on a cord through a slotted post and place it in the goal square & The ring at least $1$\,mm inside the goal square, post, cord, beads and discs on the puzzle side, the ring at rest \\
Orbits ($2$) & Untangling & ALOHA\,2 & Free a cord loop on a handle from a fixed post assembly and place it in the goal square & Cord and handle at least $1$\,mm inside the goal square, the assembly on the puzzle side, the loop's tie within $3$\,mm, at most $10$\,mm/s \\
\midrule
Rope threading ($5$) & Threading & ALOHA\,2 & Thread a rope through named holes in a given order and direction & Each required hole passed exactly once in order and direction, the full rope thickness clearing the rim by $0.1$\,mm, the rope intact and at most $10$\,mm/s \\
Fiber bundling ($5$) & Threading & Franka & Pass every fine fiber, then the sheath, through a fixed ring & Every fiber through tip first to its root and beyond the ring, the sheath's branching face through, no kink above $60^\circ$. A gripper touch of a fiber fails \\
Wire splicing ($1$) & Threading & ALOHA\,2 & Bend an eye in one wire, thread the other through it, wind a coil and fold the end back & An eye turned more than $150^\circ$ and closed, the other wire through it once and wound at least $1.5$ turns, the coil tight, the end folded back, released and still, for $2$\,s \\
\midrule
Blanket flatten ($5$) & Shape control & ALOHA\,2 & Unfold a folded blanket and spread it flat, then let go & At least $95\%$ coverage, every point at most $30$\,mm above the table, no edge stretched beyond $1.15\times$, released with $5$\,mm clearance, at rest \\
Blanket fold ($5$) & Shape control & ALOHA\,2 & Fold a blanket into a target half fold, in some tasks with a dog-ear, at a given place & Corners within $55$\,mm, the layer order of the target on at least $88\%$ of points, outline IoU at least $0.85$, placed within $40$\,mm and $5^\circ$, released and at rest \\
\bottomrule
\end{tabular}\end{fittab}
\end{table}

\begin{table}[!htbp]
\centering
\caption{\textbf{Every task of \name.} Type as in \cref{fig:stats}. Difficulty: the task's rating within its
family (IKEA scenes are unrated). Steps and Wall: length of the reference solution in control steps and its wall-clock
time in minutes.}
\label{tab:tasks}
\scriptsize
\setlength{\tabcolsep}{2.5pt}
\begin{fittab}\begin{tabular}{@{}l l l r r @{\hspace{9pt}} l l l r r@{}}
\toprule
Task & Type & Difficulty & Steps & Wall & Task & Type & Difficulty & Steps & Wall \\
\midrule
\texttt{interlocked-claws-02} & Interlocking & easier & $20.1$k & $20$ & \texttt{ikea-eket} & Assembly & -- & $70.2$k & $5$ \\
\texttt{interlocked-claws-05} & Interlocking & hard & $16.2$k & $22$ & \texttt{ikea-gladom} & Assembly & -- & $186.6$k & $85$ \\
\texttt{interlocked-claws-08} & Interlocking & easy & $4.3$k & $5$ & \texttt{ikea-ikea365} & Assembly & -- & $89.1$k & $6$ \\
\texttt{interlocked-claws-09} & Interlocking & medium & $4.8$k & $5$ & \texttt{ikea-kyrre} & Assembly & -- & $733.6$k & $46$ \\
\texttt{interlocked-claws-10} & Interlocking & hard & $6.2$k & $9$ & \texttt{ikea-lack} & Assembly & -- & $63.7$k & $9$ \\
\texttt{interlocked-claws-11} & Interlocking & medium & $6.0$k & $5$ & \texttt{ikea-stegesund} & Assembly & -- & $357.2$k & $25$ \\
\texttt{interlocked-claws-12} & Interlocking & hard & $21.1$k & $53$ & \texttt{ikea-stenerik} & Assembly & -- & $215.8$k & $84$ \\
\texttt{interlocked-claws-13} & Interlocking & hard & $9.6$k & $12$ & \texttt{ikea-svallet} & Assembly & -- & $69.4$k & $5$ \\
\texttt{cast-keyhole-02} & Interlocking & easier & $5.9$k & $4$ & \texttt{ikea-uppsta} & Assembly & -- & $430.2$k & $64$ \\
\texttt{cast-keyhole-03} & Interlocking & easy & $12.2$k & $9$ & \texttt{ikea-variera} & Assembly & -- & $243.2$k & $62$ \\
\texttt{cast-keyhole-04} & Interlocking & medium & $10.8$k & $8$ & \texttt{ikea-vesken} & Assembly & -- & $389.5$k & $13$ \\
\texttt{cast-keyhole-05} & Interlocking & hard & $14.6$k & $12$ & \texttt{knot-untying-02} & Untangling & easier & $1.1$k & $21$ \\
\texttt{cast-keyhole-06} & Interlocking & hard & $8.1$k & $4$ & \texttt{knot-untying-04} & Untangling & medium & $1.9$k & $36$ \\
\texttt{cast-coil-04} & Interlocking & medium & $9.0$k & $7$ & \texttt{knot-untying-05} & Untangling & hard & $2.4$k & $52$ \\
\texttt{cast-coil-05} & Interlocking & hard & $8.8$k & $7$ & \texttt{knot-untying-06} & Untangling & very easy & $4.8$k & $36$ \\
\texttt{cast-coil-06} & Interlocking & very easy & $8.7$k & $7$ & \texttt{knot-untying-07} & Untangling & easier & $7.4$k & $51$ \\
\texttt{cast-coil-07} & Interlocking & easier & $28.0$k & $18$ & \texttt{knot-untying-09} & Untangling & medium & $10.6$k & $56$ \\
\texttt{cast-coil-08} & Interlocking & easy & $33.0$k & $24$ & \texttt{slotted-post-ring-01} & Untangling & very easy & $0.4$k & $0.5$ \\
\texttt{horseshoe-ring-01} & Interlocking & very easy & $11.5$k & $14$ & \texttt{slotted-post-ring-02} & Untangling & easier & $1.0$k & $1$ \\
\texttt{horseshoe-ring-02} & Interlocking & easier & $5.3$k & $6$ & \texttt{slotted-post-ring-03} & Untangling & easy & $9.2$k & $11$ \\
\texttt{horseshoe-ring-03} & Interlocking & easy & $5.8$k & $7$ & \texttt{slotted-post-ring-04} & Untangling & medium & $10.6$k & $13$ \\
\texttt{horseshoe-ring-04} & Interlocking & medium & $13.0$k & $15$ & \texttt{slotted-post-ring-05} & Untangling & hard & $9.4$k & $12$ \\
\texttt{horseshoe-ring-05} & Interlocking & hard & $7.8$k & $10$ & \texttt{orbits-01} & Untangling & very easy & $3.3$k & $4$ \\
\texttt{hedgehog-in-cage-02} & Narrow passage & easier & $1.9$k & $2$ & \texttt{orbits-02} & Untangling & easier & $7.9$k & $30$ \\
\texttt{hedgehog-in-cage-03} & Narrow passage & easy & $8.3$k & $7$ & \texttt{rope-threading-04} & Threading & very easy & $3.7$k & $22$ \\
\texttt{hedgehog-in-cage-06} & Narrow passage & hard & $2.9$k & $2$ & \texttt{rope-threading-05} & Threading & easier & $3.8$k & $22$ \\
\texttt{hedgehog-in-cage-07} & Narrow passage & easy & $7.2$k & $3$ & \texttt{rope-threading-06} & Threading & easy & $5.5$k & $32$ \\
\texttt{hedgehog-in-cage-08} & Narrow passage & easier & $3.0$k & $2$ & \texttt{rope-threading-07} & Threading & medium & $4.7$k & $29$ \\
\texttt{buzz-wire-3d-02} & Narrow passage & easier & $13.1$k & $14$ & \texttt{rope-threading-08} & Threading & hard & $6.1$k & $36$ \\
\texttt{buzz-wire-3d-04} & Narrow passage & medium & $9.2$k & $17$ & \texttt{fiber-bundling-02} & Threading & easier & $3.6$k & $5$ \\
\texttt{buzz-wire-3d-06} & Narrow passage & very easy & $10.4$k & $11$ & \texttt{fiber-bundling-04} & Threading & medium & $8.1$k & $22$ \\
\texttt{buzz-wire-3d-07} & Narrow passage & easier & $17.3$k & $18$ & \texttt{fiber-bundling-05} & Threading & hard & $9.9$k & $48$ \\
\texttt{buzz-wire-3d-10} & Narrow passage & hard & $18.3$k & $20$ & \texttt{fiber-bundling-08} & Threading & easy & $6.2$k & $10$ \\
\texttt{lock-picking-03} & Narrow passage & easy & $16.3$k & $24$ & \texttt{fiber-bundling-09} & Threading & medium & $14.5$k & $39$ \\
\texttt{lock-picking-04} & Narrow passage & medium & $19.2$k & $28$ & \texttt{wire-splicing-01} & Threading & hard & $14.9$k & $124$ \\
\texttt{lock-picking-05} & Narrow passage & hard & $35.3$k & $41$ & \texttt{blanket-flatten-03} & Shape control & easier & $2.4$k & $15$ \\
\texttt{bed-bolt-01} & Assembly & medium & $25.3$k & $16$ & \texttt{blanket-flatten-04} & Shape control & easy & $3.4$k & $22$ \\
\texttt{bed-bolt-02} & Assembly & hard & $25.3$k & $15$ & \texttt{blanket-flatten-06} & Shape control & hard & $11.7$k & $74$ \\
\texttt{bed-bolt-03} & Assembly & very easy & $15.5$k & $10$ & \texttt{blanket-flatten-07} & Shape control & easy & $1.6$k & $14$ \\
\texttt{bed-bolt-04} & Assembly & easier & $15.5$k & $10$ & \texttt{blanket-flatten-08} & Shape control & medium & $1.5$k & $14$ \\
\texttt{bed-bolt-05} & Assembly & easy & $15.3$k & $9$ & \texttt{blanket-fold-01} & Shape control & very easy & $0.9$k & $6$ \\
\texttt{clock-assembling-01} & Assembly & hard & $8.0$k & $11$ & \texttt{blanket-fold-02} & Shape control & easier & $2.8$k & $18$ \\
\texttt{battery-replacement-01} & Assembly & medium & $7.0$k & $7$ & \texttt{blanket-fold-03} & Shape control & easy & $9.6$k & $61$ \\
\texttt{caliper-sorting-01} & Assembly & hard & $23.0$k & $25$ & \texttt{blanket-fold-04} & Shape control & medium & $7.7$k & $48$ \\
\texttt{ikea-burvik} & Assembly & -- & $162.5$k & $29$ & \texttt{blanket-fold-05} & Shape control & hard & $10.2$k & $65$ \\
\bottomrule
\end{tabular}\end{fittab}
\end{table}

\subsection{Domain randomization}
\label{sec:suppl_seeds}
\begin{figure}[!htbp]
  \centering
  \setlength{\tabcolsep}{1.5pt}
  \renewcommand{\arraystretch}{0.6}
  \begin{tabular}{@{}cccc@{}}
   & {\small Seed $0$} & {\small Seed $1$} & {\small Seed $2$} \\
\rotatebox{90}{\parbox{0.15\linewidth}{\centering\footnotesize Interlocked claws 05}} & \includegraphics[width=0.28\linewidth]{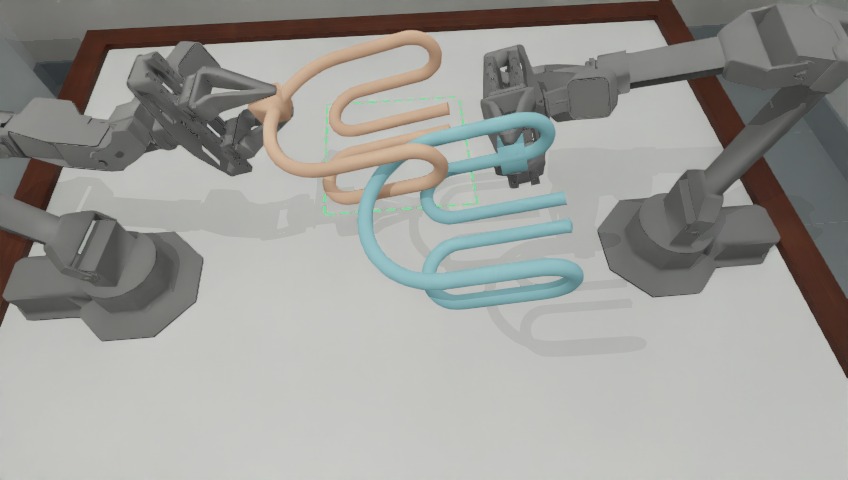} & \includegraphics[width=0.28\linewidth]{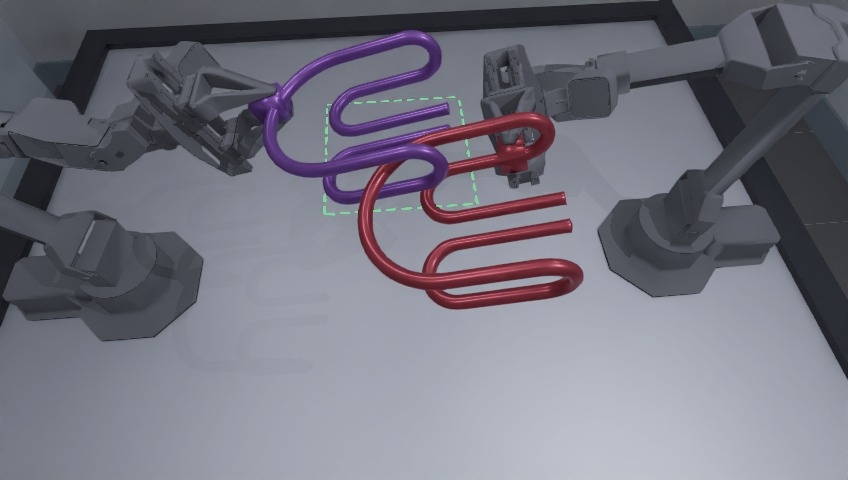} & \includegraphics[width=0.28\linewidth]{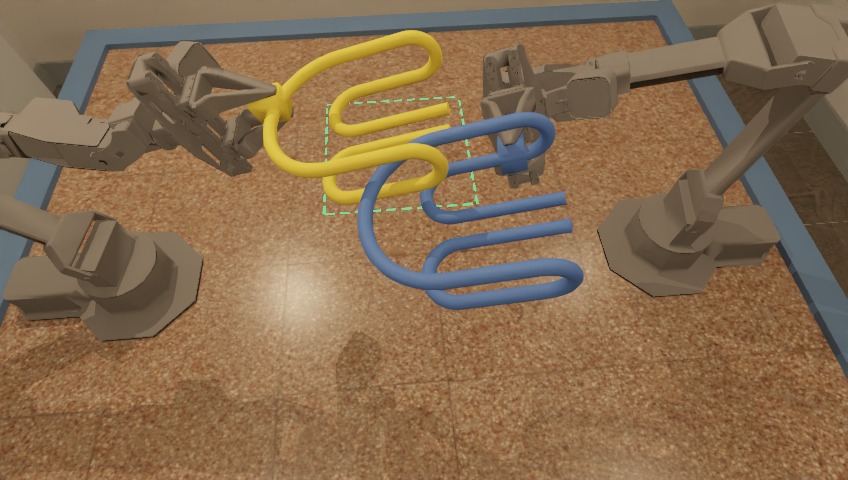} \\
\rotatebox{90}{\parbox{0.15\linewidth}{\centering\footnotesize Buzz wire 3D 06}} & \includegraphics[width=0.28\linewidth]{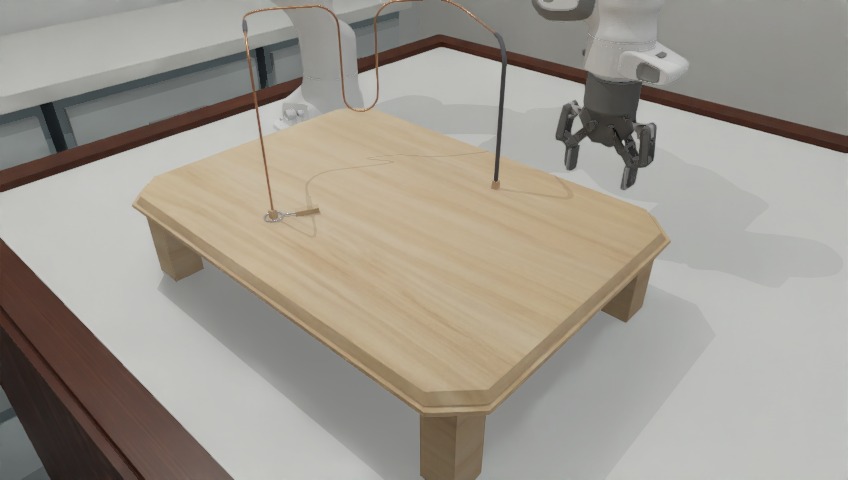} & \includegraphics[width=0.28\linewidth]{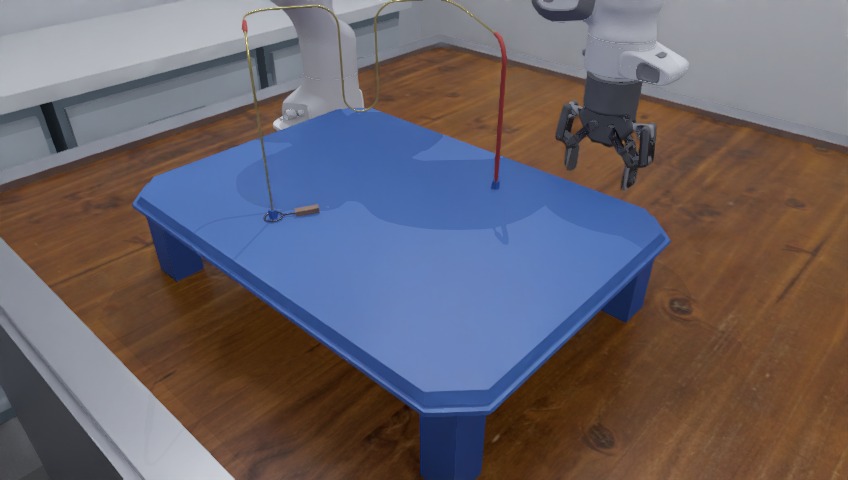} & \includegraphics[width=0.28\linewidth]{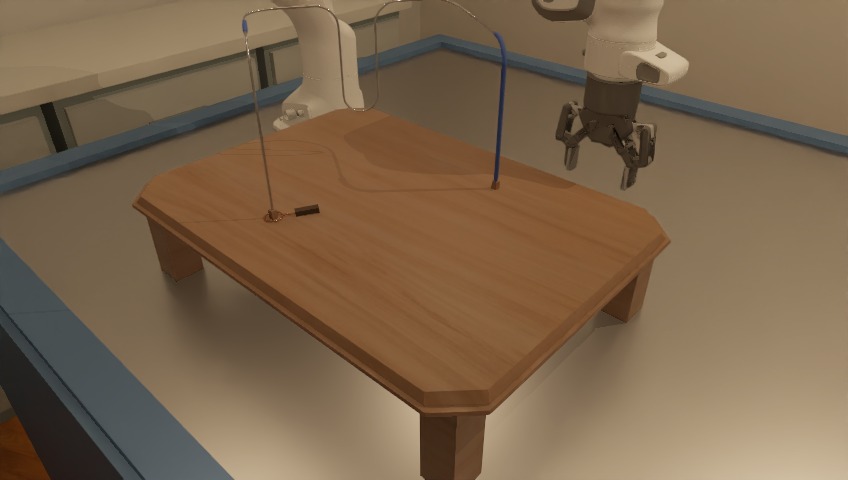} \\
\rotatebox{90}{\parbox{0.15\linewidth}{\centering\footnotesize Blanket fold 01}} & \includegraphics[width=0.28\linewidth]{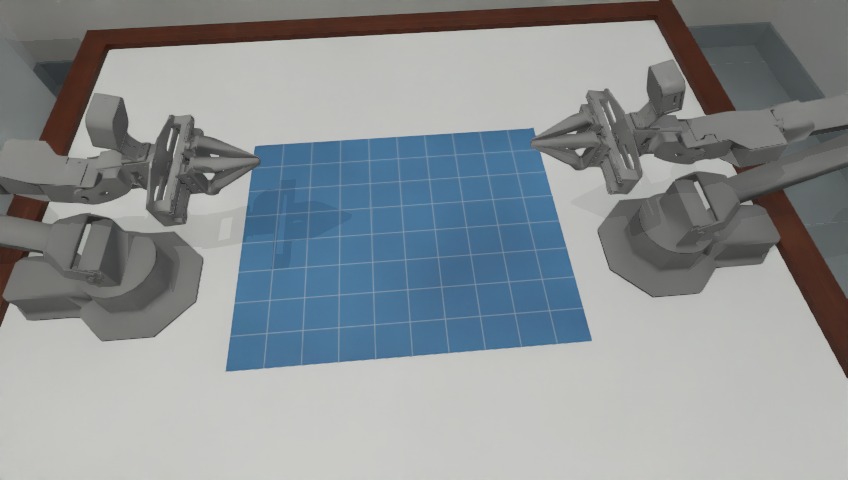} & \includegraphics[width=0.28\linewidth]{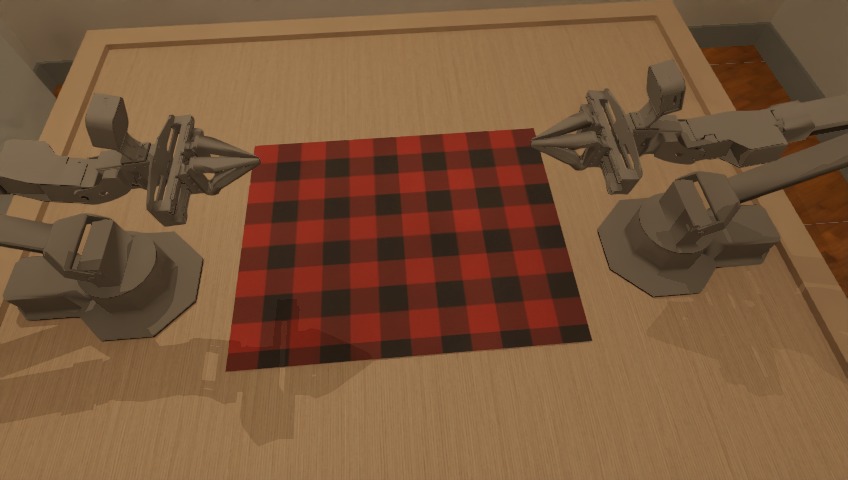} & \includegraphics[width=0.28\linewidth]{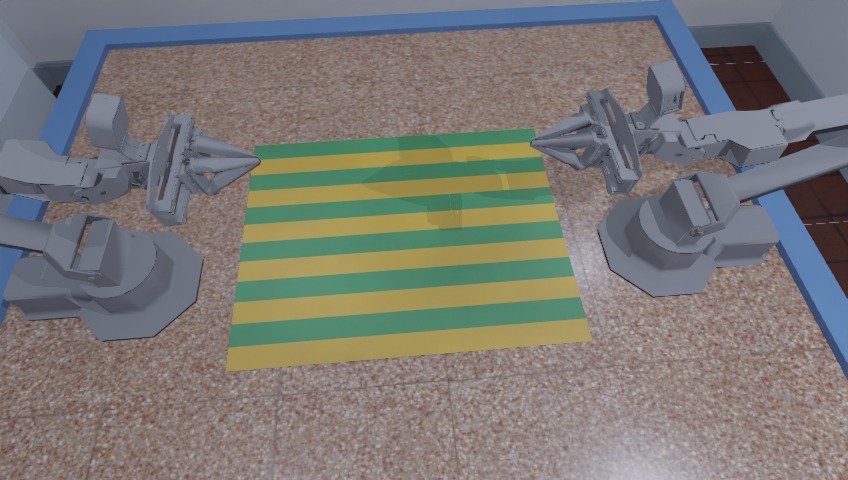} \\
\rotatebox{90}{\parbox{0.15\linewidth}{\centering\footnotesize Knot untying 07}} & \includegraphics[width=0.28\linewidth]{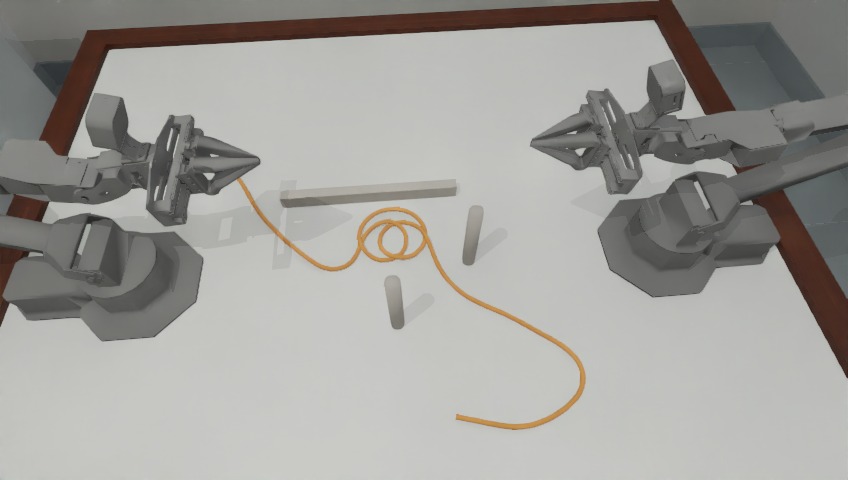} & \includegraphics[width=0.28\linewidth]{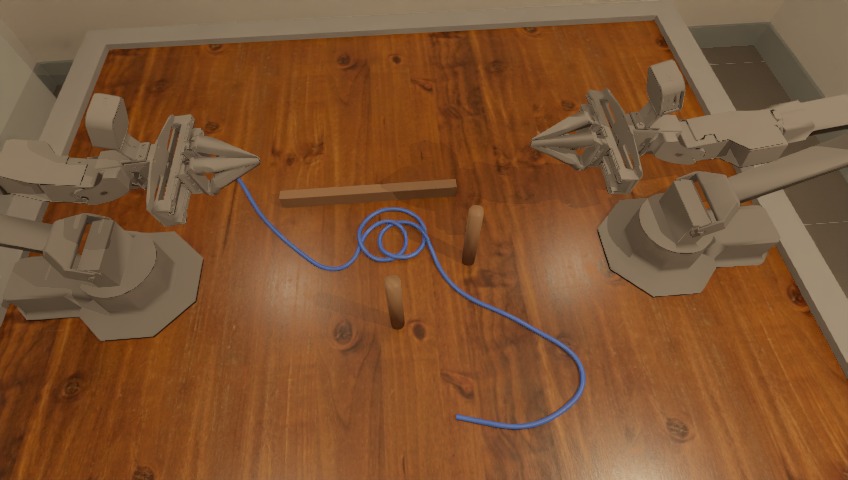} & \includegraphics[width=0.28\linewidth]{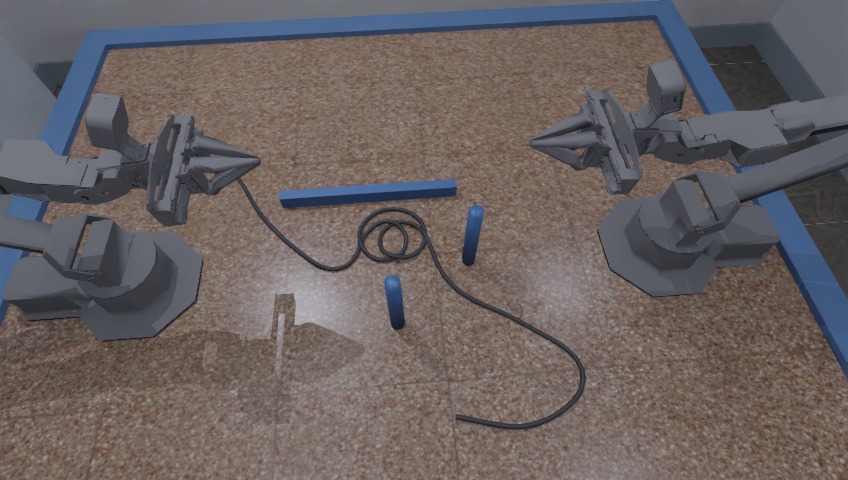} \\
  \end{tabular}
  \caption{\textbf{Domain randomization.} The first camera frame of four tasks at their three domain-randomization seeds. A seed
changes the lighting, the table, the floor and the parts' colours and materials. The geometry, physics, start state and
grader stay those of seed $0$.}
  \label{fig:seeds}
\end{figure}

Each task has three domain-randomization seeds (\cref{fig:seeds}). Seed $0$ is the scene as released. Seeds $1$ and $2$ change only
how the scene looks: the room's lighting, the table top (wood, metal or stone), the table's frame, the floor, and the
colours and materials of the task's parts. A cloth changes its weave and pattern, a wire its metal, a rope its colour.
The parts' shapes, masses and friction, the start state, the cameras, the success rule and the reference stay those of seed
$0$: each seed is derived from seed $0$'s frozen scene, its physical fields must equal seed $0$'s, and its reference
replays to the same world and robot states as seed $0$'s, bit for bit.

The instructions stay valid at every seed. A part
the instruction names by colour keeps its hue (its shade or gloss may change), a part named by its material keeps the
material, and no other part is named by colour. The green dashed goal marker is the same at every seed, and every seed
keeps the camera frames bright enough to read. For the IKEA scenes a seed changes the lighting only. The three runs of the
main evaluation each use one seed (\cref{sec:setup}), so every success rate averages over three looks of each task.

\subsection{How a task is built}
\label{sec:suppl_construction}

Each puzzle task passes the six stages of \cref{fig:pipeline}, and every stage leaves a record in the release.

\begin{figure}[!htbp]
  \centering
  \includegraphics[width=0.6\linewidth]{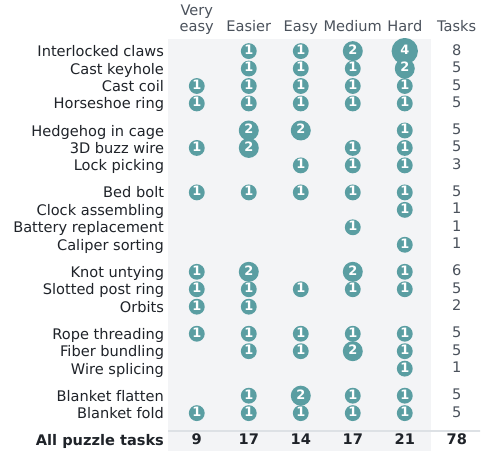}
  \caption{\textbf{Difficulty ratings.} Puzzle tasks per family and rating, with the families grouped by constraint type
as in \cref{tab:families}. The IKEA scenes are unrated.}
  \label{fig:tiers}
\end{figure}

\paragraph{Design.}
A family fixes a puzzle mechanism, a robot and a success rule, and its tasks differ in the scene. Of the $78$ puzzle
tasks, $28$ port earlier scenes to MuJoCo, $19$ come from redesigned tier sets and $31$ are new designs. Each task is rated
within its family as \emph{very easy}, \emph{easier}, \emph{easy}, \emph{medium} or \emph{hard} (\cref{fig:tiers}), and each
step up adds work. In the redesigned interlocked claws, \emph{easy} forces a hand-over and a
$180^\circ$ roll, \emph{medium} adds a third rotation axis at $3.5$\,mm clearance, and \emph{hard} starts with empty hands
and both pieces on the table at $3$\,mm. A redesigned scene must need at least as much as the scene it replaces,
clearly more in one measure such as grasps or tool path, and the old reference and simple breakers, such as a straight
escape, must fail on it.

\paragraph{Freeze.}
The release image builds each scene once, with the SIMD paths of the numerical libraries masked. The frozen instance
holds the settled start state, the geometry, the physics step and solver settings, the grasp rules and the goal. The
task's runtime binding records the SHA-256 of the instance, of every asset file ($5{,}809$ for interlocked claws 05) and
of the tooling. A change of physics makes a new task, never a new version of an old one.

\paragraph{Reference.}
Each family has an oracle program that plans every move on the known scene, often from the measured state of a live
session, and the executed joint targets become the reference. The \std reference keeps the rows, re-timed where a joint
moved faster than the service's $0.04$\,rad per control step. A reference counts as verified when two fresh-process
replays succeed on the same final state hash, the stress observer finds no event on it, a do-nothing run
fails, and it runs within the agent's budget. Its Harbor runs follow (\cref{sec:suppl_validity}).

\paragraph{Stress test.}
A round drives a scene with six strategies: random targets, directed moves that pull, twist, ram and throw parts,
mutated recorded streams, the cross-entropy method maximising a breakdown score, the reference cut at a random row and
perturbed, and periodic inputs of up to $20$\,Hz. An observer marks a breakdown at a non-finite state, a contact deeper
than $10$\,mm, energy beyond the robot's work by more than $0.1$\,J, an uncaused speed jump, or a stored-energy kick
above $3$\,m/s. Candidates are replayed in fresh processes, and a regression set, the $50$ reproducers of earlier faults, must replay without a breakdown.

\paragraph{Package.}
A generator renders both Harbor task directories of a scene from its release record (family, rating, instance, stated
facts, evidence), its texts and its references, and checks every filled text against the facts it must state, such as
the grasp rule, the control step and the rating. A directory holds \texttt{task.toml}, the instruction, the environment
pinned to the image by digest, the verifier with the runtime binding and the reference solution. Continuous integration
re-renders every directory and checks, among other things, that no \std text states a world coordinate
(\cref{sec:suppl_modes}).

\subsection{Assets, licences and release}
\label{sec:suppl_license}

\name's own code is released under the MIT licence: the simulator, the graders, the procedural puzzle generators, the
Gymnasium package and the reference solutions. Its data is released under CC~BY~4.0: the task texts, the frozen scenes,
the reference trajectories and the agent runs. Third-party material keeps its own licence (\cref{tab:license}), and
its licence texts ship next to it. All of these licences permit use, modification and redistribution, including
commercial use, and none is share-alike, so they combine with the MIT and CC~BY~4.0 terms without further conditions.
The robot models are copied unchanged from MuJoCo Menagerie together with their upstream licence files and the hash of
every file. The Poly Haven lighting map and domain-randomization textures are downloaded when the images are built
and checked by hash.

We will release all $180$ task directories, the $90$ tasks in both the \std and the \priv regime, together with their
graders, the reference solution of every task, the pinned image recipes, the Gymnasium package and the full record of
every agent run in this paper. We designed all tasks, graders and reference solutions for this benchmark, including the
puzzle-part meshes.

\begin{table}[!htbp]
\centering
\caption{\textbf{Assets and licences.} Third-party material keeps its upstream licence. The last two rows are \name's own.}
\label{tab:license}
\scriptsize
\setlength{\tabcolsep}{3pt}
\renewcommand{\arraystretch}{1.12}
\begin{fittab}\begin{tabular}{@{}>{\raggedright\arraybackslash}p{9.4cm} >{\raggedright\arraybackslash}p{3.3cm} >{\raggedright\arraybackslash}p{2.6cm}@{}}
\toprule
Asset group & Source & Licence \\
\midrule
ALOHA\,2 model & MuJoCo Menagerie & BSD-3-Clause \\
Franka Panda model & MuJoCo Menagerie & Apache-2.0 \\
Robotiq 2F-85 gripper model & MuJoCo Menagerie & BSD-2-Clause \\
Studio lighting map (HDRI) & Poly Haven & CC0 \\
Table, frame and floor textures (domain randomization) & Poly Haven & CC0 \\
Ground texture, two camera kernels & Genesis & Apache-2.0 \\
Cloth solver kernels (adapted) & Newton & Apache-2.0 \\
Physics engine (with a backported EPA fix) & MuJoCo & Apache-2.0 \\
IKEA scene models and part meshes & built for \name & MIT \\
Puzzle-part meshes & built for \name & MIT \\
\midrule
Simulator, graders, puzzle generators, Gymnasium package, reference solutions & this work & MIT \\
Task texts, frozen scenes, reference trajectories, agent runs & this work & CC~BY~4.0 \\
\bottomrule
\end{tabular}\end{fittab}
\end{table}

\FloatBarrier

\section{Regimes and inputs}
\label{sec:suppl_regimes}

\subsection{The two regimes and the interfaces}
\label{sec:suppl_modes}

Each task ships as two Harbor~\cite{harbor} task directories, one per regime (\cref{fig:modes}), that share the frozen scene, the physics and the success rule. We describe the regimes as a coding agent meets them.

\paragraph{Privileged regime.}
The simulator runs inside the agent's container, so the agent has full access to it. The agent may read the simulator code and the full state, reset to the frozen start, render
any camera and replay as often as it likes, and it delivers one trajectory of control targets. The verifier replays that trajectory twice from the frozen state in a fresh process. The task is solved only if both replays pass the success rule and at least
one reaches the goal before the delivered trajectory ends, not during the hold appended to confirm it. In this regime the agent plans and checks its
motions against a known model.

\paragraph{Standard regime.}
The simulator runs in a separate container, and the agent reaches it only through a fixed observation and action API. The agent observes RGB
images with optional depth, proprioception and calibrated cameras (intrinsics and extrinsics in the robot frame), and
submits chunks of joint or end-effector targets. There is no reset and no privileged state: positions and poses of
objects, the table height and physics parameters are withheld, and the task texts contain no world coordinates. Alongside the images, the simulator publishes the safety readings the grader judges, contact forces and penetration depths. It grades the live episode, and a fresh-process replay of the executed actions confirms it. This regime adds perception from real-sensor inputs and a single committed attempt to the same task.

\paragraph{Interfaces for any policy class.}
Besides the Harbor tasks for coding agents, a Python package wraps every task as a Gymnasium
environment~\cite{towers2024gymnasium} with the same physics and grader, and a reference trajectory stepped through it
ends on exactly the recorded state hash. VLA policies run through this environment, served over an openpi-style
websocket~\cite{black2024pi0}: the runner sends camera images, the robot state and a one-sentence instruction and
receives action chunks. Any other policy steps the environment directly.

\subsection{What a policy receives}
\label{sec:suppl_inputs}

\begin{figure}[!htbp]
  \centering
  \includegraphics[width=\linewidth]{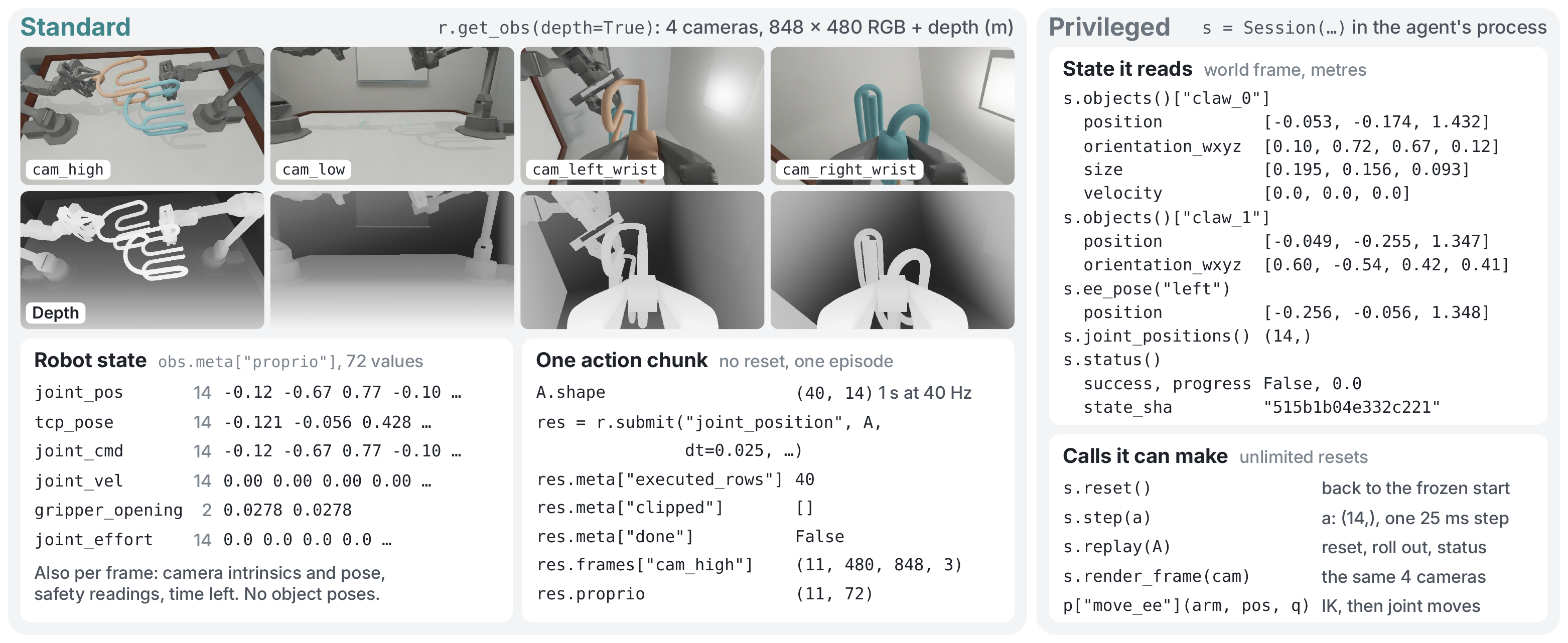}
  \caption{\textbf{What a policy receives in each regime.} Interlocked claws 05 at the frozen start, captured from the released tasks.
\textcolor{stdink}{\std}: the four cameras' RGB and depth (grey, near is light) as one observation returns them, the six groups of the robot state, and one action chunk of $40$ joint-target rows ($1$\,s) with its response. Poses are given in the robot's base frame and no object pose is given.
\textcolor{privink}{\priv}: an excerpt of the state the agent's session reads (world frame, metres) and the calls it can make.}
  \label{fig:inputs}
\end{figure}

\Cref{fig:inputs} shows both regimes on interlocked claws 05, the task of \cref{fig:pipeline}, at its frozen start.
In the \std regime one observation call returns $848\times480$ RGB and metric depth from four cameras, two fixed and two on the wrists, together with their calibration and $72$ proprioceptive values, and one action call executes a chunk of joint targets at the $40$\,Hz control rate and returns the frames and robot states it produced.
In the \priv regime the agent's own Python session reads the pose, size and velocity of every object in the world frame, renders the same four cameras pixel for pixel, and can return to this state and roll out any action sequence as often as it likes.
Both regimes receive the same task sentence and success rule and command the same $14$ joint targets, so the two regimes differ in access.

\subsection{What an agent reads}
\label{sec:suppl_instruction}

\begin{figure}[!tbp]
  \centering
  \includegraphics[width=\linewidth]{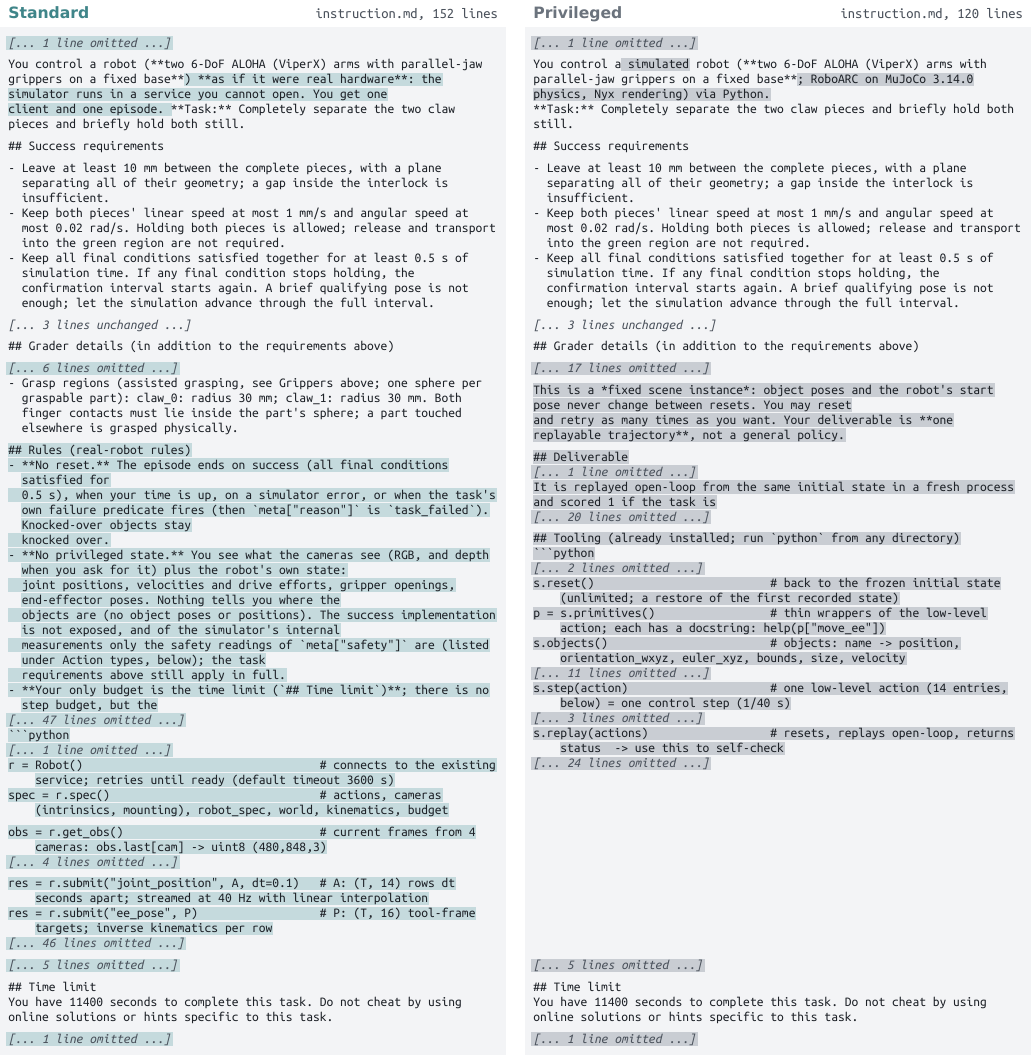}
  \caption{\textbf{What an agent reads.} The two \texttt{instruction.md} files of interlocked claws 05 as Harbor hands them to the agent, \textcolor{stdink}{\std} (left) and \textcolor{privink}{\priv} (right). Every line shown is copied from the file, and long lines are wrapped for print. Text that appears in only one of the two files is tinted. A marker stands for hidden lines, ``unchanged'' when the other file has the same lines and ``omitted'' (tinted) when some of them differ.}
  \label{fig:instruction}
\end{figure}

Harbor hands the agent the task's \texttt{instruction.md} as it is (\cref{sec:suppl_agents}). \Cref{fig:instruction} shows the two files of interlocked claws 05 side by side, $152$ lines for the \std regime and $120$ for the \priv regime.
Both state the same task sentence, success requirements, failure conditions and time limit of $11{,}400$\,s.
The \std file says that the agent drives the robot as if it were real hardware, through a service it cannot open, with one client and one episode.
It allows no reset and gives no object poses, only camera images, depth on request and the robot's own state.
Its client offers an observation call, action chunks of joint or end-effector targets and stateless kinematics, and the grasp regions come with their radius only.
The service records the episode and grades it.
The \priv file gives a Python session on the simulator that resets without limit and reads every object's pose.
It names the success function in the simulator's code and the file with each grasp region's centre.
The deliverable is one trajectory, replayed open-loop in a fresh process and scored $1$ or $0$.
The \std agent's container holds a client library and no simulator, while the \priv agent's container holds the simulator, its source code and the frozen scene.

\section{Validity checks}
\label{sec:suppl_validity}

\paragraph{Solvable: verified references.}
Each puzzle and IKEA scene ships a reference solution that Harbor's reference runner runs through the full grading
path: it scores $1$ in both regimes (for the puzzles at all three domain-randomization seeds), and a do-nothing agent scores $0$.

\paragraph{Reproducible: deterministic replay.}
The containers mask the SIMD code paths of the numerical libraries, so every CPU class builds the same frozen state and
a replay of the same actions ends on the same state hash. The grading replays of every reference are bit-identical on
AMD Zen\,4 and on Intel CPUs with AVX-512 and with AVX2. A verdict in the \priv regime is therefore a property of the delivered trajectory, whichever machine grades it.

\paragraph{Robust: stress acceptance.}
Before release, $46{,}662$ stress episodes of random, directed, optimized and periodic controls drove the puzzle
scenes to check that no motion an agent could command breaks the simulator or yields a success the task text
does not grant. Risky behaviour is clamped: speed and force limits stop the motion and the episode
continues. A success reached with a deep contact between task parts is held for review, a simulator breakdown that
follows an agent's forcing counts as the agent's failure, and any other breakdown is an infrastructure error that
counts as no result. The strategies of a stress round and the thresholds of its observer are in
\cref{sec:suppl_construction}, and the limits of the contact rule in \cref{sec:suppl_audit}.

\paragraph{Fair: grader audits.}
The \priv grader credits a success only when the goal is reached within the delivered trajectory,
before the short hold that confirms it. The release's text checks find no leak in any of the $180$ task directories: no \std task text states a world coordinate, a table height or a camera's distance to the scene, no instruction points to the verifier's files, and no file an agent can read links to an upstream solution, dataset or demonstration, while the reference solutions are mounted for the reference runner only.

\subsection{Physical audit of the solved runs}
\label{sec:suppl_audit}

In the \priv regime an agent can read the simulator and the success rule (\cref{sec:suppl_modes}). This audit checks
that the solved runs are physically legal: no task part may pass through another part or a fixture, and no rope, cord
or fiber may pass through another body.

\paragraph{The grader's contact rule.}
In nine families of rigid parts in contact (interlocked claws, cast keyhole, cast coil, horseshoe ring, hedgehog in
cage, bed bolt, clock assembling, battery replacement and caliper sorting), the grader reads the deepest contact between
a task part and any other body at the end of every physics tick of $2.5$\,ms. It does not credit a success with a
contact deeper than $3$\,mm at one tick or deeper than $2$\,mm for $0.25$\,s, and a contact deeper than $10$\,mm ends
the episode. All $666$ solved runs of these families ($466$ \priv and $200$ \std) therefore stay below these limits at
every tick. The grader has no contact-depth rule in seven other families (3D buzz wire, lock picking, knot untying,
slotted post ring, orbits, rope threading and fiber bundling), which only the replays below cover. Over all $1{,}338$
solved puzzle runs, the verifier records no simulator breakdown, the two grading replays of each of the $926$ \priv
runs end on one state hash, and the replay of each of the $412$ \std runs ends on the hash of its live episode.

\begin{figure}[!htbp]
  \centering
  \includegraphics[width=\linewidth]{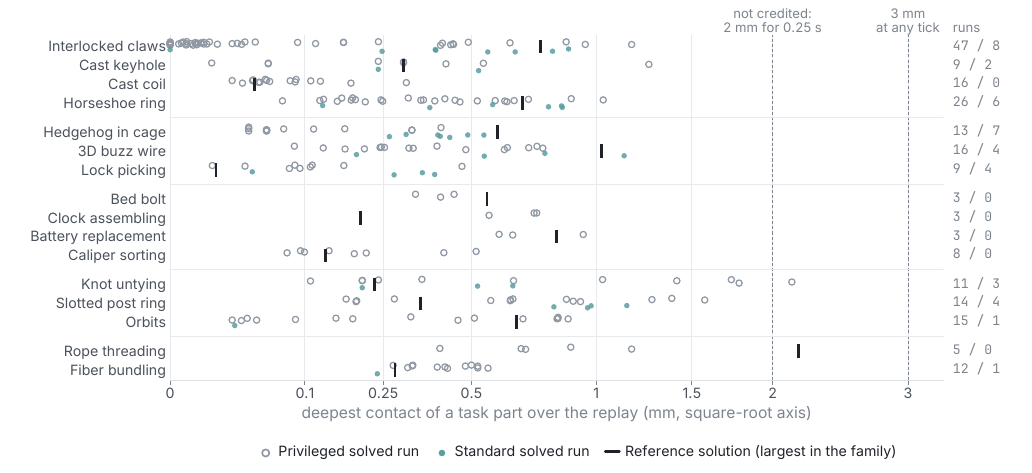}
  \caption{\textbf{Deepest contact of a task part in the solved runs.} One mark per audited run: the deepest contact
between a task part and another task part or a fixture over every simulator step of the verifier's replay. Black bar:
the largest value over the family's replayed reference solutions. Dashed lines: the limits of the grader's contact
rule in the nine families of rigid parts in contact, $2$\,mm held for $0.25$\,s and $3$\,mm at one tick. Right:
audited runs per family (\priv / \std).}
  \label{fig:audit}
\end{figure}

\paragraph{Replays.}
We measure how far below these limits the solved runs stay. We replay $210$ of the $780$ solved \priv runs and $40$ of
the $357$ solved \std runs of the seven agents in $16$ puzzle families through the verifier's own replay path and read
every contact after every simulator step, $0.625$\,ms or shorter (\cref{tab:audit}). Of the \priv runs, $139$ follow a
fixed order that rotates through the agents within each family, as do all \std runs. The other $71$ \priv runs complete
the set of all $87$ solved \priv runs of these families that are shorter than a tenth of their task's reference solution,
the runs that depart furthest from it. Every replay ends in success on the state hash that the verifier recorded. For each
run we take the deepest contact between a task part and another task part or a fixture over the whole episode, and we
measure the reference solutions of $50$ tasks in the same way (\cref{fig:audit}).

\paragraph{Contact depth.}
The deepest contact of a solved run has a median of $0.29$\,mm and a $90$th percentile of $0.86$\,mm, against a median
of $0.22$\,mm and a maximum of $2.18$\,mm for the reference solutions. Of the $250$ runs, $14$ exceed $1$\,mm, $1$
exceeds $2$\,mm and none exceeds $3$\,mm. The largest value, $2.13$\,mm, belongs to a \priv knot-untying run of
Gemini 4 Argon, in which a rope link of $4$\,mm radius sinks into the clamp plate for $25$\,ms. The longest unbroken
stretch above $1$\,mm in any run lasts $1.92$\,s. In $13$ of the $14$ runs above $1$\,mm the deepest contact is with the
table or a fixture, and the deepest contact between two task parts over all runs reads $1.22$\,mm. The $87$ shortest
runs have a median of $0.12$\,mm and a maximum of $2.13$\,mm. Of the $229$ audited runs whose task has a replayed
reference, $26$ are deeper than the reference by more than $0.5$\,mm. Of the $84$ audited runs on tasks with a
documented design clearance, $6$ read a contact deeper than that clearance: $0.33$, $0.40$ and $0.44$\,mm against
$0.075$\,mm on bed bolt, $0.31$ against $0.30$\,mm on cast coil and $0.31$ and $0.49$ against $0.30$\,mm on hedgehog in
cage.

\begin{table}[!htbp]
\centering
\caption{\textbf{Replay audit of the solved runs, per family.} Rule: whether the grader has a contact-depth rule for the family. Solved runs: solved puzzle runs of the seven agents. Audited runs: runs replayed step by step. Deepest contact: the deepest contact of a task part with another task part or a fixture over the replay, with its median, $90$th percentile and maximum over the audited runs of both regimes. Reference: the largest such value over the family's replayed reference solutions, with their number in grey. The blanket families and the IKEA scenes are not in this audit.}
\label{tab:audit}
\footnotesize
\setlength{\tabcolsep}{4.5pt}
\begin{fittab}\begin{tabular}{@{}l c rr rr rrr rr r@{}}
\toprule
 & & \multicolumn{2}{c}{Solved runs} & \multicolumn{2}{c}{Audited runs} & \multicolumn{3}{c}{Deepest contact (mm)} & \multicolumn{2}{c}{Runs above} & \\
\cmidrule(lr){3-4} \cmidrule(lr){5-6} \cmidrule(lr){7-9} \cmidrule(lr){10-11}
Family & Rule & Priv. & Std. & Priv. & Std. & Median & $90$th pct. & Max & $1$\,mm & $2$\,mm & Reference \\
\midrule
Interlocked claws & yes & $103$ & $68$ & $47$ & $8$ & $0.01$ & $0.65$ & $1.17$ & $1$ & $0$ & $0.76$ \textcolor{black!50}{($8$)} \\
Cast keyhole & yes & $54$ & $14$ & $9$ & $2$ & $0.30$ & $0.54$ & $1.26$ & $1$ & $0$ & $0.30$ \textcolor{black!50}{($5$)} \\
Cast coil & yes & $65$ & $21$ & $16$ & $0$ & $0.05$ & $0.15$ & $0.31$ & $0$ & $0$ & $0.04$ \textcolor{black!50}{($5$)} \\
Horseshoe ring & yes & $77$ & $49$ & $26$ & $6$ & $0.38$ & $0.84$ & $1.03$ & $1$ & $0$ & $0.68$ \textcolor{black!50}{($5$)} \\
\addlinespace[2pt]
Hedgehog in cage & yes & $60$ & $39$ & $13$ & $7$ & $0.22$ & $0.44$ & $0.54$ & $0$ & $0$ & $0.59$ \textcolor{black!50}{($5$)} \\
3D buzz wire & no & $40$ & $9$ & $16$ & $4$ & $0.36$ & $0.77$ & $1.14$ & $1$ & $0$ & $1.03$ \textcolor{black!50}{($5$)} \\
Lock picking & no & $40$ & $28$ & $9$ & $4$ & $0.11$ & $0.38$ & $0.47$ & $0$ & $0$ & $0.01$ \textcolor{black!50}{($3$)} \\
\addlinespace[2pt]
Bed bolt & yes & $62$ & $1$ & $3$ & $0$ & $0.40$ & $0.44$ & $0.44$ & $0$ & $0$ & $0.55$ \textcolor{black!50}{($2$)} \\
Clock assembling & yes & $15$ & $6$ & $3$ & $0$ & $0.73$ & $0.74$ & $0.74$ & $0$ & $0$ & $0.20$ \textcolor{black!50}{($1$)} \\
Battery replacement & yes & $11$ & $1$ & $3$ & $0$ & $0.65$ & $0.88$ & $0.94$ & $0$ & $0$ & $0.82$ \textcolor{black!50}{($1$)} \\
Caliper sorting & yes & $19$ & $1$ & $8$ & $0$ & $0.16$ & $0.44$ & $0.52$ & $0$ & $0$ & $0.13$ \textcolor{black!50}{($1$)} \\
\addlinespace[2pt]
Knot untying & no & $62$ & $43$ & $11$ & $3$ & $0.58$ & $1.77$ & $2.13$ & $5$ & $1$ & $0.23$ \textcolor{black!50}{($2$)} \\
Slotted post ring & no & $32$ & $26$ & $14$ & $4$ & $0.84$ & $1.31$ & $1.57$ & $4$ & $0$ & $0.34$ \textcolor{black!50}{($2$)} \\
Orbits & no & $30$ & $27$ & $15$ & $1$ & $0.25$ & $0.83$ & $0.87$ & $0$ & $0$ & $0.66$ \textcolor{black!50}{($2$)} \\
\addlinespace[2pt]
Rope threading & no & $57$ & $17$ & $5$ & $0$ & $0.70$ & $1.06$ & $1.17$ & $1$ & $0$ & $2.18$ \textcolor{black!50}{($1$)} \\
Fiber bundling & no & $53$ & $7$ & $12$ & $1$ & $0.42$ & $0.52$ & $0.56$ & $0$ & $0$ & $0.28$ \textcolor{black!50}{($2$)} \\
\midrule
All & & $780$ & $357$ & $210$ & $40$ & $0.29$ & $0.86$ & $2.13$ & $14$ & $1$ & $2.18$ \textcolor{black!50}{($50$)} \\
\bottomrule
\end{tabular}\end{fittab}
\end{table}

\paragraph{Ropes, cords and fibers.}
In the rope, cord and fiber families a link tunnels when its centre line enters another body, which is a contact as
deep as the link's radius. Over the $63$ audited runs of these $5$ families the deepest contact of a link with any
body, the robot's fingers included, reaches $0.81$ of its radius (a finger on a cord of orbits), and the deepest contact
between two links $0.21$ of the sum of their radii. No centre line enters another body in any of these runs.

\paragraph{Stress observer.}
The stress observer of \cref{sec:suppl_construction} follows $119$ of the replays in full, with its energy accounts and
its causal check of every speed jump, and the other $131$ with its contact checks alone. It reports no breakdown, and no
replay has a non-finite state or a solver warning. Its triggers for violent motion mark $11$ runs: in $6$ a robot link
presses on a fixture with $2.5$ to $7.5$\,kN, which the rules allow, and in $5$ a rope, cord or chain link moves faster
than $3$\,m/s. The verifier records a forcing event, an arm that presses at its torque limit, in $104$
of the $926$ solved \priv puzzle runs and in $127$ of the $412$ solved \std puzzle runs.

\paragraph{IKEA scenes.}
The verifier of the IKEA scenes records the deepest contact of a part with any body in every replay
(\cref{tab:audit_ikea}). Of the $70$ solved IKEA runs, $62$ stay within $0.5$\,mm of their scene's reference solution.
Of the other $8$, $4$ are EKET runs of GPT-6 Astra and GPT-6.1 Sol, out of $11$ solved EKET runs. EKET models the ribs
of its dowels as a compliant envelope of $30$\,N/m, which a side panel overlaps while it is fitted: the reference
solution reads $4.9$\,mm there and these runs $6.0$ to $10.0$\,mm, where a contact deeper than $10$\,mm would end the
episode. Each of them passes the scene's seating gates, which accept a panel within $0.25$ to $0.8$\,mm of its goal
pose. We opened two of these runs at the level of single collision shapes. Their deep contacts lie between the dowels'
envelope and pieces of the side panel, at forces of at most $0.28$\,N.

\begin{table}[!htbp]
\centering
\caption{\textbf{Deepest part contact in the solved IKEA runs,} from the verifier's record of each replay. Smallest and Largest: over the solved runs of the scene. Reference: the scene's reference solution. Deeper runs: solved runs whose deepest contact exceeds the reference by more than $0.5$\,mm. The scene \texttt{ikea-stenerik} has no solved run.}
\label{tab:audit_ikea}
\footnotesize
\setlength{\tabcolsep}{6pt}
\begin{fittab}\begin{tabular}{@{}l r rr r r@{}}
\toprule
 & & \multicolumn{2}{c}{Deepest contact (mm)} & & \\
\cmidrule(lr){3-4}
Scene & Solved runs & Smallest & Largest & Reference (mm) & Deeper runs \\
\midrule
\texttt{ikea-burvik} & $4$ & $0.34$ & $0.56$ & $0.63$ & $0$ \\
\texttt{ikea-eket} & $11$ & $4.92$ & $9.96$ & $4.95$ & $4$ \\
\texttt{ikea-gladom} & $3$ & $0.53$ & $0.70$ & $0.82$ & $0$ \\
\texttt{ikea-ikea365} & $12$ & $0.30$ & $1.13$ & $0.61$ & $1$ \\
\texttt{ikea-kyrre} & $2$ & $0.75$ & $0.83$ & $0.96$ & $0$ \\
\texttt{ikea-lack} & $14$ & $0.17$ & $1.53$ & $0.42$ & $1$ \\
\texttt{ikea-stegesund} & $1$ & $1.59$ & $1.59$ & $0.92$ & $1$ \\
\texttt{ikea-svallet} & $5$ & $0.30$ & $0.49$ & $0.16$ & $0$ \\
\texttt{ikea-uppsta} & $2$ & $2.67$ & $3.36$ & $2.70$ & $1$ \\
\texttt{ikea-variera} & $11$ & $0.45$ & $0.81$ & $0.73$ & $0$ \\
\texttt{ikea-vesken} & $5$ & $1.60$ & $2.50$ & $2.07$ & $0$ \\
\midrule
All & $70$ & & & & $8$ \\
\bottomrule
\end{tabular}\end{fittab}
\end{table}

\paragraph{Coverage.}
The replay sample holds $11$ of the $62$ solved \priv knot-untying runs, $5$ of $57$ in rope threading, $12$ of $53$
in fiber bundling and $9$ of $40$ in lock picking, and \cref{tab:audit} gives the audited runs of every family in both
regimes. The two blanket families ($146$ solved \priv and $55$ solved \std runs) run their cloth on the GPU, outside
this CPU replay.

\section{Evaluation details}
\label{sec:suppl_eval}

\subsection{Agents and harnesses}
\label{sec:suppl_agents}

\begin{table}[!htbp]
\centering
\caption{\textbf{The seven agents as run.} Model: the model string given to Harbor (\texttt{-m}). Harness: the version recorded in each trial's \texttt{agent/trajectory.json}, a range where the harness was installed at its latest release for each trial. Effort: the level and the harness option that carried it. Harbor: the Harbor version that ran the trials.}
\label{tab:agents}
\scriptsize
\setlength{\tabcolsep}{3pt}
\renewcommand{\arraystretch}{1.15}
\begin{fittab}\begin{tabular}{@{}l l l l l@{}}
\toprule
Agent & Model & Harness & Effort, option & Harbor \\
\midrule
GPT-6 Astra & \texttt{gpt-6-astra} & Codex 0.165.0--0.167.2 & xhigh, \texttt{-c model\_reasoning\_effort} & 0.24.0 \\
GPT-6.1 Sol & \texttt{gpt-6.1-sol} & Codex 0.165.0--0.167.2 & xhigh, \texttt{-c model\_reasoning\_effort} & 0.24.0 \\
Claude Opus 5.5 & \texttt{claude-opus-5.5} & Claude Code 2.1.314 & xhigh, \texttt{--effort} & 0.24.0 \\
Gemini 4 Argon & \texttt{google/gemini-4-argon} & Antigravity SDK 0.1.23 & high, SDK \texttt{thinking\_level} & 0.24.0 \\
\addlinespace[2pt]
Kimi 3.1 Pro & \texttt{moonshot/kimi-31-pro-nuvora} & Kimi Code 2.5.2 & max, \texttt{KIMI\_MODEL\_THINKING\_EFFORT} & 0.25.0 \\
DeepSeek v4.1 Flash & \texttt{deepseek/deepseek-v4.1-flash} & DSH 0.2.1-rc.2 & high, ACP option \texttt{reasoning\_effort} & 0.23.0 \\
GLM 5.3 Flash & \texttt{anthropic/glm-5.3-flash} & Claude Code 2.1.318--2.1.325 & high, \texttt{--effort} & 0.23.0 \\
\bottomrule
\end{tabular}\end{fittab}
\end{table}

\Cref{tab:agents} lists how each agent was run. Codex and Claude Code are Harbor's own agents. Antigravity, Kimi Code and DSH run through Harbor agents of \name's repository (\mbox{\texttt{scripts/harbor/agents/}}), which install a pinned harness version during the agent's set-up.

\paragraph{Common settings.}
Each trial is one Harbor job that runs one task in one regime. Harbor hands the agent the task's \texttt{instruction.md} as it is, with no prompt template and nothing added. The agent gets no task-specific primitives and no demonstrations.

The agent's wall-clock limit is the \texttt{[agent] timeout\_sec} of the task's \texttt{task.toml}, which Harbor applies as the timeout of its \texttt{agent\_execution} phase, with no timeout multiplier. It is $11{,}400$\,s for every task, $190$ minutes of agent session, and $21{,}600$\,s for the wire-splicing task. The harness is installed before this clock starts, in Harbor's \texttt{agent\_setup} phase, so the installation does not count against the limit. In the solved trials this phase takes a median of $9.4$, $12.7$ and $13.1$ minutes for GPT-6 Astra, GPT-6.1 Sol and Claude Opus 5.5 and under $1.5$ minutes for the other four agents. DeepSeek v4.1 Flash and GLM 5.3 Flash ran under a limit of $10{,}800$\,s in place of the $11{,}400$\,s: all $67$ solved trials of these two agents record this limit, and all $1{,}341$ solved trials of the other five agents record $11{,}400$\,s. At the cutoff Harbor stops the agent and the trial is graded as it stands.

The agent's container reaches only an allowlist of hosts: the package registries (PyPI and the Debian and Ubuntu archives) and the model API, which every run reached through a usage gateway that records each request's tokens. The gateway holds a trial to $100$\,M input tokens, cache reads included, and $10$\,M output tokens, and a trial that reaches the limit ends as a failure. Kimi 3.1 Pro has a limit of $500$\,M input tokens, because Kimi Code counts cached tokens again on every request. Every harness runs with its web search and web fetch tools off, and Codex also with its apps, plugins and image generation off.

Each trial is pinned to one GPU (the IKEA \priv tasks use none), which it may share with other trials, and reserves the CPUs and memory that its task declares (\cref{sec:suppl_cost}). Each agent runs every task three times in each regime, at domain-randomization seeds $0$, $1$ and $2$, with the same seeds in \std and \priv. A trial that ends in an infrastructure or grader error (no verdict from the grader, a Harbor exception other than the cutoff, a model endpoint the gateway cannot reach, or a model-account failure) counts as no result and runs again from a fresh environment with the same task, seed and configuration. The agent's own end of its session, its cutoff and its token limit are results and are never rerun.

\subsection{VLA policies}
\label{sec:suppl_vla}

We evaluate the DROID checkpoints of $\pi_{0.5}$~\cite{black2025pi05}, MolmoAct2~\cite{fang2026molmoact2}, GR00T N1.7~\cite{nvidia2025gr00t} and Cosmos3-Nano-Policy~\cite{kim2026cosmospolicy} as released. Each is called
as its authors' DROID client calls it, with the same cameras, image size, action chunk and gripper rule. Three adaptations
bridge DROID's robot and ours without changing a task: the cameras move to the poses of DROID's simulated evaluation
scene, the last joint is offset by $\pi/4$ because our gripper is mounted at $-45^\circ$, and the arm starts from DROID's
reset pose. With the tasks' own cameras, $\pi_{0.5}$ came within $3$\,cm of the task object in $0$ of $24$ trial episodes, and
with DROID's camera poses in $7$ of $24$.

\paragraph{Episodes.} Each model runs each of the $18$ puzzles once at each domain-randomization seed $0$, $1$ and $2$, $54$ episodes per
model, in the task's frozen scene. The model receives the task's goal in one sentence, the first sentence of the
task statement (for example ``Separate and place the whole hedgehog ball, including every spike inside the green dashed goal region.''), with its camera
images and the arm's joint and gripper state, and returns a chunk of joint-position targets.

The targets play at DROID's
$15$\,Hz, interpolated onto the puzzles' $40$\,Hz control steps, and every limit of the \std regime applies to them. The
simulation waits for each reply, so inference time does not count. An episode ends at the task's own end or at twice
the length of its reference solution, $3{,}800$ to $70{,}600$ control steps. It is graded as in the \std regime: a live
success counts when a fresh-process replay of the executed actions confirms it. \Cref{tab:vla_models} lists what each
model receives and returns.

\begin{figure}[!htbp]
  \centering
  \includegraphics[width=\linewidth]{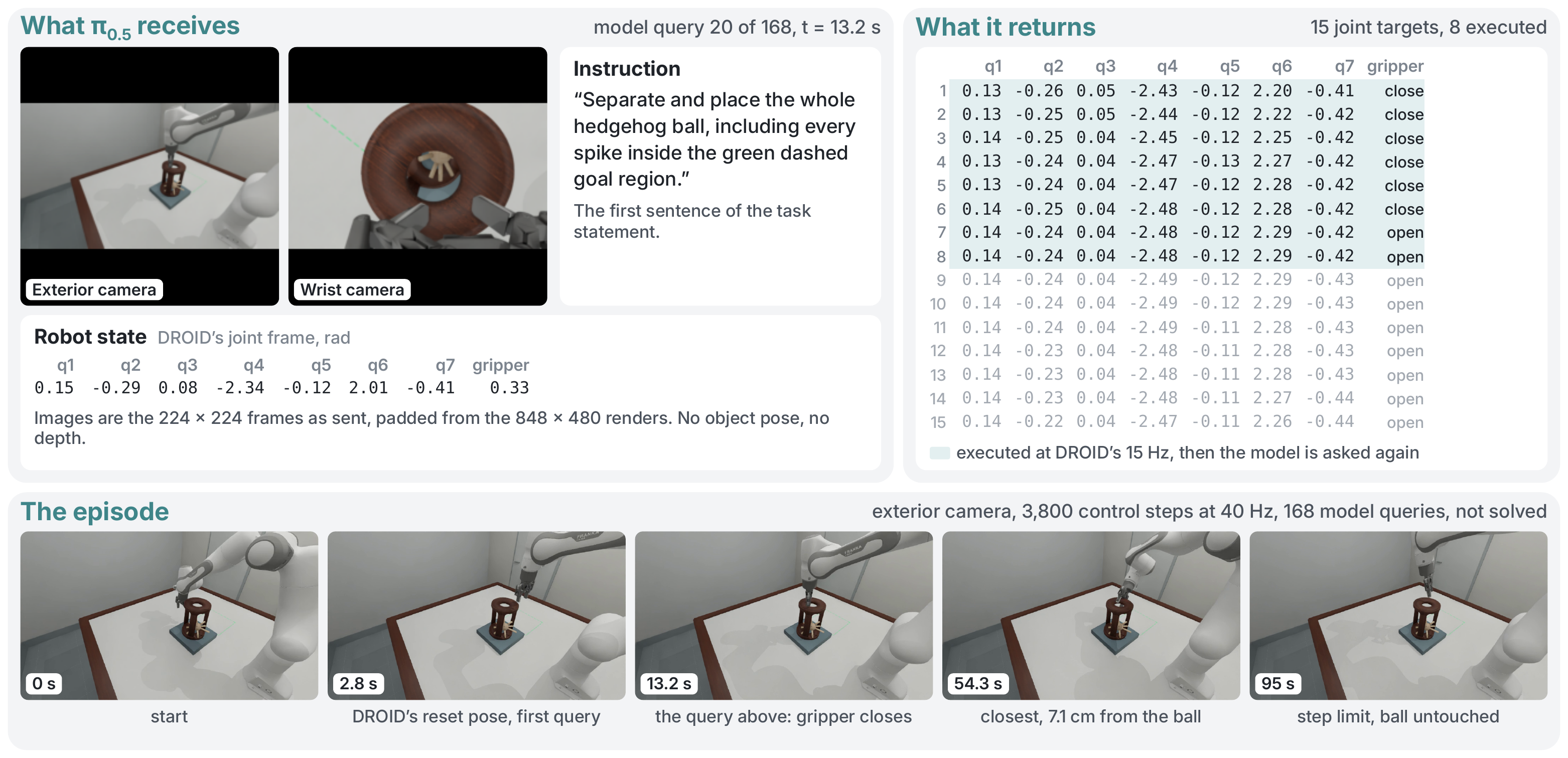}
  \caption{\textbf{One VLA episode.} $\pi_{0.5}$ on \texttt{hedgehog-in-cage-02}, seed $0$, \std regime. Top: what the
  model receives at one query and the chunk of joint targets it returns, of which rows $1$ to $8$ run. Bottom: the
  model's exterior camera over the episode.}
  \label{fig:vla_example}
\end{figure}

\Cref{fig:vla_example} shows one $\pi_{0.5}$ episode on \texttt{hedgehog-in-cage-02} at seed $0$, run again at these
settings for this figure. Each query sends two $224\times224$ images, the instruction and the arm's state, and the model
returns $15$ joint targets, of which the first $8$ run. Over $168$ queries and $3{,}800$ control steps the arm moves
above and around the cage and commands its gripper closed in $3$ queries. It never reaches into the cage, comes no
closer than $7.1$\,cm to the ball and stops at the step limit.

\begin{table}[!htbp]
\centering
\caption{\textbf{The four VLA policies as run.} Cameras: exterior views and the wrist view, as each model's DROID client
sends them. Chunk: actions returned per query and executed before the next query.}
\label{tab:vla_models}
\scriptsize
\setlength{\tabcolsep}{4.5pt}
\begin{fittab}\begin{tabular}{@{}l l l c l@{}}
\toprule
Policy & Checkpoint & Cameras, image size & Chunk & Gripper \\
\midrule
$\pi_{0.5}$ & openpi \texttt{pi05\_droid\_jointpos} & $1$ exterior + wrist, $224\times224$ & $15$, $8$ & binarised at $0.5$ \\
MolmoAct2 & \texttt{MolmoAct2-DROID}, float32 & $1$ exterior + wrist, $320\times180$ & $15$, $15$ & continuous in $[0,1]$ \\
GR00T N1.7 & \texttt{GR00T-N1.7-DROID}, bfloat16 & $1$ exterior + wrist, $320\times180$ & $40$, $8$ & binarised at $0.5$ \\
Cosmos3-Nano-Policy & \texttt{Cosmos3-Nano-Policy-DROID} & $2$ exterior + wrist, $640\times360$ & $32$, $32$ & binarised at $0.5$ \\
\bottomrule
\end{tabular}\end{fittab}
\end{table}

\paragraph{Coding agents on the one-arm puzzles.} For reference, \cref{tab:vla_agents} lists the seven coding agents on the
$18$ one-arm puzzles in the \std regime. Each policy class runs under its own protocol, with its own inputs, interface
and budget (\cref{sec:suppl_modes}).

\begin{table}[!htbp]
\centering
\caption{\textbf{Coding agents on the $\mathbf{18}$ one-arm puzzles.} Success rate (\%) in the \std regime, three runs per
puzzle. Passage: narrow passage, Thread.: threading. Numbers in parentheses count the puzzles.}
\label{tab:vla_agents}
\footnotesize
\setlength{\tabcolsep}{4pt}
\begin{fittab}\begin{tabular}{@{}l c c c@{}}
\toprule
Agent & Passage ($13$) & Thread. ($5$) & All ($18$) \\
\midrule
GPT-6 Astra (Codex) & $56.4$ & $6.7$ & $42.6$ \\
GPT-6.1 Sol (Codex) & $51.3$ & $13.3$ & $40.7$ \\
Claude Opus 5.5 (Claude Code) & $38.5$ & $6.7$ & $29.6$ \\
Gemini 4 Argon (Antigravity) & $43.6$ & $13.3$ & $35.2$ \\
Kimi 3.1 Pro (Kimi Code) & $2.6$ & $6.7$ & $3.7$ \\
DeepSeek v4.1 Flash (DSH) & $0.0$ & $0.0$ & $0.0$ \\
GLM 5.3 Flash (Claude Code) & $2.6$ & $0.0$ & $1.9$ \\
\bottomrule
\end{tabular}\end{fittab}
\end{table}

\subsection{An agent's run, step by step}
\label{sec:suppl_run}

\begin{figure}[!htbp]
  \centering
  \includegraphics[width=\linewidth]{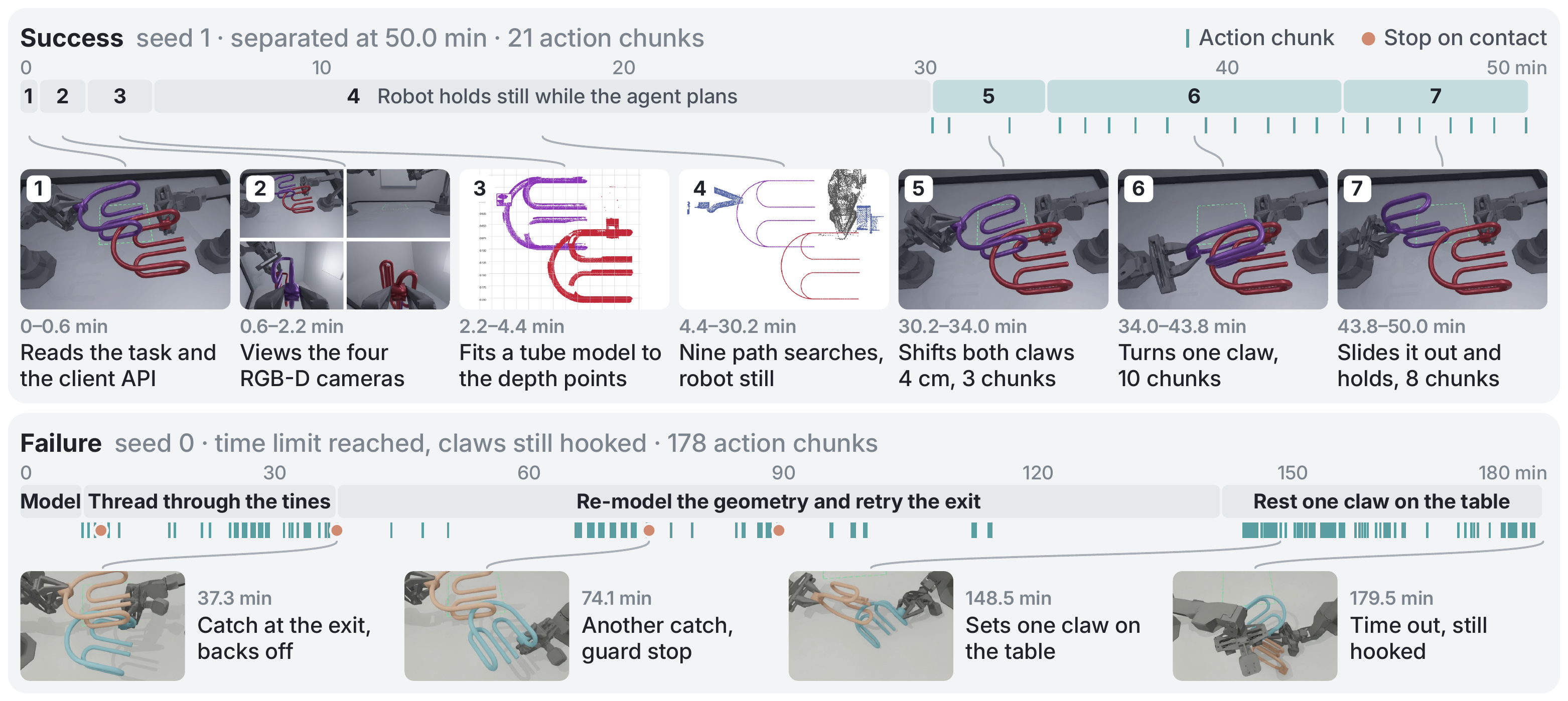}
  \caption{\textbf{One task, two runs of GPT-6 Astra in the \std regime} (\texttt{interlocked-claws-05}). Top: the
  successful run in seven phases cut from its transcript, on the agent's clock. Tiles $1$ to $4$ are images the agent itself
  viewed (a camera frame, all four cameras, its depth-point plot and its gripper-aware model), tiles $5$ to $7$ come from
  the grader's replay. Bottom: the failed run on a $3$-hour axis. Ticks mark action chunks, dots mark stops on contact.}
  \label{fig:run}
\end{figure}

\Cref{fig:run} follows GPT-6 Astra (Codex, xhigh effort) through two \std runs on \texttt{interlocked-claws-05}, where the
two ALOHA\,2 arms hold two hooked claw pieces that must end at least $10$\,mm apart. Minutes count from the agent's start.
In the successful run (seed $1$) the agent sends no motion for the first $30$ minutes. The agent reads the task and the client
API, looks through the four RGB-D cameras, separates the two claws by colour in the depth images and fits a tube model to
the points.

It then searches for an exit path over the claws' relative pose. The first path clears the claws but collides
with the grippers and reaches a wrist limit, so the agent adds both grippers to the model, checks each path continuously
between waypoints and finally plans backward from a pose centred in the narrow slot. Nine tree searches and three path
optimizations run in this phase.

The first motion comes at $30.2$ minutes. The agent shifts both claws $4$\,cm toward the
front of the table, turns one claw through the slot and slides it out in $21$ action chunks. After each chunk it lets the
arms settle, compares the measured hand pose with the plan and usually views the top camera image, and three small
chunks correct the drift. The service reports success at $50.0$ minutes after $119.8$\,s of simulated time, and
the grader's replay confirms it with no review or forcing events.

In the failed run (seed $0$) the agent builds a similar
camera model and starts moving after $7.3$ minutes along a route through the tines. It stops on contact four times, at $9.5$,
$37.3$, $74.1$ and $89.4$ minutes, and after the later catches refits the geometry from new camera views. At $94$ minutes it writes that ``the
fuller reconstruction found a section of the bend that the earlier model missed''. At $141.5$ minutes it rests one claw on
the table to let it pivot free. The claws are still hooked when the run times out, after $178$ action chunks, with
the agent's context summarized twice along the way.

\subsection{Information-access ablation}
\label{sec:suppl_ablation}

\paragraph{Tasks.} The ablation uses $15$ rigid puzzles, chosen from those that GPT-6 Astra in Codex (high effort) solved in the \priv regime and failed
in the \std regime in control runs: battery-replacement-01, bed-bolt-01, -02, -04 and -05, buzz-wire-3d-02, -06 and -10, caliper-sorting-01, cast-coil-04, -07 and -08, cast-keyhole-05 and -06, and hedgehog-in-cage-07. By construction its \priv success on them is $100\,\%$ at selection.

\paragraph{Arms.} Each arm is a copy of the \std task with only its additions: object poses, part geometry, both,
or both and the internal state, each without and with resets. Every addition arrives through the \std service in the robot's frame, the agent still acts only
through action chunks, and the simulator, rollouts, the success rule and the reference stay hidden. The budget and the
grader are those of the main evaluation.
\begin{itemize}[leftmargin=1.2em,itemsep=1pt,topsep=2pt]
  \item \emph{Resets}: a reset call restores the frozen start any number of times within the budget. A failure ends only
  the attempt, and the last attempt is graded.
  \item \emph{Object poses}: every response carries the pose of every task part.
  \item \emph{Part geometry}: a read-only call returns the geometry of every task part and fixture as the simulator holds it, including
  the collision shapes (type, pose, sizes, convex vertices), with moving links at their start configuration, and the regions of
  the assisted grasp. Masses, friction and other physical parameters are not given.
  \item \emph{Internal state}: every response also carries the simulator's internal state, including joint positions and velocities, the current poses of moving
  links, the grasp phase and mechanism readings.
\end{itemize}

\paragraph{Runs.} \Cref{tab:ablation} gives the counts behind \cref{fig:ladder}. Every arm ran three times per task at
domain-randomization seed $0$, $45$ runs per arm, and each run of $15$ tasks gives one success rate. The geometry-only arm gives each part's shapes in its
own frame, fixed parts included, and no position of any part. Kimi 3.1 Pro, run at the plain \std regime
and with internal state, solves $0$ ($0/45$), $6.7$ ($1/15$) and, with resets,
$0$ ($0/15$), against $13.3$ ($6/45$) in the \priv regime on these tasks.

\paragraph{The same puzzles in the main evaluation.} The ablation runs at high effort and at seed $0$, and the main
evaluation at xhigh effort and at seeds $0$, $1$ and $2$. In the main evaluation GPT-6 Astra solves $7$ of its $45$ trials
on these $15$ puzzles in the \std regime ($15.6\,\%$, with $2$, $4$ and $1$ of $15$ at the three seeds) and $41$ of $45$ in
the \priv regime ($91.1\,\%$, with $12$, $14$ and $15$ of $15$). The \std arm of the ablation without additions solves
$11$ of $45$ ($24.4\,\%$), and the two \std rates do not differ significantly (Fisher's exact test, two-sided, $p=0.43$).
The grey line of \cref{fig:ladder} marks the \priv level at selection, $100\,\%$.

\begin{table}[!htbp]
\centering
\caption{\textbf{Information-access ablation, GPT-6 Astra in the \std regime.} Success (\%) on the $15$ tasks, with successes
and trials, and the standard deviation over the three runs. Rows as in \cref{fig:ladder}.}
\label{tab:ablation}
\small\setlength{\tabcolsep}{6pt}
\begin{fittab}\begin{tabular}{@{}l r r@{}}
\toprule
Added to \std & No reset & + resets \\
\midrule
Nothing & $24.4\pm10.2$ ($11/45$) & $42.2\pm10.2$ ($19/45$) \\
Object poses only & $28.9\pm3.8$ ($13/45$) & $51.1\pm16.8$ ($23/45$) \\
Part geometry only & $37.8\pm10.2$ ($17/45$) & $51.1\pm16.8$ ($23/45$) \\
Poses + geometry & $60.0\pm6.7$ ($27/45$) & $73.3\pm6.7$ ($33/45$) \\
Poses + geometry + internal state & $71.1\pm3.8$ ($32/45$) & $93.3\pm0.0$ ($42/45$) \\
\bottomrule
\end{tabular}\end{fittab}
\end{table}

\FloatBarrier

\section{Additional results}
\label{sec:suppl_results}

\subsection{Per-family results}
\label{sec:suppl_family}

\begin{figure}[!htbp]
  \centering
  \includegraphics[width=0.6\linewidth]{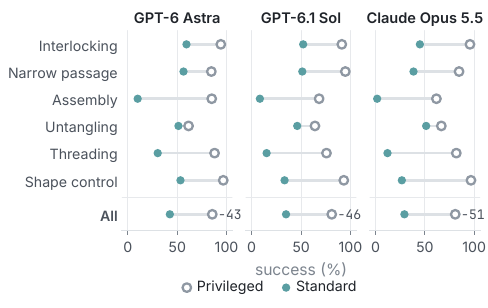}
  \caption{\textbf{The drop from the \priv to the \std regime, per constraint type.} Success of GPT-6 Astra, GPT-6.1 Sol and Claude Opus 5.5 in the \priv regime (\textcolor{privink}{ring}) and the \std regime (\textcolor{stdink}{dot}). The number beside
each agent is its drop over all tasks. The drop is largest in threading and assembly and smallest in untangling.}
  \label{fig:gap}
\end{figure}

\begin{figure}[!htbp]
  \centering
  \includegraphics[width=\linewidth]{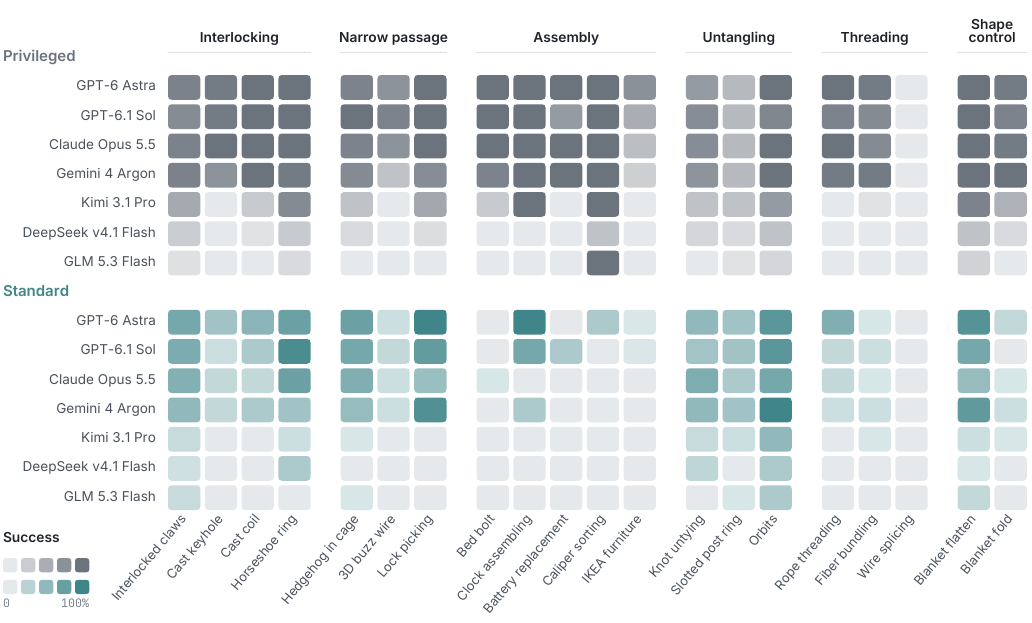}
  \caption{\textbf{Success per agent and task family.} For each agent the \priv regime (grey) sits above the \std
regime (teal), and a darker cell means a higher rate. Families are grouped by constraint type, and \cref{tab:family}
gives the numbers.}
  \label{fig:heatmap}
\end{figure}

\Cref{tab:main_sd} gives every cell of \cref{tab:main} with its run-to-run spread. Resampling the $20$ families, the $19$ puzzle families and the IKEA scenes,
$10{,}000$ times keeps GPT-6 Astra first in the \std regime in more than $99\%$ of draws and
places every proprietary agent above every open-weight agent in both regimes in every draw, while the closer
proprietary agents trade places in part of the draws. Weighting every family equally gives the same \std ranking and
swaps GPT-6.1 Sol and Claude Opus 5.5 in the \priv regime. \Cref{tab:family} breaks
\cref{tab:main} down by family, with every agent that has trials.
\Cref{fig:heatmap} shows every family at a glance and \cref{fig:gap} the drop from the \priv to the \std regime
by constraint type. \Cref{tab:tiers} splits the same trials by the difficulty rating of each task.

\begin{table}[!htbp]
\centering
\caption{\textbf{Success rate (\%) with its run-to-run spread.} Each cell of \cref{tab:main} as the mean
$\pm$ the standard deviation over the three domain-randomization seeds.}
\label{tab:main_sd}
\scriptsize
\setlength{\tabcolsep}{2.2pt}
\begin{fittab}\begin{tabular}{@{}l @{\hspace{4pt}} r@{\,}l r@{\,}l r@{\,}l r@{\,}l r@{\,}l r@{\,}l r@{\,}l @{\hspace{10pt}} r@{\,}l r@{\,}l r@{\,}l r@{\,}l r@{\,}l r@{\,}l r@{\,}l@{}}
\toprule
 & \multicolumn{14}{c}{Privileged} & \multicolumn{14}{c}{Standard} \\
\cmidrule(lr){2-15} \cmidrule(lr){16-29}
Agent & \multicolumn{2}{c}{Interl.} & \multicolumn{2}{c}{Passage} & \multicolumn{2}{c}{Assem.} & \multicolumn{2}{c}{Untang.} & \multicolumn{2}{c}{Thread.} & \multicolumn{2}{c}{Shape} & \multicolumn{2}{c}{All} & \multicolumn{2}{c}{Interl.} & \multicolumn{2}{c}{Passage} & \multicolumn{2}{c}{Assem.} & \multicolumn{2}{c}{Untang.} & \multicolumn{2}{c}{Thread.} & \multicolumn{2}{c}{Shape} & \multicolumn{2}{c}{All} \\
\midrule
\multicolumn{29}{@{}l}{\textsc{proprietary}} \\
GPT-6 Astra & $94$ & {\scriptsize\color{black!50}$\pm 3$} & $85$ & {\scriptsize\color{black!50}$\pm 8$} & $85$ & {\scriptsize\color{black!50}$\pm 9$} & $62$ & {\scriptsize\color{black!50}$\pm 8$} & $88$ & {\scriptsize\color{black!50}$\pm 5$} & $97$ & {\scriptsize\color{black!50}$\pm 6$} & $86$ & {\scriptsize\color{black!50}$\pm 4$} & $59$ & {\scriptsize\color{black!50}$\pm 11$} & $56$ & {\scriptsize\color{black!50}$\pm 9$} & $10$ & {\scriptsize\color{black!50}$\pm 0$} & $51$ & {\scriptsize\color{black!50}$\pm 4$} & $30$ & {\scriptsize\color{black!50}$\pm 10$} & $53$ & {\scriptsize\color{black!50}$\pm 15$} & $43$ & {\scriptsize\color{black!50}$\pm 5$} \\
GPT-6.1 Sol & $91$ & {\scriptsize\color{black!50}$\pm 4$} & $95$ & {\scriptsize\color{black!50}$\pm 4$} & $68$ & {\scriptsize\color{black!50}$\pm 6$} & $64$ & {\scriptsize\color{black!50}$\pm 4$} & $76$ & {\scriptsize\color{black!50}$\pm 10$} & $93$ & {\scriptsize\color{black!50}$\pm 6$} & $81$ & {\scriptsize\color{black!50}$\pm 1$} & $52$ & {\scriptsize\color{black!50}$\pm 12$} & $51$ & {\scriptsize\color{black!50}$\pm 9$} & $8$ & {\scriptsize\color{black!50}$\pm 6$} & $46$ & {\scriptsize\color{black!50}$\pm 0$} & $15$ & {\scriptsize\color{black!50}$\pm 14$} & $33$ & {\scriptsize\color{black!50}$\pm 15$} & $35$ & {\scriptsize\color{black!50}$\pm 5$} \\
Claude Opus 5.5 & $96$ & {\scriptsize\color{black!50}$\pm 4$} & $85$ & {\scriptsize\color{black!50}$\pm 8$} & $62$ & {\scriptsize\color{black!50}$\pm 3$} & $67$ & {\scriptsize\color{black!50}$\pm 4$} & $82$ & {\scriptsize\color{black!50}$\pm 9$} & $97$ & {\scriptsize\color{black!50}$\pm 6$} & $81$ & {\scriptsize\color{black!50}$\pm 1$} & $45$ & {\scriptsize\color{black!50}$\pm 3$} & $38$ & {\scriptsize\color{black!50}$\pm 8$} & $2$ & {\scriptsize\color{black!50}$\pm 3$} & $51$ & {\scriptsize\color{black!50}$\pm 12$} & $12$ & {\scriptsize\color{black!50}$\pm 5$} & $27$ & {\scriptsize\color{black!50}$\pm 12$} & $29$ & {\scriptsize\color{black!50}$\pm 2$} \\
Gemini 4 Argon & $88$ & {\scriptsize\color{black!50}$\pm 9$} & $62$ & {\scriptsize\color{black!50}$\pm 8$} & $50$ & {\scriptsize\color{black!50}$\pm 5$} & $64$ & {\scriptsize\color{black!50}$\pm 9$} & $85$ & {\scriptsize\color{black!50}$\pm 5$} & $100$ & {\scriptsize\color{black!50}$\pm 0$} & $73$ & {\scriptsize\color{black!50}$\pm 4$} & $38$ & {\scriptsize\color{black!50}$\pm 3$} & $44$ & {\scriptsize\color{black!50}$\pm 12$} & $2$ & {\scriptsize\color{black!50}$\pm 3$} & $54$ & {\scriptsize\color{black!50}$\pm 8$} & $12$ & {\scriptsize\color{black!50}$\pm 5$} & $47$ & {\scriptsize\color{black!50}$\pm 6$} & $31$ & {\scriptsize\color{black!50}$\pm 5$} \\
\addlinespace[3pt]
\multicolumn{29}{@{}l}{\textsc{open-weight}} \\
Kimi 3.1 Pro & $42$ & {\scriptsize\color{black!50}$\pm 7$} & $26$ & {\scriptsize\color{black!50}$\pm 4$} & $17$ & {\scriptsize\color{black!50}$\pm 3$} & $38$ & {\scriptsize\color{black!50}$\pm 8$} & $3$ & {\scriptsize\color{black!50}$\pm 5$} & $67$ & {\scriptsize\color{black!50}$\pm 6$} & $31$ & {\scriptsize\color{black!50}$\pm 4$} & $9$ & {\scriptsize\color{black!50}$\pm 4$} & $3$ & {\scriptsize\color{black!50}$\pm 4$} & $0$ & {\scriptsize\color{black!50}$\pm 0$} & $21$ & {\scriptsize\color{black!50}$\pm 4$} & $3$ & {\scriptsize\color{black!50}$\pm 5$} & $10$ & {\scriptsize\color{black!50}$\pm 10$} & $7$ & {\scriptsize\color{black!50}$\pm 2$} \\
DeepSeek v4.1 Flash & $16$ & {\scriptsize\color{black!50}$\pm 3$} & $8$ & {\scriptsize\color{black!50}$\pm 0$} & $2$ & {\scriptsize\color{black!50}$\pm 3$} & $18$ & {\scriptsize\color{black!50}$\pm 4$} & $0$ & {\scriptsize\color{black!50}$\pm 0$} & $23$ & {\scriptsize\color{black!50}$\pm 6$} & $11$ & {\scriptsize\color{black!50}$\pm 1$} & $12$ & {\scriptsize\color{black!50}$\pm 5$} & $0$ & {\scriptsize\color{black!50}$\pm 0$} & $0$ & {\scriptsize\color{black!50}$\pm 0$} & $15$ & {\scriptsize\color{black!50}$\pm 8$} & $0$ & {\scriptsize\color{black!50}$\pm 0$} & $3$ & {\scriptsize\color{black!50}$\pm 6$} & $6$ & {\scriptsize\color{black!50}$\pm 2$} \\
GLM 5.3 Flash & $6$ & {\scriptsize\color{black!50}$\pm 7$} & $0$ & {\scriptsize\color{black!50}$\pm 0$} & $5$ & {\scriptsize\color{black!50}$\pm 0$} & $5$ & {\scriptsize\color{black!50}$\pm 4$} & $0$ & {\scriptsize\color{black!50}$\pm 0$} & $10$ & {\scriptsize\color{black!50}$\pm 0$} & $4$ & {\scriptsize\color{black!50}$\pm 2$} & $6$ & {\scriptsize\color{black!50}$\pm 7$} & $3$ & {\scriptsize\color{black!50}$\pm 4$} & $0$ & {\scriptsize\color{black!50}$\pm 0$} & $8$ & {\scriptsize\color{black!50}$\pm 8$} & $0$ & {\scriptsize\color{black!50}$\pm 0$} & $10$ & {\scriptsize\color{black!50}$\pm 0$} & $4$ & {\scriptsize\color{black!50}$\pm 3$} \\
\bottomrule
\end{tabular}\end{fittab}
\end{table}

\begin{table}[!htbp]
\centering
\caption{\textbf{Success rate (\%) per task family.} Astra: GPT-6 Astra, Sol: GPT-6.1 Sol, Opus 5.5: Claude
Opus 5.5, Argon: Gemini 4 Argon, Kimi: Kimi 3.1 Pro, DeepSeek: DeepSeek v4.1 Flash, GLM: GLM 5.3 Flash.
Wire splicing has a $6$\,h budget. Bold marks the best per row and regime, and underline the second.}
\label{tab:family}
\scriptsize
\setlength{\tabcolsep}{2.6pt}
\begin{fittab}\begin{tabular}{@{}l r *{7}{c} @{\hspace{8pt}} *{7}{c}@{}}
\toprule
 & & \multicolumn{7}{c}{Privileged} & \multicolumn{7}{c}{Standard} \\
\cmidrule(lr){3-9} \cmidrule(lr){10-16}
Constraint type / family & \#Tasks & Astra & Sol & Opus 5.5 & Argon & Kimi & DeepSeek & GLM & Astra & Sol & Opus 5.5 & Argon & Kimi & DeepSeek & GLM \\
\midrule
\multicolumn{16}{@{}l}{\textsc{rigid}} \\
\textbf{Interlocking} & \textbf{$\mathbf{23}$} & \underline{$94$} & $91$ & \textbf{$\mathbf{96}$} & $88$ & $42$ & $16$ & $6$ & \textbf{$\mathbf{59}$} & \underline{$52$} & $45$ & $38$ & $9$ & $12$ & $6$ \\
\quad Interlocked claws & $8$ & \textbf{$\mathbf{88}$} & \underline{$79$} & \textbf{$\mathbf{88}$} & \textbf{$\mathbf{88}$} & $54$ & $25$ & $8$ & \textbf{$\mathbf{67}$} & \underline{$62$} & $58$ & $50$ & $17$ & $12$ & $17$ \\
\quad Cast keyhole & $5$ & \underline{$93$} & \underline{$93$} & \textbf{$\mathbf{100}$} & $73$ & $0$ & $0$ & $0$ & \textbf{$\mathbf{40}$} & $13$ & \underline{$20$} & \underline{$20$} & $0$ & $0$ & $0$ \\
\quad Cast coil & $5$ & \textbf{$\mathbf{100}$} & \textbf{$\mathbf{100}$} & \textbf{$\mathbf{100}$} & \textbf{$\mathbf{100}$} & \underline{$27$} & $7$ & $0$ & \textbf{$\mathbf{53}$} & \underline{$33$} & $20$ & \underline{$33$} & $0$ & $0$ & $0$ \\
\quad Horseshoe ring & $5$ & \textbf{$\mathbf{100}$} & \textbf{$\mathbf{100}$} & \textbf{$\mathbf{100}$} & \underline{$93$} & $80$ & $27$ & $13$ & \underline{$73$} & \textbf{$\mathbf{93}$} & \underline{$73$} & $40$ & $13$ & $33$ & $0$ \\
\textbf{Narrow passage} & \textbf{$\mathbf{13}$} & \underline{$85$} & \textbf{$\mathbf{95}$} & \underline{$85$} & $62$ & $26$ & $8$ & $0$ & \textbf{$\mathbf{56}$} & \underline{$51$} & $38$ & $44$ & $3$ & $0$ & $3$ \\
\quad Hedgehog in cage & $5$ & \underline{$87$} & \textbf{$\mathbf{100}$} & \underline{$87$} & $80$ & $33$ & $13$ & $0$ & \textbf{$\mathbf{73}$} & \underline{$67$} & $60$ & $47$ & $7$ & $0$ & $7$ \\
\quad 3D buzz wire & $5$ & \underline{$73$} & \textbf{$\mathbf{87}$} & \underline{$73$} & $33$ & $0$ & $0$ & $0$ & \underline{$13$} & \textbf{$\mathbf{20}$} & \underline{$13$} & \underline{$13$} & $0$ & $0$ & $0$ \\
\quad Lock picking & $3$ & \textbf{$\mathbf{100}$} & \textbf{$\mathbf{100}$} & \textbf{$\mathbf{100}$} & \underline{$78$} & $56$ & $11$ & $0$ & \textbf{$\mathbf{100}$} & $78$ & $44$ & \underline{$89$} & $0$ & $0$ & $0$ \\
\textbf{Assembly} & \textbf{$\mathbf{20}$} & \textbf{$\mathbf{85}$} & \underline{$68$} & $62$ & $50$ & $17$ & $2$ & $5$ & \textbf{$\mathbf{10}$} & \underline{$8$} & $2$ & $2$ & $0$ & $0$ & $0$ \\
\quad Bed bolt & $5$ & \textbf{$\mathbf{100}$} & \textbf{$\mathbf{100}$} & \textbf{$\mathbf{100}$} & \underline{$87$} & $27$ & $0$ & $0$ & $0$ & $0$ & \textbf{$\mathbf{7}$} & $0$ & $0$ & $0$ & $0$ \\
\quad Clock assembling & $1$ & \textbf{$\mathbf{100}$} & \textbf{$\mathbf{100}$} & \textbf{$\mathbf{100}$} & \textbf{$\mathbf{100}$} & \textbf{$\mathbf{100}$} & $0$ & $0$ & \textbf{$\mathbf{100}$} & \underline{$67$} & $0$ & $33$ & $0$ & $0$ & $0$ \\
\quad Battery replacement & $1$ & \textbf{$\mathbf{100}$} & \underline{$67$} & \textbf{$\mathbf{100}$} & \textbf{$\mathbf{100}$} & $0$ & $0$ & $0$ & $0$ & \textbf{$\mathbf{33}$} & $0$ & $0$ & $0$ & $0$ & $0$ \\
\quad Caliper sorting & $1$ & \textbf{$\mathbf{100}$} & \textbf{$\mathbf{100}$} & \textbf{$\mathbf{100}$} & \textbf{$\mathbf{100}$} & \textbf{$\mathbf{100}$} & \underline{$33$} & \textbf{$\mathbf{100}$} & \textbf{$\mathbf{33}$} & $0$ & $0$ & $0$ & $0$ & $0$ & $0$ \\
\quad IKEA furniture & $12$ & \textbf{$\mathbf{75}$} & \underline{$50$} & $36$ & $22$ & $0$ & $0$ & $0$ & \textbf{$\mathbf{6}$} & \textbf{$\mathbf{6}$} & $0$ & $0$ & $0$ & $0$ & $0$ \\
\addlinespace
\multicolumn{16}{@{}l}{\textsc{soft}} \\
\textbf{Untangling} & \textbf{$\mathbf{13}$} & $62$ & \underline{$64$} & \textbf{$\mathbf{67}$} & \underline{$64$} & $38$ & $18$ & $5$ & \underline{$51$} & $46$ & \underline{$51$} & \textbf{$\mathbf{54}$} & $21$ & $15$ & $8$ \\
\quad Knot untying & $6$ & $67$ & \textbf{$\mathbf{78}$} & \textbf{$\mathbf{78}$} & \underline{$72$} & $33$ & $17$ & $0$ & \underline{$50$} & $39$ & \textbf{$\mathbf{61}$} & \underline{$50$} & $17$ & $22$ & $0$ \\
\quad Slotted post ring & $5$ & \textbf{$\mathbf{40}$} & \textbf{$\mathbf{40}$} & \textbf{$\mathbf{40}$} & \textbf{$\mathbf{40}$} & \underline{$33$} & $13$ & $7$ & \textbf{$\mathbf{40}$} & \textbf{$\mathbf{40}$} & \underline{$33$} & \textbf{$\mathbf{40}$} & $13$ & $0$ & $7$ \\
\quad Orbits & $2$ & \textbf{$\mathbf{100}$} & \underline{$83$} & \textbf{$\mathbf{100}$} & \textbf{$\mathbf{100}$} & $67$ & $33$ & $17$ & \underline{$83$} & \underline{$83$} & $67$ & \textbf{$\mathbf{100}$} & $50$ & $33$ & $33$ \\
\textbf{Threading} & \textbf{$\mathbf{11}$} & \textbf{$\mathbf{88}$} & $76$ & $82$ & \underline{$85$} & $3$ & $0$ & $0$ & \textbf{$\mathbf{30}$} & \underline{$15$} & $12$ & $12$ & $3$ & $0$ & $0$ \\
\quad Rope threading & $5$ & \textbf{$\mathbf{100}$} & $87$ & \textbf{$\mathbf{100}$} & \underline{$93$} & $0$ & $0$ & $0$ & \textbf{$\mathbf{60}$} & \underline{$20$} & \underline{$20$} & $13$ & $0$ & $0$ & $0$ \\
\quad Fiber bundling & $5$ & \textbf{$\mathbf{93}$} & \underline{$80$} & \underline{$80$} & \textbf{$\mathbf{93}$} & $7$ & $0$ & $0$ & \underline{$7$} & \textbf{$\mathbf{13}$} & \underline{$7$} & \textbf{$\mathbf{13}$} & \underline{$7$} & $0$ & $0$ \\
\quad Wire splicing & $1$ & $0$ & $0$ & $0$ & $0$ & $0$ & $0$ & $0$ & $0$ & $0$ & $0$ & $0$ & $0$ & $0$ & $0$ \\
\textbf{Shape control} & \textbf{$\mathbf{10}$} & \underline{$97$} & $93$ & \underline{$97$} & \textbf{$\mathbf{100}$} & $67$ & $23$ & $10$ & \textbf{$\mathbf{53}$} & $33$ & $27$ & \underline{$47$} & $10$ & $3$ & $10$ \\
\quad Blanket flatten & $5$ & \textbf{$\mathbf{100}$} & \textbf{$\mathbf{100}$} & \textbf{$\mathbf{100}$} & \textbf{$\mathbf{100}$} & \underline{$87$} & $33$ & $20$ & \textbf{$\mathbf{87}$} & $67$ & $47$ & \underline{$80$} & $13$ & $7$ & $20$ \\
\quad Blanket fold & $5$ & \underline{$93$} & $87$ & \underline{$93$} & \textbf{$\mathbf{100}$} & $47$ & $13$ & $0$ & \textbf{$\mathbf{20}$} & $0$ & $7$ & \underline{$13$} & $7$ & $0$ & $0$ \\
\addlinespace
\midrule
\textbf{All} & $90$ & \textbf{$\mathbf{86}$} & \underline{$81$} & \underline{$81$} & $73$ & $31$ & $11$ & $4$ & \textbf{$\mathbf{43}$} & \underline{$35$} & $29$ & $31$ & $7$ & $6$ & $4$ \\
\midrule
\#Trials & & $270$ & $270$ & $270$ & $270$ & $270$ & $270$ & $270$ & $270$ & $270$ & $270$ & $270$ & $270$ & $270$ & $270$ \\
\bottomrule
\end{tabular}\end{fittab}
\end{table}

\begin{table}[!htbp]
\centering
\caption{\textbf{Success rate (\%) per difficulty rating}, over the trials of \cref{tab:main}. Ratings are set within a family (\cref{tab:tasks}). V.\,easy: very easy. The IKEA scenes have no rating.}
\label{tab:tiers}
\scriptsize
\setlength{\tabcolsep}{3.6pt}
\begin{fittab}\begin{tabular}{@{}l *{6}{c} @{\hspace{8pt}} *{6}{c}@{}}
\toprule
 & \multicolumn{6}{c}{\textcolor{stdink}{\textbf{Standard}}} & \multicolumn{6}{c}{\textcolor{privink}{\textbf{Privileged}}} \\
\cmidrule(lr){2-7} \cmidrule(l){8-13}
Agent & V.\,easy & Easier & Easy & Medium & Hard & IKEA & V.\,easy & Easier & Easy & Medium & Hard & IKEA \\
\textcolor{black!50}{Tasks} & $9$ & $17$ & $14$ & $17$ & $21$ & $12$ & $9$ & $17$ & $14$ & $17$ & $21$ & $12$ \\
\midrule
GPT-6 Astra & $55.6$ & $60.8$ & $50.0$ & $29.4$ & $49.2$ & $5.6$ & $96.3$ & $94.1$ & $88.1$ & $80.4$ & $82.5$ & $75.0$ \\
GPT-6.1 Sol & $51.9$ & $54.9$ & $33.3$ & $29.4$ & $33.3$ & $5.6$ & $96.3$ & $94.1$ & $81.0$ & $80.4$ & $82.5$ & $50.0$ \\
Claude Opus 5.5 & $70.4$ & $43.1$ & $35.7$ & $15.7$ & $23.8$ & $0.0$ & $100.0$ & $92.2$ & $88.1$ & $82.4$ & $82.5$ & $36.1$ \\
Gemini 4 Argon & $59.3$ & $51.0$ & $26.2$ & $21.6$ & $30.2$ & $0.0$ & $81.5$ & $96.1$ & $73.8$ & $80.4$ & $74.6$ & $22.2$ \\
\addlinespace[2pt]
Kimi 3.1 Pro & $14.8$ & $23.5$ & $2.4$ & $2.0$ & $1.6$ & $0.0$ & $55.6$ & $37.3$ & $40.5$ & $21.6$ & $36.5$ & $0.0$ \\
DeepSeek v4.1 Flash & $14.8$ & $5.9$ & $11.9$ & $2.0$ & $3.2$ & $0.0$ & $22.2$ & $21.6$ & $2.4$ & $5.9$ & $12.7$ & $0.0$ \\
GLM 5.3 Flash & $7.4$ & $7.8$ & $7.1$ & $0.0$ & $3.2$ & $0.0$ & $11.1$ & $3.9$ & $4.8$ & $0.0$ & $7.9$ & $0.0$ \\
\bottomrule
\end{tabular}\end{fittab}
\end{table}

\subsection{Wire splicing and slotted post ring 03 to 05}
\label{sec:suppl_hardfam}

No agent solves wire splicing or slotted post ring 03, 04 and 05 in either regime (\cref{tab:family}). These four tasks
account for $168$ runs of the seven agents, $165$ of them with a verdict. This section records where these runs stop,
how they use their time, and what replays of the reference solutions show (\cref{tab:hardfam}).

\begin{table}[!htbp]
\centering
\caption{\textbf{The four tasks that no agent solves.} Runs: runs of the seven agents, three per agent. Staged: runs with stage data. Stages passed: number of runs that pass $0$ to $4$ of the task's six stages in a row. Minutes: median length of the agent's session. Executed: median number of control steps that a \std run executes. Reference: length of the reference solution in control steps. Resets: median number of resets of the scene in a \priv run. No trajectory: \priv runs that deliver no trajectory.}
\label{tab:hardfam}
\footnotesize
\setlength{\tabcolsep}{3.5pt}
\begin{fittab}\begin{tabular}{@{}l l rr rrrrr r rr rr@{}}
\toprule
 & & & & \multicolumn{5}{c}{Stages passed} & & \multicolumn{2}{c}{Control steps} & & \\
\cmidrule(lr){5-9} \cmidrule(lr){11-12}
Task & Regime & Runs & Staged & $0$ & $1$ & $2$ & $3$ & $4$ & Minutes & Executed & Reference & Resets & No trajectory \\
\midrule
\texttt{wire-splicing-01} & Priv. & $21$ & $19$ & $8$ & $0$ & $4$ & $7$ & $0$ & $348$ & -- & $14{,}882$ & $3$ & $2$ \\
 & Std. & $21$ & $19$ & $7$ & $0$ & $7$ & $5$ & $0$ & $318$ & $12{,}778$ & $14{,}882$ & -- & -- \\
\addlinespace[2pt]
\texttt{slotted-post-ring-03} & Priv. & $21$ & $19$ & $5$ & $3$ & $0$ & $6$ & $5$ & $180$ & -- & $9{,}193$ & $17$ & $2$ \\
 & Std. & $21$ & $21$ & $3$ & $4$ & $1$ & $9$ & $4$ & $178$ & $9{,}381$ & $9{,}193$ & -- & -- \\
\addlinespace[2pt]
\texttt{slotted-post-ring-04} & Priv. & $21$ & $19$ & $7$ & $1$ & $1$ & $6$ & $4$ & $180$ & -- & $10{,}580$ & $5$ & $3$ \\
 & Std. & $21$ & $21$ & $3$ & $6$ & $1$ & $6$ & $5$ & $178$ & $8{,}044$ & $10{,}580$ & -- & -- \\
\addlinespace[2pt]
\texttt{slotted-post-ring-05} & Priv. & $21$ & $20$ & $7$ & $3$ & $1$ & $3$ & $6$ & $180$ & -- & $9{,}407$ & $11$ & $2$ \\
 & Std. & $21$ & $21$ & $2$ & $4$ & $2$ & $10$ & $3$ & $179$ & $9{,}246$ & $9{,}407$ & -- & -- \\
\bottomrule
\end{tabular}\end{fittab}
\end{table}

\paragraph{Where the runs stop.}
Each of the four tasks has six stages (\cref{fig:suppl_ladder}), and $159$ of the runs have stage data. No run passes
more than three stages of wire splicing or more than four stages of a slotted post ring task. GPT-6 Astra and GPT-6.1
Sol pass three stages of wire splicing in all $6$ of their \priv runs, and in $5$ of these $6$ runs the final message
says that the coil or the fold of the red end is not completed. On the slotted post ring tasks the two agents pass three
or four stages in all $36$ runs, and in $24$ of them the final message says that the cord is still on the ring or that
the cord, a bead or a disc still crosses the boundary of the goal.

\paragraph{Time.}
The \std runs use a median of $178$ minutes of agent session on the slotted post ring tasks and $318$ minutes on wire
splicing, under limits of $190$ and $360$ minutes ($180$ minutes for two agents, \cref{sec:suppl_agents}).
One control step of these scenes takes $0.45$ to $0.48$\,s of wall clock in the \std service on the slotted post ring
tasks and $0.70$\,s on wire splicing (medians over the runs), and executing actions takes $40$ to $47\,\%$ of an agent's time.
The \std runs execute a median of $9.4$k, $8.0$k and $9.2$k control steps on slotted post ring 03, 04 and 05 and $12.8$k
on wire splicing. The reference solutions take $9.2$k, $10.6$k, $9.4$k and $14.9$k control steps (\cref{tab:tasks}), and
$31$ of the $82$ \std runs with a record execute at least as many as their reference.
In the \priv regime an agent resets the scene a median of $17$, $5$ and $11$ times on the three slotted post ring tasks
and $3$ times on wire splicing. Of the $84$ \priv runs, $9$ deliver no trajectory, and $22$ of the $75$ delivered
trajectories are at least half as long as the reference.

\paragraph{References.}
The reference of a slotted post ring task is recorded from the family's oracle program, which plans each move on the
known scene (\cref{sec:suppl_construction}). We replay two references at seed $0$ in the release image.
The reference of slotted post ring 03 reaches the goal within its own $9{,}193$ rows.
The reference of wire splicing reaches the goal within its own $14{,}882$ rows, on the state hash of the release record,
in $137$ minutes of wall clock on one core, $38\,\%$ of the task's time limit. Seeds $1$ and $2$ use the same action rows.

To measure the margin of a reference, we shift the commanded tool point of each arm by one Gaussian offset that stays
fixed over the whole trajectory, and replay the shifted rows. The shifted reference does not reach the goal at any
standard deviation we replay, with one replay each: $0.05$, $0.1$, $0.5$, $1$ and $2$\,mm per axis on slotted post ring
03 and $2$\,mm on slotted post ring 05. The smallest shift moves the two tool points by $0.07$ and $0.05$\,mm. The
recorded rows of this reference therefore work as an exact replay. A shifted replay follows the rows without
feedback, while the oracle program that recorded them runs in a live session and checks the scene at the end of each
stage, so the test measures the sensitivity of the open-loop rows.

\paragraph{Simulator.}
The verifier's records of these runs ($157$ records) hold no simulator instability, no contact held for review and no
infrastructure error, and no shifted replay raises one. A forcing event, an arm that presses at its torque limit,
occurs in $94$ of the runs. The largest recorded contact force is $4{,}442$\,N, between the gripper of DeepSeek v4.1
Flash and the fixed jig in a \std wire-splicing run, and five runs exceed $500$\,N.

\subsection{Per-model analysis}
\label{sec:suppl_models}

\Cref{tab:suppl_models} lists per agent how the failed runs end, how many are stuck on their stop step and how often a step is planned first. Astra, Sol and Argon end $82$ to $94\,\%$ of their failed runs with the budget used up, and none of their labelled failed runs ends with a give-up. Opus~5.5, DeepSeek and GLM use up the budget in $1$ to $14\,\%$ of their failed runs and give up in $54$ to $86\,\%$, most often saying that time is running out. Kimi is in between ($58\,\%$ and $32\,\%$). Being stuck differs less between agents: $42$ to $56\,\%$ of the timed failed runs of each of the five agents with step times. Planning first does not follow success. Argon and GLM plan first on $83$ and $77\,\%$ of their labelled steps, but Argon solves far more trials than GLM (\cref{tab:main}). Astra and Sol, the two agents with the highest success rates, plan first on $35$ and $32\,\%$.

\begin{table}[!htbp]
\centering
\caption{\textbf{Failed runs per agent.} Endings are shares of the failed runs, as in \cref{fig:endings}a: Budget exhausted
is counted over all failed runs, and the give-ups and other endings come from the labelled sample, weighted to all failed runs.
Stuck is the share of timed failed runs whose stop step takes at least a third of the run time and has at least $3$ failed
attempts, as in \cref{fig:endings}b, for the five agents with step times (raw counts, timed failed runs in parentheses).
Plan first is the share of the $43$ labelled steps per agent that the agent plans before its first attempt, weighted, as in
\cref{fig:endings}d.}
\label{tab:suppl_models}
\footnotesize
\setlength{\tabcolsep}{4pt}
\begin{fittab}\begin{tabular}{@{}l r rrrr r r@{}}
\toprule
 & & \multicolumn{4}{c}{Ending (\%)} & & \\
\cmidrule(lr){3-6}
Agent & Failed runs & Budget exhausted & Gave up for time & Gave up otherwise & Other & Stuck (\%) & Plan first (\%) \\
\midrule
Astra & $181$ & $82.3$ & $0$ & $0$ & $17.7$ & $47$ \textcolor{black!50}{($66$)} & $35$ \\
Sol & $222$ & $88.3$ & $0$ & $0$ & $11.7$ & $56$ \textcolor{black!50}{($73$)} & $32$ \\
Opus 5.5 & $230$ & $1.7$ & $56.3$ & $29.8$ & $12.2$ & $42$ \textcolor{black!50}{($72$)} & $42$ \\
Argon & $239$ & $94.1$ & $0$ & $0$ & $5.9$ & -- & $83$ \\
Kimi & $436$ & $58.0$ & $26.0$ & $5.6$ & $10.3$ & $48$ \textcolor{black!50}{($136$)} & $60$ \\
DeepSeek & $447$ & $1.1$ & $42.3$ & $21.5$ & $35.1$ & -- & $73$ \\
GLM & $459$ & $13.5$ & $43.1$ & $11.2$ & $32.1$ & $47$ \textcolor{black!50}{($139$)} & $77$ \\
\midrule
All & $2{,}214$ & $40.4$ & $28.5$ & $10.8$ & $20.3$ & $48$ \textcolor{black!50}{($486$)} & $55$ \\
\bottomrule
\end{tabular}\end{fittab}
\end{table}

\subsection{Success against the time limit and a lite subset}
\label{sec:suppl_budget}

\paragraph{Time to solve.}
For every solved trial we read the time from the start of the agent's session, Harbor's \texttt{agent\_execution} phase,
to the moment the trial is solved. In the \std regime this moment is the success entry of the service's episode log. In
the \priv regime it is the last write of the trajectory file that the verifier replays. We read the time for all
$1{,}408$ solved trials ($992$ \priv and $416$ \std), and for $14$ of them, one per agent and regime, we checked it
against the agent's transcript. \Cref{fig:budget} and \cref{tab:budget} give the share of an agent's $270$ trials per
regime that are solved within $t$ minutes, and \cref{tab:budget_times} the time to solve of the solved trials.

Every trial ran under its full limit (\cref{sec:suppl_agents}), so a trial solved at minute $40$ counts for every budget
from $40$ minutes on. The agents read their remaining time and pace themselves for the full limit. In
the \priv regime the last write of the accepted file can be later than the first working save. The session log of the
task tooling gives a lower bound: $774$ of the $992$ solved \priv trials log a successful recording, $338$ of them more
than one minute before the final write. With that earlier time GPT-6 Astra reaches $24.4\,\%$, $51.1\,\%$ and $72.2\,\%$
at $30$, $60$ and $120$ minutes, against $21.1\,\%$, $46.3\,\%$ and $68.2\,\%$ with the final write.

\paragraph{Results.}
No agent solves a trial within $5$ minutes, and the earliest solved trial of any agent is solved at minute $5.9$.
GPT-6 Astra solves $5.6\,\%$ of its \priv trials and $2.6\,\%$ of its \std trials within $15$ minutes,
$46.3\,\%$ and $16.7\,\%$ within one hour, and $68.2\,\%$ and $29.3\,\%$ within two hours, against $85.6\,\%$ and
$42.6\,\%$ under the full limit. Its median solved trial takes $55$ minutes in the \priv regime and $75$ minutes in the
\std regime. Over the seven agents the median is $65$ minutes in the \priv regime and $77$ minutes in the \std regime,
and $45\,\%$ of the $992$ solved \priv trials and $38\,\%$ of the $416$ solved \std trials are solved within one hour.

The curves of three agents still rise at the limit. After minute $160$ GPT-6 Astra solves $14$ \priv and $11$ \std
trials, GPT-6.1 Sol $13$ and $14$ and Gemini 4 Argon $14$ and $15$, of $270$ each, while Claude Opus 5.5 solves $0$ and
$2$. Of all solved trials, $26$ are solved between minute $180$ and minute $190$: $3$ \priv and $7$ \std trials of
GPT-6 Astra, $1$ \priv trial of GPT-6.1 Sol, $3$ \priv and $7$ \std trials of Gemini 4 Argon and $5$ \priv trials of
Kimi 3.1 Pro. Cut at $180$ minutes, GPT-6 Astra solves $84.4\,\%$ and $40.0\,\%$, Gemini 4 Argon $72.2\,\%$ and
$28.1\,\%$ and Kimi 3.1 Pro $29.6\,\%$ of its \priv trials, and in the \std regime Claude Opus 5.5 has solved more
trials than Gemini 4 Argon ($79$ against $76$, where the full limit gives $79$ against $83$).

The order of the agents depends on the budget in the \priv regime. GPT-6 Astra has solved the most trials up to minute
$77$ and from minute $163$ on, and Claude Opus 5.5 from minute $80$ to minute $160$ ($74.1\,\%$ at two hours, against
$68.2\,\%$ for GPT-6 Astra and $65.9\,\%$ for GPT-6.1 Sol). In the \std regime GPT-6 Astra has solved the most trials
at every budget from minute $24$ on.

\begin{table}[!htbp]
\centering
\caption{\textbf{Success (\%) within a time budget.} Share of the $270$ trials of an agent and regime that are solved within $t$ minutes of the start of the agent's session, read from runs made under the full limit. Limit: the rate of \cref{tab:main}, at $190$ minutes ($180$ minutes for DeepSeek v4.1 Flash and GLM 5.3 Flash). All: the seven agents together, $1{,}890$ trials per regime.}
\label{tab:budget}
\footnotesize
\setlength{\tabcolsep}{7pt}
\begin{fittab}\begin{tabular}{@{}l *{12}{r}@{}}
\toprule
 & \multicolumn{12}{c}{Budget in minutes} \\
\cmidrule(l){2-13}
Agent & $5$ & $10$ & $15$ & $20$ & $30$ & $45$ & $60$ & $90$ & $120$ & $150$ & $180$ & Limit \\
\midrule
\multicolumn{13}{@{}l}{\textsc{privileged}} \\
GPT-6 Astra & $0.0$ & $1.5$ & $5.6$ & $10.7$ & $21.1$ & $32.2$ & $46.3$ & $59.3$ & $68.2$ & $78.9$ & $84.4$ & $85.6$ \\
GPT-6.1 Sol & $0.0$ & $2.2$ & $5.2$ & $8.9$ & $15.6$ & $25.6$ & $34.4$ & $55.9$ & $65.9$ & $73.3$ & $80.7$ & $81.1$ \\
Claude Opus 5.5 & $0.0$ & $0.4$ & $1.5$ & $4.4$ & $13.0$ & $26.7$ & $39.3$ & $62.6$ & $74.1$ & $80.0$ & $80.7$ & $80.7$ \\
Gemini 4 Argon & $0.0$ & $0.0$ & $0.0$ & $2.6$ & $6.3$ & $17.4$ & $28.1$ & $43.3$ & $57.0$ & $64.8$ & $72.2$ & $73.3$ \\
\addlinespace[2pt]
Kimi 3.1 Pro & $0.0$ & $0.0$ & $0.4$ & $0.7$ & $4.1$ & $7.4$ & $13.0$ & $15.9$ & $23.7$ & $27.0$ & $29.6$ & $31.5$ \\
DeepSeek v4.1 Flash & $0.0$ & $0.0$ & $0.0$ & $0.0$ & $0.0$ & $0.7$ & $2.2$ & $5.2$ & $8.2$ & $10.0$ & $10.7$ & $10.7$ \\
GLM 5.3 Flash & $0.0$ & $0.0$ & $0.4$ & $0.4$ & $0.4$ & $0.7$ & $1.9$ & $2.2$ & $2.6$ & $3.3$ & $4.4$ & $4.4$ \\
\addlinespace[2pt]
All & $0.0$ & $0.6$ & $1.9$ & $4.0$ & $8.6$ & $15.8$ & $23.6$ & $34.9$ & $42.8$ & $48.2$ & $51.9$ & $52.5$ \\
\midrule
\multicolumn{13}{@{}l}{\textsc{standard}} \\
GPT-6 Astra & $0.0$ & $0.7$ & $2.6$ & $4.1$ & $7.0$ & $11.1$ & $16.7$ & $24.8$ & $29.3$ & $34.4$ & $40.0$ & $42.6$ \\
GPT-6.1 Sol & $0.0$ & $0.4$ & $2.2$ & $3.0$ & $5.2$ & $9.3$ & $14.1$ & $19.6$ & $26.3$ & $29.3$ & $34.8$ & $34.8$ \\
Claude Opus 5.5 & $0.0$ & $0.0$ & $0.0$ & $0.7$ & $2.2$ & $7.0$ & $10.7$ & $18.5$ & $23.7$ & $28.1$ & $29.3$ & $29.3$ \\
Gemini 4 Argon & $0.0$ & $0.0$ & $0.0$ & $0.0$ & $0.4$ & $4.1$ & $9.3$ & $15.9$ & $20.0$ & $24.1$ & $28.1$ & $30.7$ \\
\addlinespace[2pt]
Kimi 3.1 Pro & $0.0$ & $0.0$ & $0.0$ & $0.0$ & $1.1$ & $2.2$ & $3.3$ & $4.8$ & $5.6$ & $6.7$ & $7.0$ & $7.0$ \\
DeepSeek v4.1 Flash & $0.0$ & $0.0$ & $0.0$ & $0.0$ & $0.4$ & $1.5$ & $2.6$ & $3.7$ & $4.8$ & $5.2$ & $5.6$ & $5.6$ \\
GLM 5.3 Flash & $0.0$ & $0.0$ & $0.0$ & $0.4$ & $0.4$ & $1.1$ & $1.1$ & $1.9$ & $2.2$ & $3.0$ & $4.1$ & $4.1$ \\
\addlinespace[2pt]
All & $0.0$ & $0.2$ & $0.7$ & $1.2$ & $2.4$ & $5.2$ & $8.2$ & $12.8$ & $16.0$ & $18.7$ & $21.3$ & $22.0$ \\
\bottomrule
\end{tabular}\end{fittab}
\end{table}

\begin{table}[!htbp]
\centering
\caption{\textbf{Time to solve of the solved trials.} Solved: solved trials of the agent in the regime. Minutes: quartiles (Q1, Median, Q3) of the time from the start of the agent's session to the moment the trial is solved. Within: share of the solved trials that are solved within one and two hours.}
\label{tab:budget_times}
\footnotesize
\setlength{\tabcolsep}{4pt}
\begin{fittab}\begin{tabular}{@{}l r rrr rr @{\hspace{12pt}} r rrr rr@{}}
\toprule
 & \multicolumn{6}{c}{Privileged} & \multicolumn{6}{c}{Standard} \\
\cmidrule(lr){2-7} \cmidrule(l){8-13}
 & & \multicolumn{3}{c}{Minutes} & \multicolumn{2}{c}{Within (\%)} & & \multicolumn{3}{c}{Minutes} & \multicolumn{2}{c}{Within (\%)} \\
\cmidrule(lr){3-5} \cmidrule(lr){6-7} \cmidrule(lr){9-11} \cmidrule(l){12-13}
Agent & Solved & Q1 & Median & Q3 & $60$\,min & $120$\,min & Solved & Q1 & Median & Q3 & $60$\,min & $120$\,min \\
\midrule
GPT-6 Astra & $231$ & $31$ & $55$ & $100$ & $54$ & $80$ & $115$ & $43$ & $75$ & $132$ & $39$ & $69$ \\
GPT-6.1 Sol & $219$ & $36$ & $65$ & $102$ & $42$ & $81$ & $94$ & $43$ & $76$ & $120$ & $40$ & $76$ \\
Claude Opus 5.5 & $218$ & $40$ & $62$ & $86$ & $49$ & $92$ & $79$ & $48$ & $76$ & $111$ & $37$ & $81$ \\
Gemini 4 Argon & $198$ & $46$ & $76$ & $113$ & $38$ & $78$ & $83$ & $57$ & $83$ & $142$ & $30$ & $65$ \\
\addlinespace[2pt]
Kimi 3.1 Pro & $85$ & $46$ & $85$ & $120$ & $41$ & $75$ & $19$ & $43$ & $60$ & $112$ & $47$ & $79$ \\
DeepSeek v4.1 Flash & $29$ & $60$ & $92$ & $118$ & $21$ & $76$ & $15$ & $46$ & $82$ & $95$ & $47$ & $87$ \\
GLM 5.3 Flash & $12$ & $50$ & $83$ & $139$ & $42$ & $58$ & $11$ & $55$ & $90$ & $147$ & $27$ & $54$ \\
\midrule
All & $992$ & $40$ & $65$ & $104$ & $45$ & $82$ & $416$ & $48$ & $77$ & $126$ & $38$ & $73$ \\
\bottomrule
\end{tabular}\end{fittab}
\end{table}

\paragraph{Lite subset.}
A full evaluation of one agent takes $540$ trials and, over the $3{,}780$ scored trials of the seven agents, an average
of $1{,}612$ trial-hours of wall clock from the start of the environment to the verdict. The lite subset keeps $30$ of the
$90$ tasks (\cref{tab:lite_tasks}). Each constraint type keeps a third of its tasks, and each family keeps a third of
its tasks rounded down or up. Among such subsets we take the one whose $14$ success rates and $7$ regime gaps of the
seven agents lie closest to those of the full suite, found from $200{,}000$ random stratified draws whose $12$ best
are improved by swaps within a constraint type. The subset holds $18$ of the $20$ families. Battery replacement and
caliper sorting, with one task each, are absent.

On the lite subset every rate and every gap is within $1.5$ points of its full-suite value, and the seven agents keep
their order in both regimes (\cref{tab:lite}). The subset takes $180$ trials and $556$ trial-hours per agent, $34\,\%$
of the full suite, and $185$ trial-hours for one seed. Wire splicing is in the subset and takes $45$ of the $556$ hours.

The subset was fitted to these seven agents. When it is chosen without one agent (five repeats, $70$ held-out rates),
that agent's two rates on it differ from its full-suite rates by $3.2$ points (root mean square, largest $9.3$) and its
gap by $4.7$ points (largest $13.0$). A random subset with the same strata gives $4.3$ and $4.6$ points and keeps the
exact order of the seven agents in $22\,\%$ (\std) and $53\,\%$ (\priv) of $20{,}000$ draws. With one seed, $30$ trials
per rate, the largest deviation of a rate from its full-suite value is $5.6$ to $8.9$ points. When the families are
resampled as in \cref{sec:suppl_family}, the $95\,\%$ interval of a rate is $25.6$ points wide on the lite subset
against $19.7$ on the full suite (mean over the $14$ rates).

\begin{table}[!htbp]
\centering
\caption{\textbf{Lite subset against the full suite.} Success (\%) over three seeds on all $90$ tasks (Full, solved trials of $270$ in grey) and on the $30$ lite tasks (Lite, solved trials of $90$ in grey). Gap: \priv minus \std, in points.}
\label{tab:lite}
\footnotesize
\setlength{\tabcolsep}{5pt}
\begin{fittab}\begin{tabular}{@{}l rr rr rr@{}}
\toprule
 & \multicolumn{2}{c}{Privileged} & \multicolumn{2}{c}{Standard} & \multicolumn{2}{c}{Gap} \\
\cmidrule(lr){2-3} \cmidrule(lr){4-5} \cmidrule(l){6-7}
Agent & Full & Lite & Full & Lite & Full & Lite \\
\midrule
GPT-6 Astra & $85.6$ \textcolor{black!50}{($231$)} & $85.6$ \textcolor{black!50}{($77$)} & $42.6$ \textcolor{black!50}{($115$)} & $43.3$ \textcolor{black!50}{($39$)} & $43.0$ & $42.2$ \\
GPT-6.1 Sol & $81.1$ \textcolor{black!50}{($219$)} & $81.1$ \textcolor{black!50}{($73$)} & $34.8$ \textcolor{black!50}{($94$)} & $34.4$ \textcolor{black!50}{($31$)} & $46.3$ & $46.7$ \\
Claude Opus 5.5 & $80.7$ \textcolor{black!50}{($218$)} & $80.0$ \textcolor{black!50}{($72$)} & $29.3$ \textcolor{black!50}{($79$)} & $27.8$ \textcolor{black!50}{($25$)} & $51.5$ & $52.2$ \\
Gemini 4 Argon & $73.3$ \textcolor{black!50}{($198$)} & $73.3$ \textcolor{black!50}{($66$)} & $30.7$ \textcolor{black!50}{($83$)} & $31.1$ \textcolor{black!50}{($28$)} & $42.6$ & $42.2$ \\
\addlinespace[2pt]
Kimi 3.1 Pro & $31.5$ \textcolor{black!50}{($85$)} & $32.2$ \textcolor{black!50}{($29$)} & $7.0$ \textcolor{black!50}{($19$)} & $7.8$ \textcolor{black!50}{($7$)} & $24.4$ & $24.4$ \\
DeepSeek v4.1 Flash & $10.7$ \textcolor{black!50}{($29$)} & $11.1$ \textcolor{black!50}{($10$)} & $5.6$ \textcolor{black!50}{($15$)} & $6.7$ \textcolor{black!50}{($6$)} & $5.2$ & $4.4$ \\
GLM 5.3 Flash & $4.4$ \textcolor{black!50}{($12$)} & $5.6$ \textcolor{black!50}{($5$)} & $4.1$ \textcolor{black!50}{($11$)} & $4.4$ \textcolor{black!50}{($4$)} & $0.4$ & $1.1$ \\
\bottomrule
\end{tabular}\end{fittab}
\end{table}

\begin{table}[!htbp]
\centering
\caption{\textbf{The $30$ tasks of the lite subset} by constraint type, with the number of tasks kept of the type's tasks in brackets.}
\label{tab:lite_tasks}
\scriptsize
\setlength{\tabcolsep}{3pt}
\begin{fittab}\begin{tabular}{@{}p{3.2cm} p{12.3cm}@{}}
\toprule
Type & Tasks \\
\midrule
Interlocking ($8$ of $23$) & \raggedright \texttt{cast-coil-05}, \texttt{cast-coil-06}, \texttt{cast-keyhole-03}, \texttt{cast-keyhole-05}, \texttt{horseshoe-ring-01}, \texttt{interlocked-claws-08}, \texttt{interlocked-claws-10}, \texttt{interlocked-claws-12} \tabularnewline
Narrow passage ($4$ of $13$) & \raggedright \texttt{buzz-wire-3d-02}, \texttt{hedgehog-in-cage-06}, \texttt{hedgehog-in-cage-07}, \texttt{lock-picking-04} \tabularnewline
Assembly ($7$ of $20$) & \raggedright \texttt{bed-bolt-03}, \texttt{bed-bolt-05}, \texttt{clock-assembling-01}, \texttt{ikea-ikea365}, \texttt{ikea-kyrre}, \texttt{ikea-stegesund}, \texttt{ikea-svallet} \tabularnewline
Untangling ($4$ of $13$) & \raggedright \texttt{knot-untying-05}, \texttt{knot-untying-07}, \texttt{orbits-01}, \texttt{slotted-post-ring-02} \tabularnewline
Threading ($4$ of $11$) & \raggedright \texttt{fiber-bundling-09}, \texttt{rope-threading-06}, \texttt{rope-threading-07}, \texttt{wire-splicing-01} \tabularnewline
Shape control ($3$ of $10$) & \raggedright \texttt{blanket-flatten-04}, \texttt{blanket-fold-01}, \texttt{blanket-fold-04} \tabularnewline
\bottomrule
\end{tabular}\end{fittab}
\end{table}

\subsection{How the agents solve tasks in the \priv regime}
\label{sec:suppl_privmethods}

We read what each agent did from the code and
the commands it wrote in all $1{,}890$ \priv runs ($992$ solved and $898$ failed), and from the session log of the task
tooling, which records every reset of the simulator. \Cref{fig:privmethods} shows the result per agent, and
\cref{tab:suppl_privmethods} gives the numbers.

\paragraph{Classes.}
We label each run by fixed rules on the code the agent wrote. A \emph{planner} run contains a
motion planner over configurations of a piece or of the arms: a sampling-based planner (RRT, RRT-Connect, a roadmap or
another random tree) or a graph search (A*, Dijkstra, breadth-first search, dynamic programming or beam search). A
\emph{simulator search} run has no planner, and it loops over at least three numeric candidates with one simulation each
or calls a global optimiser. Every other run is \emph{scripted}: the agent computes waypoints from the state and the
geometry, solves inverse kinematics for them and corrects the motion over a few trials. A \emph{rollout} is one stretch
of simulation between two resets. The tooling of the $12$ IKEA tasks writes no session log, so we count rollouts on the
other $78$ tasks, in $926$ of the $992$ solved runs and $704$ of the $898$ failed runs.

\paragraph{Check of the rules.}
We read $110$ runs by hand. On two samples of $42$ runs each, labelled by hand before the rules ran on them, the rules
give the hand label in $34$ and $36$ runs ($81\,\%$ and $86\,\%$), and in the second sample they agree with the hand
label on whether a run contains a planner in $40$ runs ($95\,\%$). A planner run is a run whose code contains a planner.
In a sample of $22$ solved planner runs the delivered motion follows the planner's path in $18$, partly in $3$ and not in
$1$.

\paragraph{Two thirds of the solved runs are scripted.}
Of the $992$ solved runs, $638$ ($64\,\%$) are scripted, $305$ ($31\,\%$) contain a planner and $49$ ($5\,\%$) are
simulator-search runs. Of the planner runs, $207$ contain a sampling-based planner and $98$ a graph search only. The four
proprietary agents differ little: their planner shares lie between $24\,\%$ and $37\,\%$ and their scripted shares
between $62\,\%$ and $66\,\%$.

\paragraph{Planners follow the constraint type.}
Of the $305$ solved planner runs, $290$ belong to the interlocking and narrow-passage tasks, where a rigid piece has to
find a free path through another. There $65\,\%$ of the $299$ and $69\,\%$ of the $140$ solved runs contain a planner.
By family the share is $89\,\%$ for cast coil ($58$ of $65$), $87\,\%$ for hedgehog in cage ($52$ of $60$), $80\,\%$ for
cast keyhole ($43$ of $54$), $75\,\%$ for interlocked claws ($77$ of $103$), $70\,\%$ for 3D buzz wire ($28$ of $40$) and
$42\,\%$ for lock picking ($17$ of $40$), where the planner is a graph search over a discrete model of the lock. In the
other four constraint types, $501$ of the $553$ solved runs ($91\,\%$) are scripted and $15$ ($3\,\%$) contain a
planner: the agents move the cloth, the rope, the fibers or the part with motions computed from the state, and read the
result back from the state.

\paragraph{The agents write their planners and their models.}
In $259$ of the $305$ solved planner runs every planner is the agent's own code. The other $46$ runs ($22$ of GPT-6.1
Sol, $20$ of GPT-6 Astra and $4$ of Claude Opus 5.5) also install the OMPL library~\cite{sucan2012ompl}.
The planner works on a model that the agent builds. In $284$ of the $305$ runs ($93\,\%$) the code sets up a geometric
model of its own from the task's meshes, with a collision library, signed distances or a second MuJoCo model, and the
agent plans the path of the free piece or of the arm on it. $34$ of the solved planner runs also sweep numeric
candidates on the simulator. Picking the piece up, following the path with the arms and letting it go are scripted in
the same run. A solved planner run takes a median of $8$ rollouts and $1{,}424$ program lines, a solved scripted run $7$
rollouts and $900$ lines.

\paragraph{A solved run takes few rollouts.}
A solved run takes a median of $7$ rollouts (interquartile range $5$ to $12$). Of the $926$ solved runs with a session
log, $301$ ($33\,\%$) take at most $5$ rollouts, $638$ ($69\,\%$) at most $10$ and $831$ ($90\,\%$) at most $20$, and
$9$ take more than $50$, up to $131$. GPT-6 Astra and GPT-6.1 Sol take a median of $5$ and $6$ rollouts, and none of
their solved runs takes more than $20$. Simulated time gives the same picture. A solved run simulates a median of $4.1$
times the length of the trajectory it delivers, replays for self-checking included, and a solved run of GPT-6 Astra
$3.0$ times. The session log counts resets through the task tooling. Besides these, $551$ of the solved runs
($56\,\%$) save and restore the simulator state without a reset, and $49$ ($5\,\%$) step a MuJoCo model of their own.

\paragraph{Failed runs search more.}
Of the $898$ failed runs, $250$ ($28\,\%$) contain a planner, $184$ ($20\,\%$) are simulator-search runs, $461$
($51\,\%$) are scripted and $3$ contain no motion code. A failed run takes a median of $18$ rollouts and $2{,}415$
program lines, against $7$ and $1{,}069$ for a solved run. For the four proprietary agents, $271$ of the $343$ runs with
a planner are solved ($79\,\%$), $556$ of the $649$ scripted runs ($86\,\%$) and $39$ of the $88$ simulator-search runs
($44\,\%$). Within the two rigid constraint types the success rates with and without a planner are close: the four
proprietary agents solve $166$ of $183$ interlocking runs with a planner ($91\,\%$) and $89$ of $93$ without
($96\,\%$), and $90$ of $109$ narrow-passage runs with a planner ($83\,\%$) and $37$ of $47$ without ($79\,\%$).

\begin{table}[!htbp]
\centering
\caption{\textbf{How \priv runs are solved.} Solved runs per agent and per constraint type, and all failed runs for comparison. Planner, Sim.\ search (simulator search) and Scripted are exclusive classes and sum to $100\,\%$ up to rounding, and Scripted includes the $3$ failed runs without motion code. Sampling is the part of the planner runs with a sampling-based planner. Rollouts is the median number of rollouts per run and $\le 10$ the share of runs with at most $10$, on the $78$ tasks whose tooling writes a session log ($926$ of the $992$ solved runs and $704$ of the $898$ failed runs). Simulated is the median of the control steps the agent simulated divided by the length of the trajectory it delivered, replays for self-checking included. Lines is the median number of program lines the agent wrote in a run.}
\label{tab:suppl_privmethods}
\footnotesize
\setlength{\tabcolsep}{5pt}
\begin{fittab}\begin{tabular}{@{}l r rrrr rrr r@{}}
\toprule
 & & \multicolumn{4}{c}{Method (\% of runs)} & \multicolumn{3}{c}{Simulator use} & \\
\cmidrule(lr){3-6}\cmidrule(lr){7-9}
 & Runs & Planner & Sampling & Sim.\ search & Scripted & Rollouts & $\le 10$ (\%) & Simulated & Lines \\
\midrule
\multicolumn{10}{@{}l}{\emph{Solved runs, by agent}} \\
Astra & $231$ & $33$ & $26$ & $1$ & $65$ & $5$ & $97$ & $3.0$ & $651$ \\
Sol & $219$ & $37$ & $32$ & $1$ & $62$ & $6$ & $92$ & $3.2$ & $753$ \\
Opus 5.5 & $218$ & $30$ & $17$ & $4$ & $66$ & $9$ & $56$ & $6.1$ & $1{,}292$ \\
Argon & $198$ & $24$ & $7$ & $12$ & $64$ & $10$ & $52$ & $6.2$ & $2{,}022$ \\
Kimi & $85$ & $31$ & $26$ & $6$ & $64$ & $11$ & $45$ & $6.0$ & $1{,}303$ \\
DeepSeek & $29$ & $21$ & $17$ & $17$ & $62$ & $21$ & $7$ & $13.8$ & $1{,}943$ \\
GLM & $12$ & $17$ & $8$ & $0$ & $83$ & $17.5$ & $33$ & $7.2$ & $1{,}704$ \\
\midrule
\multicolumn{10}{@{}l}{\emph{Solved runs, by constraint type}} \\
Interlocking & $299$ & $65$ & $40$ & $3$ & $33$ & $8$ & $59$ & $5.5$ & $1{,}202$ \\
Narrow passage & $140$ & $69$ & $53$ & $3$ & $28$ & $8$ & $69$ & $4.4$ & $1{,}402$ \\
Assembly & $173$ & $5$ & $5$ & $8$ & $87$ & $8$ & $58$ & $3.3$ & $1{,}132$ \\
Untangling & $124$ & $1$ & $0$ & $6$ & $94$ & $6$ & $76$ & $4.3$ & $710$ \\
Threading & $110$ & $5$ & $4$ & $5$ & $90$ & $5$ & $85$ & $3.0$ & $1{,}164$ \\
Shape control & $146$ & $0$ & $0$ & $7$ & $93$ & $6.5$ & $79$ & $4.3$ & $948$ \\
\midrule
All solved runs & $992$ & $31$ & $21$ & $5$ & $64$ & $7$ & $69$ & $4.1$ & $1{,}069$ \\
All failed runs & $898$ & $28$ & $21$ & $20$ & $52$ & $18$ & $32$ & $11.3$ & $2{,}415$ \\
\bottomrule
\end{tabular}\end{fittab}
\end{table}

\paragraph{Searches for reference files.}
About $27$ runs search the container for reference or solution files ($11$ of DeepSeek v4.1 Flash, $9$ of GLM 5.3
Flash, $6$ of Claude Opus 5.5 and $1$ of Gemini 4 Argon), $4$ of them solved runs of Claude Opus 5.5. None finds one,
as the reference solutions are mounted for the reference runner only (\cref{sec:suppl_validity}).

\paragraph{The session-log flag.}
The verifier sets a flag on a run when a monitor outside the agent's container registers that the session log in the
agent's output directory was removed or rewritten, or when the collected log differs from the monitor's own copy. The
flag is a record. The verdict comes from the verifier's two replays of the delivered trajectory,
and the session log feeds five self-reported fields (sessions, resets, simulated steps, recordings and replays).
The flag is set in $194$ of the $1{,}890$ \priv runs, $123$ of them solved. In $104$ of the $194$ runs the agent does
nothing to the log: in $99$ the monitor registers the removal $10.8$ to $19.4$\,s after the agent has finished, while
the container stops, and in $5$ the log is collected while a process still appends to it. In the other $90$ runs the
agent removes the log while it tidies its output directory. Of these, $31$ runs clean the directory in bulk and keep a
list of deliverables ($20$ of Kimi 3.1 Pro, $6$ of GPT-6 Astra and $5$ of GPT-6.1 Sol), and $59$ runs name the log in
the command ($56$ of Kimi 3.1 Pro and $1$ each of GPT-6 Astra, DeepSeek v4.1 Flash and GLM 5.3 Flash), with $58$
deletions and $1$ move. No run edits the content of the log. The monitor's copy of the log exists for all $194$ runs,
and the verifier takes the five fields of all $194$ from that copy. The verdict fields of the $123$ solved flagged runs equal those of the other $803$ solved runs outside IKEA:
both replays succeed on one state hash, and no review is pending.

\subsection{A classical planner on the rigid puzzles}
\label{sec:suppl_planner}

The spatial constraints of \name are defined for the robot that executes the motion with its grasp. We run a classical
sampling-based planner on the $36$ puzzles of the two
rigid constraint types, interlocking and narrow passage (\cref{tab:tasks}), in three stages: a route for the free piece,
the execution of a planned route by the robot, and the verifier's grade of that execution.

\paragraph{The free-piece problem.}
The free-piece problem of a puzzle takes the task's simulation model with its collision geometry, removes the robot and
lets the loose piece move as a free body in $\mathrm{SE}(3)$, from the frozen start state to the piece's pose in the
final state of the reference solution. It has no gripper, no arm, no gravity and no contact dynamics, so a solution is a
collision-free route of the piece. In the families with two loose pieces (interlocked claws, cast keyhole and cast coil)
one piece moves relative to the other and the table is left out, since the robot holds both pieces, and in the
three-piece interlocked-claws task two pieces move relative to the third. In the horseshoe ring the ring, one horseshoe
and the two chain links move relative to the other horseshoe. In the 3D buzz wire every configuration between the two
ends of the wire keeps the wire inside the loop, as the task's rule demands. In lock picking the angles of the rotor
plates are free coordinates within their stops, which relaxes the task, where a rotor turns only when the key pushes it.

\paragraph{Planners.}
We run RRT-Connect~\cite{kuffner2000rrtconnect} and a probabilistic roadmap~\cite{kavraki1996prm} with obstacle-based
sampling~\cite{amato1998obprm} from OMPL 2.0.1~\cite{sucan2012ompl}, with its default settings except a range of
$0.03$\,m for RRT-Connect, each run on one CPU core. Collisions are checked with MuJoCo's own collision detection on the
task's collision shapes every $0.5$\,mm of piece motion. A configuration is valid when no contact of a moving piece is
deeper than $0.1$\,mm, and every route is checked again every $0.1$\,mm. Time is the CPU time to the first route. Each
planner runs $5$ seeds of $10$ minutes per puzzle, and a puzzle counts as solved within $1$ or $10$ minutes when at least
$3$ of its $5$ seeds find a route within that time. RRT-Connect also runs $2$ seeds of $60$ minutes on the puzzles that
it does not solve within $10$ minutes, and such a puzzle counts at $60$ minutes when both seeds find a route. These are
$404$ planning runs in all.

\paragraph{Execution by the robot.}
For each puzzle with a route, a fixed script lets the robot execute planned routes. The reference solution's own rows
bring the robot to the reference solution's last change of grasp, and RRT-Connect plans the piece's route again from
that state, with $10$ minutes per route and one route per seed ($5$ seeds, $3$ in the 3D buzz wire). The tool frame
follows the route by inverse kinematics within the joint limits, at waypoints $2$\,mm apart and at most $1$\,cm/s. The
robot's own collision geometry is checked along the route, where a hand's contacts with the piece it holds do not
count, and the joint trajectory runs once through the verifier's replay, which grades it with the task's success rule.
In a two-arm puzzle only the arm that holds the moving piece follows the route, and the other arm holds its piece still
at its pose of the grasp row. A route is executable when inverse kinematics follows it within $2$\,mm and $0.02$\,rad and
the robot stays free of collisions. As a control, the reference solution's own piece routes run through the same script
with both arms following, and a puzzle is assessed only when its control passes every stage. This holds for $12$ of the
$25$ controls.

\paragraph{Results.}
RRT-Connect solves $9$ of the $36$ free-piece problems within $1$ minute, $13$ within $10$ minutes and $14$ within $60$
minutes (\cref{tab:planner_free}). The roadmap planner solves $0$ within $1$ minute and $4$ within $10$ minutes. For $17$
puzzles at least one run of either planner finds a route. The $9$ puzzles that RRT-Connect solves within a minute are
buzz-wire-3d-02, -06 and -07, cast-keyhole-02, hedgehog-in-cage-02 and interlocked-claws-02, -08, -09 and -10, and
RRT-Connect solves all $5$ buzz-wire problems within $10$ minutes. No run finds a route for the other $19$ puzzles. For
$14$ of them the reference solution's own piece path is a valid route of the free-piece problem.

Of the $17$ puzzles with a route, $9$ can be assessed by the robot script (\cref{tab:planner_robot}). On $4$ of the others
(cast-keyhole-02, cast-coil-06, buzz-wire-3d-02 and interlocked-claws-08) the control passes inverse kinematics and the
collision check, but its rollout is not graded as success. On the $4$ horseshoe-ring puzzles the free-piece route moves
four loose bodies, while the robot has two hands. On the $9$ assessed puzzles RRT-Connect finds $31$ of $37$ routes from
the grasp, inverse kinematics follows $7$ of them, and none is executable. The other $24$ routes leave the reach or the
joint limits of the arm, with a largest error of the tool frame between $2$ and $480$\,mm per route. The $7$ routes that
inverse kinematics follows drive the robot $11$ to $96$\,mm deep into the puzzle ($5$ routes) or the table ($2$ routes).
The verifier grades $1$ of the $31$ executions as success, a route of interlocked-claws-09. This route fails the
geometric collision check at $8$ of its $812$ waypoints, where the moving arm overlaps the piece in the other hand, which
the check keeps at its pose of the grasp row, by up to $28$\,mm.

A sampling-based planner thus finds a route for the free piece on $17$ of the $36$ puzzles, $9$ of them within a minute,
and none of the $31$ routes planned from the reference solution's grasp is executable by the robot. The hard part of
these puzzles is turning a feasible route of the piece into a motion that the robot can carry out. In their \priv
trials the agents solve the whole task with the robot from the task text and deliver one trajectory. On the $19$ puzzles for which no planner run finds a route, the four proprietary agents succeed
in $199$ of their $228$ \priv trials.

\begin{table}[!htbp]
\centering
\caption{\textbf{A classical planner on the free-piece problems.} Puzzles per family that RRT-Connect solves in at least
$3$ of $5$ seeds within $1$ and $10$ minutes of single-core time. A puzzle also counts at $60$ minutes when both of its $2$
seeds with $60$ minutes find a route. Time: median time of RRT-Connect to the first route in seconds, over the family's
puzzles whose median seed finds one. PRM: the probabilistic roadmap with obstacle-based sampling, counted as RRT-Connect
within $1$ and $10$ minutes. Any run: puzzles with a route in at least one run of either planner. DoF: degrees of freedom
of the free-piece problem. $^{\dagger}$Relaxed problem with free rotor angles.}
\label{tab:planner_free}
\footnotesize
\setlength{\tabcolsep}{5pt}
\begin{fittab}\begin{tabular}{@{}l r r ccc r cc r@{}}
\toprule
 & & & \multicolumn{3}{c}{RRT-Connect} & & \multicolumn{2}{c}{PRM} & \\
\cmidrule(lr){4-6} \cmidrule(lr){8-9}
Family & Puzzles & DoF & $1$\,min & $10$\,min & $60$\,min & Time (s) & $1$\,min & $10$\,min & Any run \\
\midrule
Interlocked claws & $8$ & $6/12$ & $4$ & $4$ & $4$ & $3.3$ & $0$ & $1$ & $5$ \\
Cast keyhole & $5$ & $6$ & $1$ & $1$ & $1$ & $58$ & $0$ & $0$ & $1$ \\
Cast coil & $5$ & $6$ & $0$ & $1$ & $1$ & $583$ & $0$ & $0$ & $1$ \\
Horseshoe ring & $5$ & $24$ & $0$ & $1$ & $2$ & $236$ & $0$ & $0$ & $4$ \\
\addlinespace[2pt]
Hedgehog in cage & $5$ & $6/7$ & $1$ & $1$ & $1$ & $0.7$ & $0$ & $1$ & $1$ \\
3D buzz wire & $5$ & $6$ & $3$ & $5$ & $5$ & $11$ & $0$ & $2$ & $5$ \\
Lock picking$^{\dagger}$ & $3$ & $9/10$ & $0$ & $0$ & $0$ & -- & $0$ & $0$ & $0$ \\
\midrule
All & $36$ & & $9$ & $13$ & $14$ & & $0$ & $4$ & $17$ \\
\bottomrule
\end{tabular}\end{fittab}
\end{table}

\begin{table}[!htbp]
\centering
\caption{\textbf{From a route of the free piece to an execution by the robot.} Of these puzzles: Free piece, those with a
route in at least one run of either planner. Assessed, puzzles among these on which the robot script reproduces the reference
solution, so that its control passes every stage. Executable, assessed puzzles with a planned route that the robot
follows by inverse kinematics without a collision. Graded, assessed puzzles with a planned route whose execution the
verifier grades as success. Routes: the routes planned from the reference solution's grasp on the assessed puzzles,
those found, those followed by inverse kinematics, those also free of collisions, and those whose execution the
verifier grades as success. --: no puzzle assessed.}
\label{tab:planner_robot}
\footnotesize
\setlength{\tabcolsep}{5pt}
\begin{fittab}\begin{tabular}{@{}l r rrrr rrrrr@{}}
\toprule
 & & \multicolumn{4}{c}{Of these puzzles} & \multicolumn{5}{c}{Routes} \\
\cmidrule(lr){3-6} \cmidrule(l){7-11}
Family & Puzzles & Free piece & Assessed & Executable & Graded & Planned & Found & IK & Collision-free & Graded \\
\midrule
Interlocked claws & $8$ & $5$ & $4$ & $0$ & $1$ & $20$ & $14$ & $4$ & $0$ & $1$ \\
Cast keyhole & $5$ & $1$ & $0$ & -- & -- & -- & -- & -- & -- & -- \\
Cast coil & $5$ & $1$ & $0$ & -- & -- & -- & -- & -- & -- & -- \\
Horseshoe ring & $5$ & $4$ & $0$ & -- & -- & -- & -- & -- & -- & -- \\
\addlinespace[2pt]
Hedgehog in cage & $5$ & $1$ & $1$ & $0$ & $0$ & $5$ & $5$ & $3$ & $0$ & $0$ \\
3D buzz wire & $5$ & $5$ & $4$ & $0$ & $0$ & $12$ & $12$ & $0$ & $0$ & $0$ \\
Lock picking & $3$ & $0$ & $0$ & -- & -- & -- & -- & -- & -- & -- \\
\midrule
All & $36$ & $17$ & $9$ & $0$ & $1$ & $37$ & $31$ & $7$ & $0$ & $1$ \\
\bottomrule
\end{tabular}\end{fittab}
\end{table}

\paragraph{Inputs of the baseline.}
The planners are given the exact geometry, the frozen start state and the goal configuration of the piece, taken from
the final state of the reference solution. In the 3D buzz wire they are also given two intermediate configurations
from the reference solution, with the loop on the wire within $10$\,mm of each end. The execution script is also given
the reference solution's rows up to its final grasp and that grasp, and in the 3D buzz wire the reference solution has
traversed the wire before its final grasp. The first stage plans for the piece alone, and the second stage is a fixed
script with one grasp per piece, in which the arm follows the piece's route by inverse kinematics.

\subsection{Tokens}
\label{sec:suppl_more}

\begin{figure}[!htbp]
  \centering
  \includegraphics[width=0.55\linewidth]{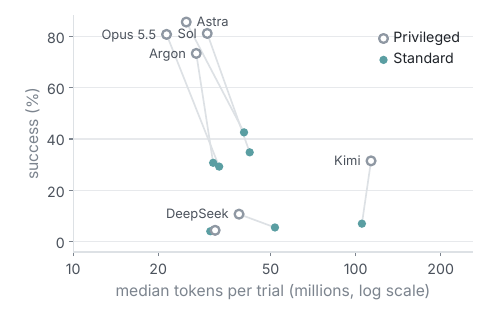}
  \caption{\textbf{Tokens against success.} Median tokens per trial (input including cached, plus output) and success, one ring (\priv) and one dot (\std) per agent. }
  \label{fig:time}
\end{figure}

Token use does not track success (\cref{fig:time}). In the \priv regime Claude Opus 5.5 uses the
fewest tokens per trial ($21$\,M) and GPT-6 Astra $25$\,M, while Kimi 3.1 Pro uses about $113$\,M and solves $31\%$. Every
proprietary agent uses more tokens in the \std regime, where it must look before it acts.

\subsection{Compute and cost of an evaluation}
\label{sec:suppl_cost}

\begin{figure}[!htbp]
  \centering
  \includegraphics[width=0.6\linewidth]{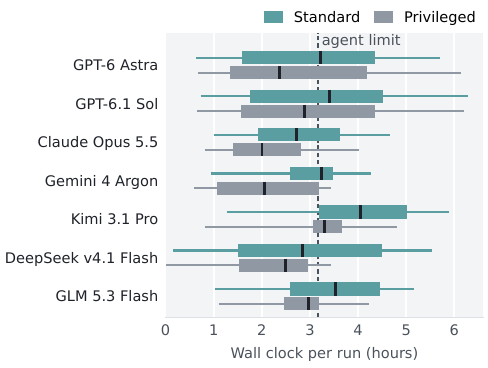}
  \caption{\textbf{Wall clock per run.} From the start of the trial's environment to the grader's verdict, over the
  $3{,}761$ runs of the seven agents. Boxes span the middle half of an agent's runs in one regime, whiskers the $5$th to
  $95$th percentile, and the black bar is the median. The dashed line is the agent's limit of $11{,}400$\,s ($10{,}800$\,s for DeepSeek v4.1 Flash and GLM 5.3 Flash, \cref{sec:suppl_agents}). Set-up and
  grading come on top of it.}
  \label{fig:walltime}
\end{figure}

These numbers cover the $3{,}761$ runs of the seven agents in both regimes that were frozen for analysis: the $3{,}780$ scored trials of the main tables less $19$ (DeepSeek $9$, GLM $9$, Sol $1$) that were rerun later (\cref{sec:suppl_endings}).

\paragraph{Time per run.}
A run takes a median of $2.7$\,h of wall clock in the \priv regime (middle half $1.6$--$3.3$\,h) and $3.3$\,h in the
\std regime ($2.1$--$4.3$\,h), from the start of its environment to the grader's verdict (\cref{fig:walltime}). This
spans the environment, the harness installation, the agent's session under the limit of \cref{sec:suppl_agents} and the
grading. A puzzle run is graded by replaying its recorded actions in a fresh process that also renders the video. The
replay takes a median of $15$\,min in the \priv regime and $41$\,min in the \std regime, and it makes up $25\,\%$ of the
puzzle runs' wall clock.

\paragraph{Hardware per trial.}
A trial reserves the CPUs and memory that its task declares. A \priv puzzle trial holds $16$ CPU cores and $30$ or
$40$\,GB ($8$ cores for the blanket tasks), a \std puzzle trial $8$ cores and $24$\,GB over the agent's container and the
simulator service ($12$ cores and $28$\,GB for the blanket tasks), and an IKEA trial $4$ cores and $10$\,GB (\priv) or
$8$ cores and $18$\,GB (\std). The agent's container peaks at a median of $1.1$\,GB of memory. The runs used NVIDIA RTX
A6000 ($2{,}015$ runs), L4 ($1{,}308$) and L40 ($186$) GPUs, and the $252$ IKEA \priv runs used none.

\paragraph{Total compute.}
The $3{,}761$ runs took $11{,}094$ trial-hours ($5{,}013$ in \priv, $6{,}082$ in \std), from $1{,}349$ for Claude Opus
5.5 to $1{,}954$ for Kimi 3.1 Pro, and reserved $119{,}139$ CPU-core-hours. The grader's replays account for
$2{,}356$ of the trial-hours. Tokens per run are in \cref{fig:time}.

\FloatBarrier
\section{Real-robot experiment}
\label{sec:suppl_real}

\paragraph{Setup.} The robot is a single YAM arm. Three Intel RealSense cameras watch the scene, two D405 and one D435i.
The agent controls the arm in absolute end-effector mode: each command gives a target pose of the end effector, not a step
from the current one. The interface follows the \std regime. The agent sees the camera images and the arm's own state, is
given no model or position of the wire, and plays one live episode per trial.

\paragraph{Task.} The gripper holds a loop around a bent wire, and the agent moves the loop along the wire without touching
it. A touch closes a circuit that lights a lamp and sounds a buzzer, and ends the trial. Progress is the share of the wire
covered before the first touch, $100\,\%$ when the loop reaches the end. Each agent plays eight consecutive trials.

\begin{table}[!htbp]
\centering
\caption{\textbf{Real buzz wire, progress (\%) per trial.} Values rounded to $5$ points, means over trials $1$--$4$ and
$5$--$8$.}
\label{tab:real}
\small
\setlength{\tabcolsep}{2.5pt}
\begin{fittab}\begin{tabular}{@{}l *{8}{r} r r@{}}
\toprule
Agent & 1 & 2 & 3 & 4 & 5 & 6 & 7 & 8 & 1--4 & 5--8 \\
\midrule
GPT-6 Astra & $60$ & $60$ & $80$ & $60$ & $85$ & $90$ & $100$ & $100$ & $65.0$ & $93.8$ \\
Claude Opus 5.5 & $0$ & $50$ & $50$ & $40$ & $50$ & $80$ & $40$ & $80$ & $35.0$ & $62.5$ \\
\bottomrule
\end{tabular}\end{fittab}
\end{table}

\section{Failure analysis}
\label{sec:suppl_failures}

\begin{table}[!tbp]
\centering
\caption{\textbf{Units and terms of the failure analysis.} Run counts are over the $3{,}609$ kept runs. Counts of labelled runs are samples, and proportions are weighted to the $3{,}609$.}
\label{tab:suppl_units}
\scriptsize
\setlength{\tabcolsep}{3pt}
\renewcommand{\arraystretch}{1.15}
\begin{fittab}\begin{tabular}{@{}>{\raggedright\arraybackslash}p{2.15cm} p{10.4cm} >{\raggedright\arraybackslash}p{2.9cm}@{}}
\toprule
Name & Definition & Count \\
\midrule
Per run & Each run counts once. Used for the approach at the key step & $902$ runs with attempts at the key step \\
Per sampled step & One step in each sampled run counts once. Used for the approach at all steps & $301$ steps, $43$ per agent \\
Per failed run & Each failed run counts once. Used for endings, reasons for giving up, stuck runs and the causes from the replay & $2{,}214$ \\
Per failed attempt & Each attempt that does not reach its step's goal counts once, in failed and solved runs. Used for the causes from the transcript & $25{,}469$ \\
Per action & Each step counts once and is the denominator. Used for the stop rate. The steps are split by regime and compared within each regime: $855$ key and $1{,}999$ other steps in the \std regime, $434$ and $1{,}192$ in the \priv regime, and $41$ other steps fall in cells with no key step. The solved rate uses $2{,}693$ key and $7{,}291$ other steps & $1{,}289$ key steps, $3{,}232$ other steps \\
Per family & Each of the $19$ puzzle families counts once, with shares over the runs inside it. Used for the stage with the largest drop & $19$ \\
\midrule
Key step & The step at which the spatial constraint named by the task is passed. After it only finishing remains. When the constraint repeats, the first time. Set from the task design before any data (\cref{tab:suppl_keysteps}) & $20$ families \\
Stop step & In a failed run, the first step that is reached but not done when the run ends (from the replay). Among reached steps without an order, the last one the robot worked on. The stop rate is the share of reached steps in failed runs that are the stop step. Key steps are compared with the other steps of the same action type in each cell of agent and action type, and the cells are weighted by their key steps. The other nine of the $2{,}214$ failed runs finish every reached step and have no stop step & $2{,}205$ failed runs \\
Stuck step & A step that takes at least a third of the run's wall clock, reading, thinking and coding included, and has at least $3$ failed attempts. A failed run is stuck when its stop step is a stuck step. Only five agents log the timestamps this needs: GPT-6 Astra, GPT-6.1 Sol, Claude Opus 5.5, Kimi 3.1 Pro and GLM 5.3 Flash & $486$ failed runs \\
Stage & Consecutive steps with no order among them. Stages are passed in order & -- \\
Drop & The number of runs that pass all earlier stages in a row and do not complete this stage & -- \\
\bottomrule
\end{tabular}\end{fittab}
\end{table}

\begin{table}[!t]
\centering
\caption{\textbf{Labels of the failure analysis.} Causes are labelled from the transcript (per failed attempt) and from
the replay (per failed run). Endings and reasons for giving up are per failed run, approaches per run at the key step and per sampled step at all steps. The cause is set by the commands, and each label names what happens in the attempt.}
\label{tab:suppl_labels}
\scriptsize
\setlength{\tabcolsep}{3pt}
\renewcommand{\arraystretch}{1.15}
\begin{fittab}\begin{tabular}{@{}>{\raggedright\arraybackslash}p{2.25cm} p{6.65cm} p{6.6cm}@{}}
\toprule
Cause & From the transcript, per failed attempt & From the replay, per failed run \\
\midrule
Perception & \std regime only: the agent's estimate of an object's position, orientation or size is wrong & \std regime only: the position that the agent reports for an object is more than $1$\,cm from the true one \\
Planning & The robot runs the code as written, but the commands cannot reach the step's goal: a wrong target or order, a missing step, an unreachable pose or a collision & Every joint follows its command (error at most $0.05$\,rad in every segment), yet the goal is not reached. A command at a joint limit marks an unreachable target \\
Execution & The commands are right, but the robot or the object does not follow them: it is blocked, slips, lags or stops midway & The robot does not follow its commands, tested in this order: blocked (a joint torque at its limit, logged in the \std regime only), slipped (the object leaves a closed gripper), lagging (a command error above $0.05$\,rad) \\
Code error & The attempt does not end normally: an exception in the control code, a call over its $10$\,s limit or a failed tool call. It concerns one attempt, not the run & -- \\
Unclear & The agent retries the step and does not state what went wrong & No physical cause to judge: an API or infrastructure error, an unexplained agent exit, or Budget exhausted with fewer than $10$ active segments on the stop step. Outside the shares \\
\midrule
Ending & \multicolumn{2}{l@{}}{Rule} \\
\midrule
Budget exhausted & \multicolumn{2}{p{13.45cm}@{}}{Harbor stops the agent at its time limit, the model reaches a token limit, or the run ends without an exception at that time limit: $11{,}400$\,s ($190$ minutes) of agent session in Harbor's \texttt{agent\_execution} phase, whose clock starts after the harness is installed ($10{,}800$\,s for DeepSeek v4.1 Flash and GLM 5.3 Flash, $21{,}600$\,s for wire splicing, \cref{sec:suppl_agents})} \\
Gave up & \multicolumn{2}{p{13.45cm}@{}}{The agent stops by itself and says that it cannot finish, or turns to other work and does not return} \\
Submitted & \multicolumn{2}{p{13.45cm}@{}}{The agent stops by itself and hands in a result that it calls partial, unverified or its best effort, or a deliverable that the task does not ask for} \\
Claimed done & \multicolumn{2}{p{13.45cm}@{}}{The agent stops by itself and hands in the right kind of result as solved, but the grader fails it} \\
\midrule
Reason for giving up & \multicolumn{2}{l@{}}{Rule} \\
\midrule
Time running out & \multicolumn{2}{p{13.45cm}@{}}{The agent says that the time left is too short to try again} \\
No new approach & \multicolumn{2}{p{13.45cm}@{}}{The agent has tried several ways, says that they all failed and sees no other way, but does not call the task impossible} \\
Judged infeasible & \multicolumn{2}{p{13.45cm}@{}}{The agent says plainly that it cannot do the task or that the task cannot be done as given} \\
\midrule
Approach & \multicolumn{2}{l@{}}{Rule, at the labelled step} \\
\midrule
Plan first & \multicolumn{2}{p{13.45cm}@{}}{Before its first attempt, the agent computes a clearance, path or pose with numbers or explicit geometry and uses the result, whatever it does after a miss} \\
Replan on failure & \multicolumn{2}{p{13.45cm}@{}}{No computation before the first attempt. After a miss, the agent measures positions and orientations again or changes the path or pose} \\
Try directly & \multicolumn{2}{p{13.45cm}@{}}{No computation before the first attempt. After a miss, the agent only changes a step size, a force or a threshold, or it succeeds without a miss} \\
\bottomrule
\end{tabular}\end{fittab}
\end{table}

\subsection{How the runs are analysed}
\label{sec:suppl_endings}

\Cref{sec:endings} reads the runs of the seven agents with two sets of labels and six units. This section defines the units
and terms (\cref{tab:suppl_units}), the labels (\cref{tab:suppl_labels}) and the key step of each family
(\cref{tab:suppl_keysteps}).

\paragraph{Runs.}
The analysis starts from the $3{,}761$ runs frozen for analysis: the $3{,}780$ scored trials of the main tables less $19$
that were rerun later (DeepSeek $9$, GLM $9$, Sol $1$). We replay $3{,}618$ of them. The other $143$ cannot be replayed:
$89$ hand in no action file, $47$ IKEA replays did not finish and $7$ action files are invalid. We keep $3{,}609$ of the
replayed runs, $2{,}214$ failed and $1{,}395$ solved (Checks below). Each has
its agent actions stored, and its replay reproduces the recorded end state and outcome. The tables count $2{,}372$ failed
and $1{,}408$ solved trials. The difference comes from the $19$ later reruns, the $152$ runs without a usable replay and
$5$ seeds that the tables score differently. Of the $152$ runs without a usable replay, $147$ failed, so the kept
runs solve $38.7\,\%$ against $37.2\,\%$ of the $3{,}761$ runs. Run counts in \cref{sec:endings}
and in this section are over the $3{,}609$ kept runs. Counts of labelled runs and attempts are
samples, and proportions are weighted to the $3{,}609$.

\paragraph{Steps.}
We split each family into steps, one sub-goal each, and give each step one of six action types: grasp, move into place,
insert or thread, tighten or turn, separate or untangle, and finish. A step is reached when all the steps it needs are done. Over all runs, a reached key step is solved in $68\,\%$ of cases and another step of its type in $74\,\%$ ($52\,\%$ against $63\,\%$ in the \std regime, $82\,\%$ against $80\,\%$ in the \priv regime).

By agent in the \std regime, the key-step gap in solved rate is $-13$ points ($-20$ to $-6$) for Gemini 4 Argon and $-6$, $-7$ and $-6$ for GPT-6 Astra, Claude Opus 5.5 and GPT-6.1 Sol (intervals $-13$ to $+1$, $-15$ to $+1$ and $-13$ to $+1$, pooled $-6$ ($-10$ to $-3$)), $-8$ ($-12$ to $-5$) for the four proprietary agents and $-17$ ($-22$ to $-12$) for the three open-weight agents. By action type it is $-23$ for insert or thread, $-11$ for separate or untangle, $-3$ for move into place ($-10$ to $+3$) and $+20$ for tighten or turn. In $8/19$ families, the stage that loses the most runs holds the key step, against $3.8$ expected by chance.

\begin{table}[!htbp]
\centering
\caption{\textbf{The key step of each family}, set from the task design.}
\label{tab:suppl_keysteps}
\scriptsize
\setlength{\tabcolsep}{3pt}
\renewcommand{\arraystretch}{1.15}
\begin{minipage}[t]{0.49\linewidth}\centering
\begin{tabular}[t]{@{}l p{4.6cm}@{}}
\toprule
Family & Key step \\
\midrule
Interlocked claws & Pieces separated (all three in three-piece tasks) \\
Cast keyhole & Plates separated \\
Cast coil & Pieces nearly assembled, past the tight section \\
Horseshoe ring & Ring free of the horseshoes \\
\midrule
Hedgehog in cage & Centre of the object out of the cage \\
3D buzz wire & Full length of the wire passed \\
Lock picking & Last rotor plate turned to its slot \\
\midrule
Bed bolt & Thread caught \\
Clock assembling & Bridge placed over all pivots \\
Battery replacement & First new cell seated (slot A) \\
Caliper sorting & First sleeve in its bin \\
IKEA furniture & First insertion or turn, such as a side panel slid in or the first leg screwed on \\
\bottomrule
\end{tabular}
\end{minipage}\hfill
\begin{minipage}[t]{0.49\linewidth}\centering
\begin{tabular}[t]{@{}l p{4.6cm}@{}}
\toprule
Family & Key step \\
\midrule
Knot untying & All crossings undone \\
Slotted post ring & Ring lifted over the post top \\
Orbits & Cord off the post assembly \\
\midrule
Rope threading & First hole passed (wall hole G1) \\
Fiber bundling & All fibers through the ring \\
Wire splicing & Blue wire through the eye \\
\midrule
Blanket flatten & Folded layer turned back (in some tasks: half fold opened) \\
Blanket fold & Half fold in place \\
\bottomrule
\end{tabular}
\end{minipage}
\end{table}

\begin{figure}[!htbp]
  \centering
  \includegraphics[width=0.6\linewidth]{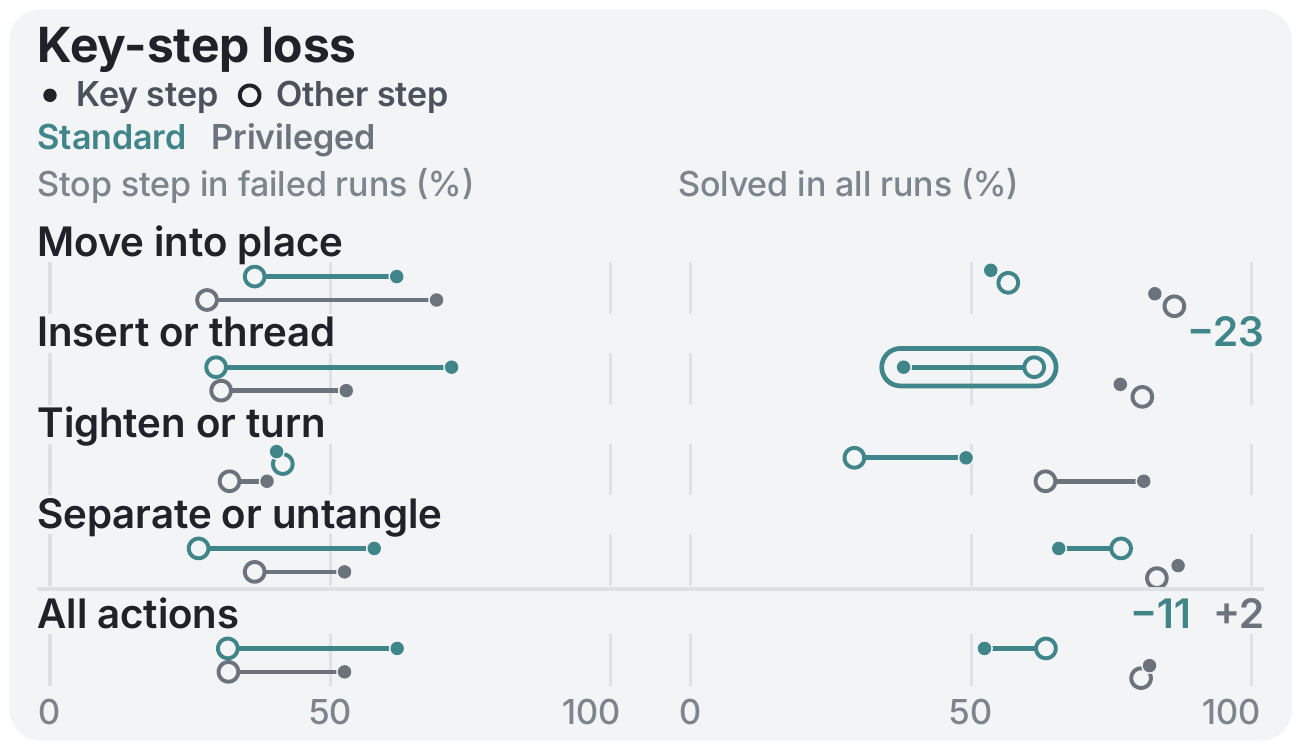}
  \caption{\textbf{Key step against the other steps of the same action.} Left: share of the reached steps that are the stop step in failed runs ($2{,}205$ failed runs with a stop step). Right: share solved, over all $3{,}609$ runs of the seven agents. Filled dot: key step. Hollow ring: other steps of the same action type. Colour: regime. Numbers: solved rate at the key step minus the other steps, in points. The ring around Insert or thread marks the hardest row.}
  \label{fig:suppl_keystep}
\end{figure}

\paragraph{Key step and stop rate.}
The \emph{key step} passes the constraint a task is named for (\cref{tab:suppl_keysteps}). A reached key step (its prerequisites done, \cref{fig:suppl_keystep}) is solved in $52\,\%$ in the \std regime and $82\,\%$ in the \priv regime, another step of its action type in $63\,\%$ and $80\,\%$.

A failed run's \emph{stop step} is its first reached step left unfinished, and the \emph{stop rate} is the share of reached steps that are the stop step. It is $62\,\%$ (\std) and $52\,\%$ (\priv) for key steps and $32\,\%$ for other
steps of the same agent and action type, and $757$ of the $2{,}205$ failed runs with a stop step ($34\,\%$) stop at the key step. Under this stop-step rule the gap is
$+30$ points in the \std regime and $+21$ in the \priv regime, where key steps are solved as often as others ($82\,\%$ against $80\,\%$): the stop rate alone shows no harder step.

Step order explains part. A second rule calls the lowest-numbered unfinished step the stop step, so that only the order of the steps decides it. Under this rule the real gap is $+36$ points in the \std regime and $+24$ in the \priv regime. To see what the order alone gives, we simulate every run on the real step dependencies with each step finished independently at the rate of the other steps of its agent and action type, so that key steps are no harder than other steps. The simulated gap is $+13$ points in the \std regime (range over $10$ simulations $11$ to $15$) and $+12$ in the \priv regime ($6$ to $16$). Only in the \std regime does the rest come with a lower solved rate. The \std solved gap is larger for open-weight agents ($-17$, proprietary $-8$) and insert steps ($-23$).

\paragraph{Two sets of labels.}
The \emph{replay labels} come from replaying the robot actions of a run in the simulator and checking the sub-goal of
each step, segment by segment, on the joints, the gripper and the objects. A segment is one agent call that moves the
robot in the \std regime and $1$\,s of simulated time in the \priv regime.

For the \emph{transcript labels} we read the transcript of a run once and list the attempts: the step that each
attempt works on, whether the attempt reaches the step's goal, and the cause when it does not. An attempt fails when
the next attempt works on the same step again. We label $1{,}210$ runs ($709$ failed, $501$ solved), at least one in
each of the $426$ cells of agent, regime, family and outcome, and weight each labelled run so that its cell counts with
all its kept runs. Intervals are $95\,\%$ intervals from $2{,}000$ resamples of runs.

\paragraph{Causes.}
Each failed attempt gets a cause from the transcript: $25{,}469$ attempts in $1{,}200$ of the $1{,}210$ labelled runs,
failed and solved ($9{,}801$ \priv and $15{,}668$ \std, $20{,}384$ in failed and $5{,}085$ in solved runs). The other
$10$ runs have no attempt to count. The shares are weighted to the kept runs. A \std run has $26$ failed attempts, a
\priv run $17$ ($+9$, $95\,\%$ interval $+7$ to $+12$), but \std runs also fail more often ($78\,\%$ against $45\,\%$):
per failed run, $30$ against $26$ ($-0.03$ to $+7$).

Per failed run, Planning is lower in the \std regime ($7.7$ against $9.7$, $-3.6$ to $-0.4$), Execution has no clear
gap ($+0.9$, $-0.9$ to $+2.6$), Unclear is higher ($5.4$ against $1.9$, $+2.4$ to $+4.7$) and Perception is $+1.4$
($5\,\%$ of \std attempts). A label names what happens in the attempt: a clearance misjudged without comment is
Unclear, and a grasp point that never holds is Execution. In the \priv regime $17$ attempts are labelled Perception.

The replay gives one cause to each failed run, from the last $4$ segments of the last attempt on its stop step. It
uses the joint limits of the robot, which are recorded for every family except five (IKEA furniture, rope threading,
slotted post ring, wire splicing, orbits), and labels $1{,}531$ of the $2{,}214$ failed runs. Of these, $236$ have no
physical cause to judge (Unclear, outside the shares), and $1{,}295$ remain ($417$ \priv, $878$ \std). A time-out, an
API error or a rate limit counts as an ending.

\paragraph{Endings.}
Each failed run gets one ending from its record, by the first rule that fits: (1) an API or infrastructure error, (2) a
non-zero agent exit (Budget exhausted when its message names the token limit, unexplained otherwise), (3) Budget
exhausted, (4) a task failure condition, and (5) otherwise a stop by the agent itself.

We read $343$ of the
$1{,}088$ self-stopped runs and label each Gave up, Submitted or Claimed done, or still trying when the transcript
ends inside an attempt ($47$ runs, $4\,\%$ of the $1{,}088$ self-stops). We apply the weighted shares to the $1{,}088$
runs, which gives estimates for these four endings: an estimated $80\,\%$ of the self-stops give up and $0.8\,\%$
wrongly claim success. Gemini 4 Argon has one read self-stopped run, and its self-stops stay unsplit. The split
describes each agent as one system of harness and model (\cref{sec:setup}).

The other
$232$ failed runs end on a task failure condition ($140$), an API or infrastructure error ($37$) or an unexplained exit
($55$). These runs and the $5$ unsplit self-stops of Gemini 4 Argon form Other in \cref{fig:endings}a. The runs still trying are a segment of their own.

For the $271$ read runs labelled Gave up
(weights sum to about $891$, and $871$ runs in all are estimated to give up), we reread the last entries of the
transcript and give the first reason that fits, in this order: an environment or tool error, Time running out, a belief that the task is
done, Judged infeasible, No new approach, another reason. The three named reasons cover $99\,\%$ of these runs. \cref{fig:endings}a hatches the give-ups whose reason is Time running out. Together with Budget exhausted, Time running out ends $68\,\%$ of the $1{,}413$ \std and $70\,\%$ of the $801$ \priv failed runs.

By the time left, $53\,\%$ of the $271$ read give-ups (weighted) stop with under $30$ minutes of the budget left ($38\,\%$ for Opus 5.5, $36\,\%$ for DeepSeek v4.1 Flash, $60\,\%$ for GLM 5.3 Flash) and $77\,\%$ with under $60$ minutes ($69\,\%$, $60\,\%$, $91\,\%$). Time running out is tested before Judged infeasible and No new
approach. In $31$ of the $194$ read runs with this reason ($15\,\%$ weighted), more than $60$ minutes of the
budget were left. In the $486$ labelled failed runs of five agents (weighted to about $1{,}510$), $67\,\%$
have an unfinished stuck step as their longest step (\emph{stuck step}, \cref{tab:suppl_units}), and $56\,\%$ when finishing steps are
left out.

\paragraph{Approach at the key step.}
Of the labelled runs, $913$ reach the key step and $902$ of them have an attempt at it that can be labelled. We split them into $14$ cells, by agent and by
whether the key step is done, and sample up to $8$ runs per cell, $105$ runs in all ($52$ done, $53$ not done).

We then label each sampled run blind to its outcome, from the transcript up to the code of the third attempt after the
first miss at the key step, or up to the last attempt when there are fewer. The outcome, the score and the ending are
removed. Each cell is weighted by the number of kept runs it stands for (about $2{,}690$ in all, against $2{,}693$ kept runs that reach the key step in the replay), and the intervals resample runs within each cell. Runs that plan first solve the key step in $67.5\,\%$ and the others in $67.8\,\%$ ($-0.3$ points, interval $-16$ to $+22$).

\begin{figure}[!p]
  \centering
  \includegraphics[width=\linewidth]{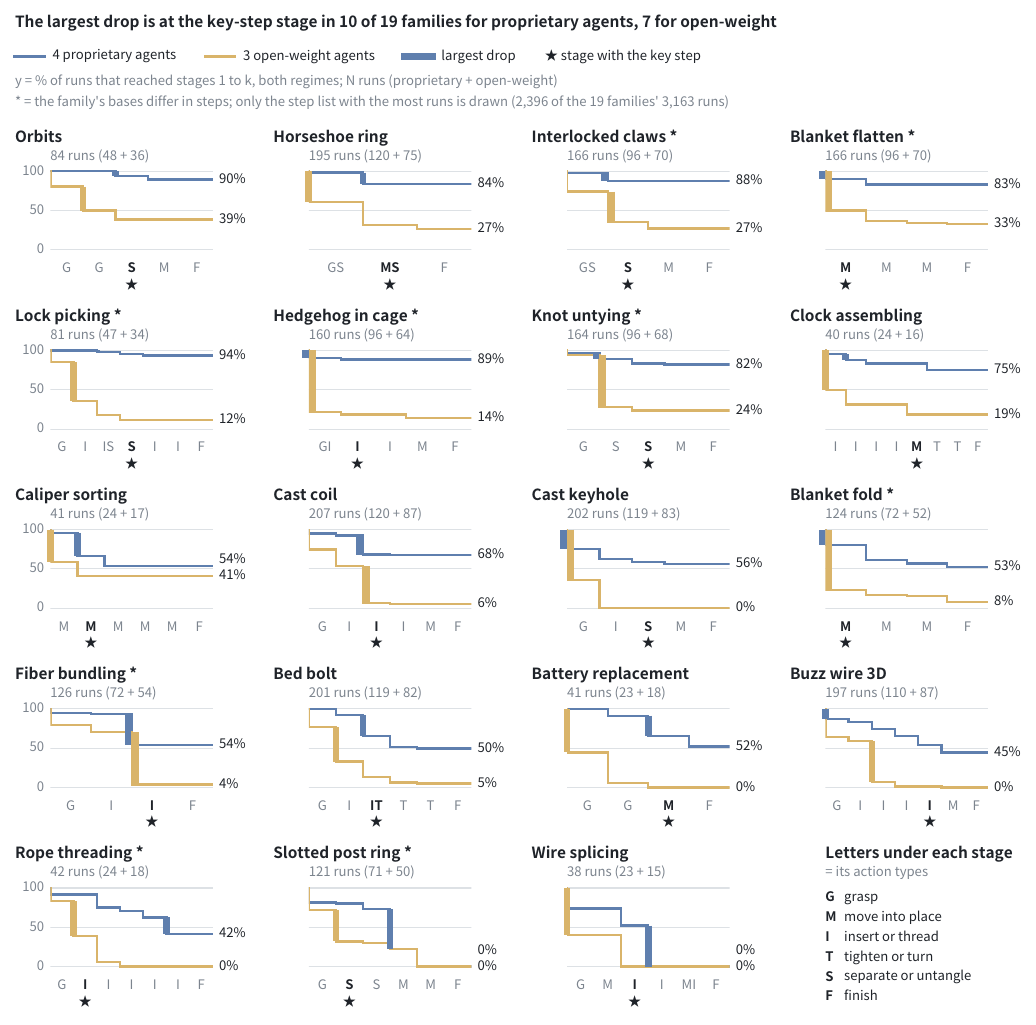}
  \caption{\textbf{Per family: where runs drop out.} One panel per puzzle family with at least $20$ runs in the drawn step list ($19$ families, $2{,}396$ runs of both regimes, solved runs included, and both counts cover the drawn step list only). Y axis: share of runs that pass stages $1$ to $k$ in a row. Each line's denominator is the runs of its group (proprietary or open-weight) in the family. The panel's $N$ is the sum of both groups. Blue: the four proprietary agents. Yellow: the three open-weight agents. The thick segment marks each line's largest drop (at the left edge: runs that never complete stage $1$), $\star$ marks the stage with the key step (\cref{tab:suppl_keysteps}), and letters give each stage's action types. The largest drop is at the key-step stage in $10$ of the $19$ families for the proprietary agents and in $7$ for the open-weight agents. $*$ marks the $9$ families whose tasks differ in steps: only the step list with the most runs is drawn, and another $767$ runs of these families are not in the figure.}
  \label{fig:suppl_ladder}
\end{figure}

\paragraph{Approach at all steps.}
\Cref{fig:endings}d uses a second labelling pass on all steps. A sample is one step of one run, one step per run,
$301$ in all and $43$ per agent: $22$ done steps and $21$ not-done steps, except that a short cell is topped up from
the other. The step has at least one real simulation attempt. The three classes (Plan first, Replan on failure, Try
directly) and their rules are those of the key step, applied to the sampled step. Each label is given blind to the
outcome, from the log cut at the code of the third attempt at the step.

Each sample counts with the run weight, scaled to the cell (agent $\times$ step done or not), and the intervals resample within each cell ($2{,}000$ draws). Plan first holds for $55.3\,\%$ of the steps ($49.7$ to $61.7\,\%$), for $47.1\,\%$ ($39.5$ to $55.2\,\%$) with the proprietary agents and $69.1\,\%$ ($60.6$ to $77.4\,\%$) with the open-weight agents. The $63$ key steps give $55.9\,\%$ and the $238$ other steps $55.2\,\%$.

Per agent the share runs from $32\,\%$ (Sol) to $83\,\%$ (Argon), each with an interval of about $\pm15$ points.
Compared within each agent and then pooled, done steps plan first $17.9$ points more often than not-done steps ($+8.1$
to $+28.1$). The log stops at the third attempt, so a step done at the first attempt has a shorter log. The key-step
result above ($-0.3$ points) counts runs, and this one counts steps.

\paragraph{Checks.}
Every one of the $3{,}618$ replays ends on the recorded hash of the final state. We keep $3{,}609$ of them. For $6$, the action file was not kept for the step analysis. In $3$, the grader records a fail although the replayed end state meets the goal, because the grader also requires a strict goal check or no safety review or forcing event. Halving or doubling the distance
and angle thresholds of the sub-goals changes the pass rate of any action type by at most $1$ point.

The two sets of labels rank the physical causes in the same order, Execution, then Planning, then Perception: per
failed attempt (transcript labels, \std $38\,\%$, $25\,\%$, $5\,\%$, and \priv $39\,\%$, $37\,\%$, $0.2\,\%$) and per
failed run (replay labels, $52\,\%$, $48\,\%$, $0\,\%$ in the \priv regime and $55\,\%$, $36\,\%$, $10\,\%$ in the \std
regime). On the same step of the same run in the \std regime, they give the same cause in $135$ of $293$ cases
($46\,\%$, $40$--$52\,\%$), and the most common split is Execution from the transcript against Planning from the
replay ($85$ cases). We compare shares within one set of labels.

On whether a step is done, the transcript labels agree with the replay labels at the last attempt on $457$ of the
$832$ steps that the replay marks done in failed runs ($55\,\%$). They mark $64$ of the $1{,}255$ steps that the
replay marks not done as done ($5\,\%$).

\subsection{Where the runs stop and the time spent there}
\label{sec:suppl_time}

\begin{figure}[!htbp]
  \centering
  \includegraphics[width=\linewidth]{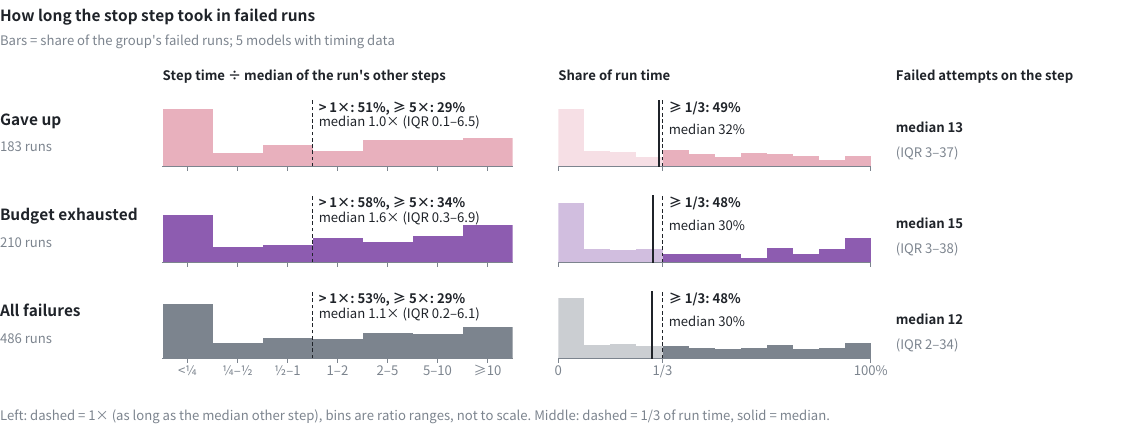}
  \caption{\textbf{Time spent on the stop step.} Per ending of the failed run: the time of the step where the run stopped, against the median time of the run's other steps (left) and against the run time (middle), and the failed attempts on that step (right).}
  \label{fig:suppl_time}
\end{figure}

\Cref{fig:suppl_time} shows how long the step took where a failed run stopped, for the five agents with timing data (Astra, GLM, Kimi, Opus, Sol). Shares are weighted to the frozen task list. Intervals are $95\,\%$ intervals.
In the $486$ failed runs, the longest single step took $59\,\%$ of the run time ($57$ to $61$), and in the $383$ solved runs it took $37\,\%$ ($35$ to $39$).
The step where a failed run stopped took a median $1.1\times$ the median time of the run's other steps (IQR $0.2$ to $6.1$, $433$ failed runs with a valid ratio). In $48\,\%$ of the $486$ failed runs it took at least a third of the run time, and the median number of failed attempts on it is $12$ (IQR $2$ to $34$).

The longest step is most often an insert-or-thread step ($24.0\,\%$ of the $486$ failed runs, $112$ runs). The other five action types take $10$ to $21\,\%$.
The longest step is a key step in $27\,\%$ of the $486$ failed runs ($26\,\%$ of the $383$ solved runs). $33\,\%$ ($29$ to $38$) of the $486$ stop steps are key steps, and so are $39\,\%$ ($33$ to $46$) of the $221$ failed runs whose stop step is a stuck step and also their longest step (\cref{tab:suppl_units}). The other $104$ of the $325$ runs whose longest step is an unfinished stuck step stop at another step. A random step is a key step in $17\,\%$.

The stuck stop steps ($233$ of the $486$, weighted shares) spread over the action types: insert or thread $28.9\,\%$ ($66$ steps), separate or untangle $23.8\,\%$ ($60$), move into place $21.5\,\%$ ($48$), grasp $15.6\,\%$ ($36$), tighten or turn $9.2\,\%$ ($21$), finish $1.0\,\%$ ($2$). No type is a majority. Insert or thread is $35.7\,\%$ of all $486$ stop steps ($164$), so it is not over-represented among the stuck ones. Separate or untangle is $23.8\,\%$ of them against $18.3\,\%$ of all stop steps, with overlapping intervals ($18.5$ to $29.1$ and $14.8$ to $22.0\,\%$). These are shares of the failed runs by action type.

\paragraph{Progress at the stop and the time on the stop step.}
Progress is the number of stages passed in a row from stage $1$, over the stages of the task (\cref{fig:suppl_ladder}). A failed run stops at a median $20\,\%$ ($2{,}214$ failed runs: $1{,}413$ \std, $801$ \priv, alike), $93.4\,\%$ stop at $50\,\%$ or less, and $881$ runs ($39.8\,\%$) pass no stage.

The shares in this paragraph and the next are raw counts over the $486$ failed runs of the five agents with step times in the labelled sample, not weighted, as in \cref{fig:endings}b. In $47.9\,\%$ of them ($43.4$ to $52.3$, $233$ runs) the stop step takes at least a third of the run time and has at least $3$ failed attempts: $53.5\,\%$ in the \std regime ($48.1$ to $59.2$, $169$ of $316$) and $37.6\,\%$ in the \priv regime ($30.0$ to $44.7$, $64$ of $170$). Among runs that pass no stage it is $56.5\,\%$ ($48.8$ to $63.5$, $96$ of $170$), and among runs that stop at $60\,\%$ or more it is $31.3\,\%$ ($15.6$ to $47.0$, $10$ of $32$). This is the stuck rule of \cref{tab:suppl_units}. Requiring also that it is the longest step gives $45.5\,\%$ ($221$ of $486$). Weighted to the kept runs, the stuck share is $47.3\,\%$ ($42.4$ to $52.0$), $52.0\,\%$ in the \std regime and $38.4\,\%$ in the \priv regime.

The ending does not change it: $49.5\,\%$ after Budget exhausted ($104$ of $210$), $48.2\,\%$ after a give-up for time ($68$ of $141$), $54.8\,\%$ after a give-up for another reason ($23$ of $42$) and $40.9\,\%$ after the other endings ($38$ of $93$). The stop step takes a median $31.6\,\%$ of the run time ($486$ failed runs), about the $34.7\,\%$ that the longest step takes in the $383$ solved runs, so the stop step is not longer than a solved run's longest step. The longest step of a failed run takes $58.3\,\%$.

\subsection{More failure cases}
\label{sec:suppl_cases}

\begin{figure}[!htbp]
  \centering
  \scriptsize
  \setlength{\tabcolsep}{1.5pt}
  \newcommand{\casefr}[1]{\includegraphics[width=0.19\linewidth]{arxiv/figures/appendix_add/cases/img/#1.jpg}}
  \begin{tabular}{@{}cccc@{}}
  \multicolumn{4}{@{}l}{\footnotesize\textbf{(a)} Argon on \texttt{cast-coil-07}, failed} \\
  \casefr{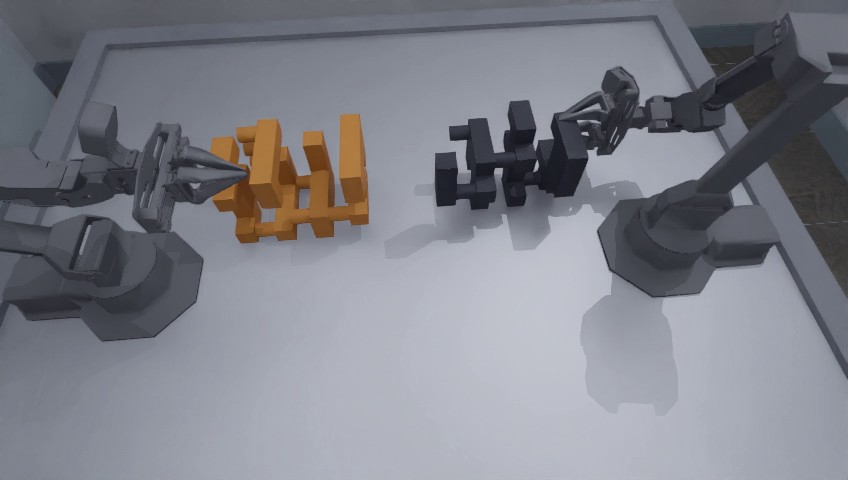} & \casefr{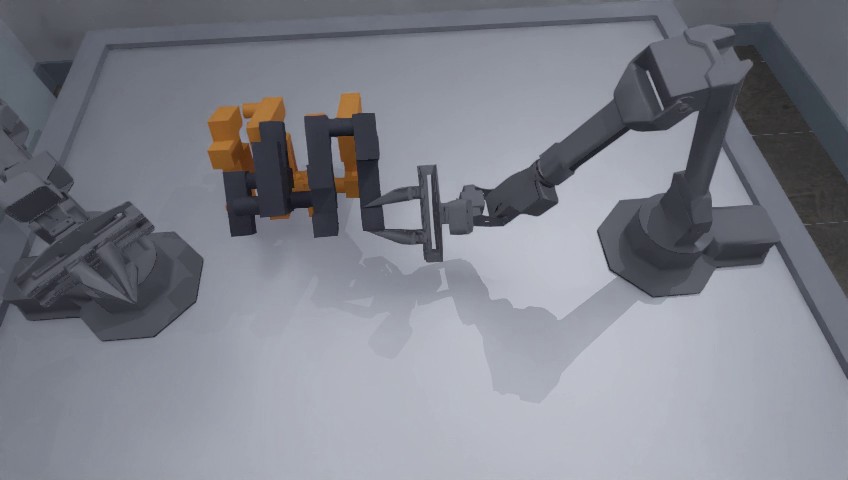} & \casefr{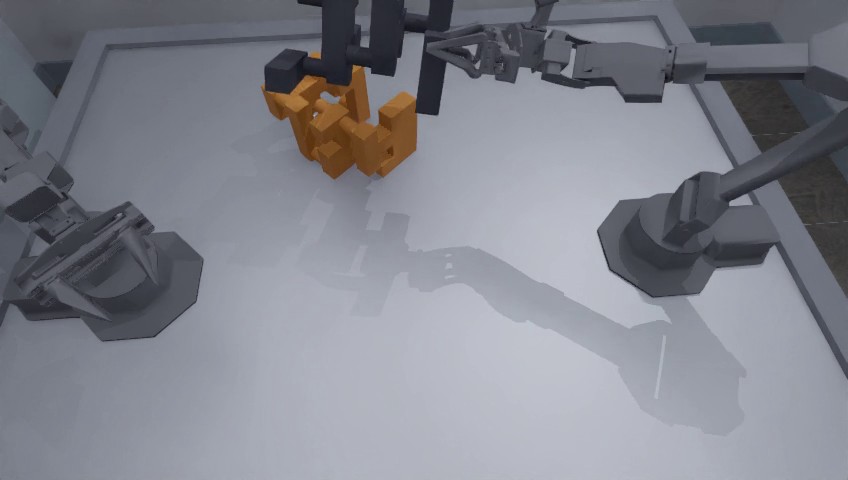} & \casefr{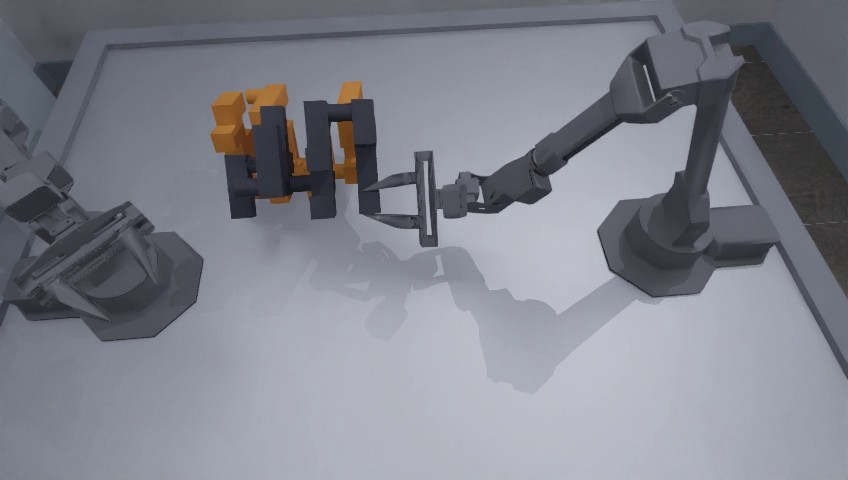} \\
  $18$\,s, grasps a piece & $34$\,s, starts the insertion & $43$\,s, the insertion backs out & $72$\,s, run ends, piece still held \\[4pt]
  \multicolumn{4}{@{}l}{\footnotesize\textbf{(b)} Sol on \texttt{ikea-svallet}, failed} \\
  \casefr{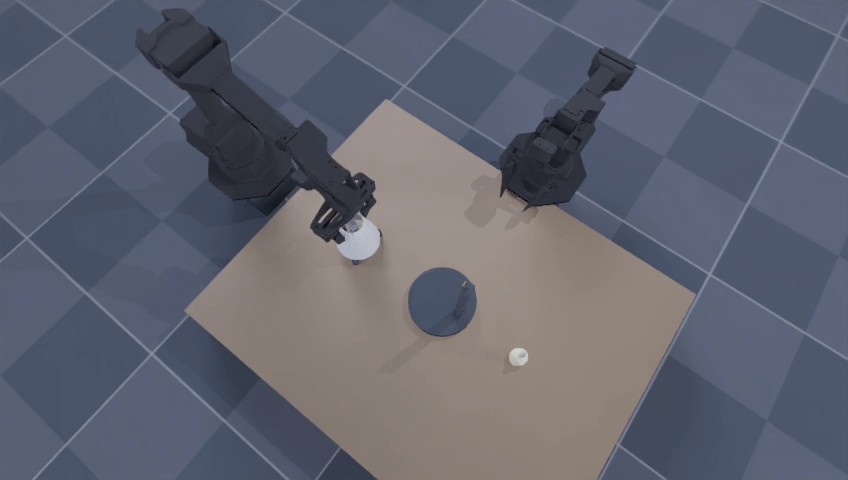} & \casefr{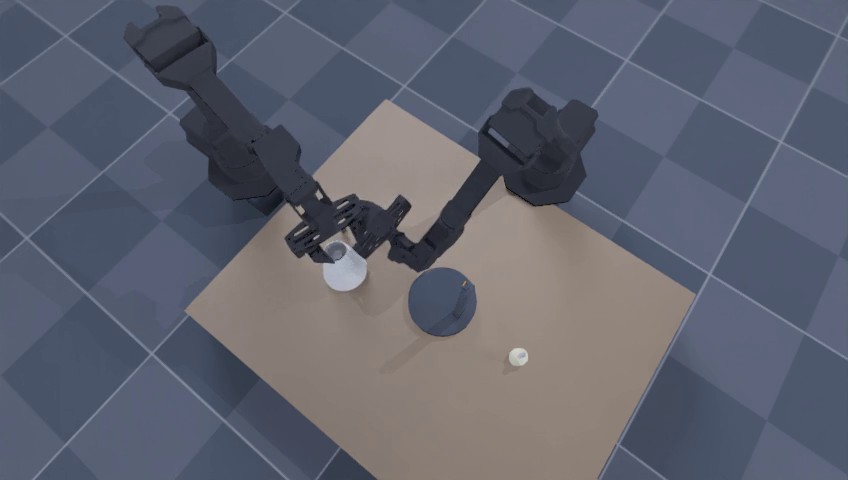} & \casefr{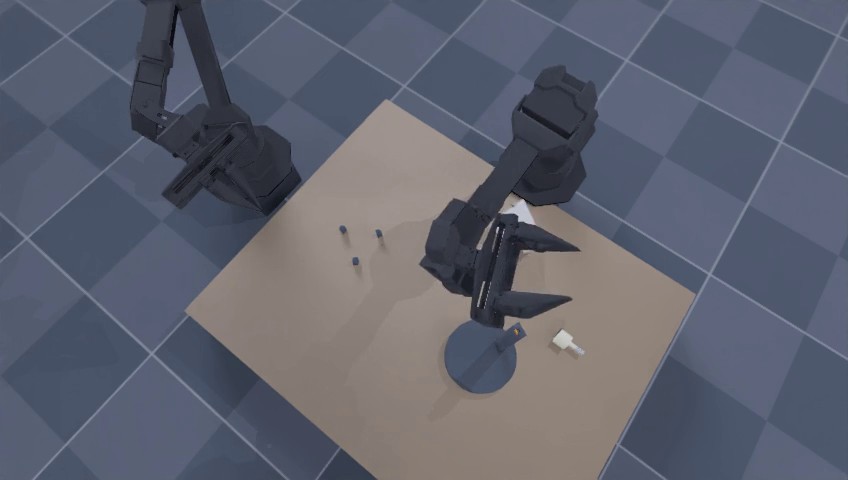} & \\
  $6.5$\,s, step 1 done, shade held & $53$\,s, shade dropped & $342$\,s, run ends, shade off the base & \\[4pt]
  \multicolumn{4}{@{}l}{\footnotesize\textbf{(c)} Argon on \texttt{slotted-post-ring-02}, solved} \\
  \casefr{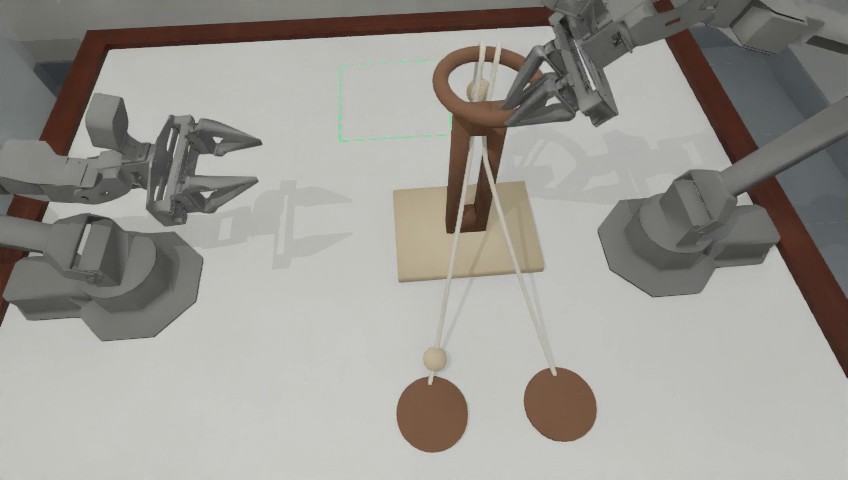} & \casefr{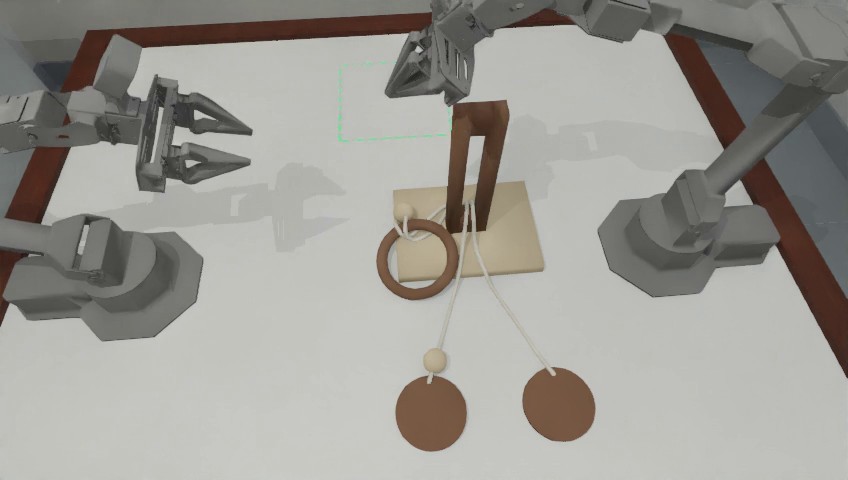} & \casefr{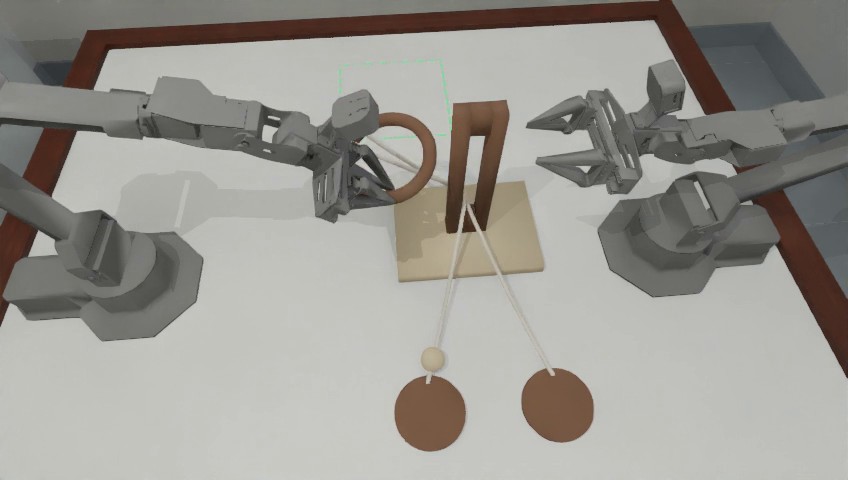} & \casefr{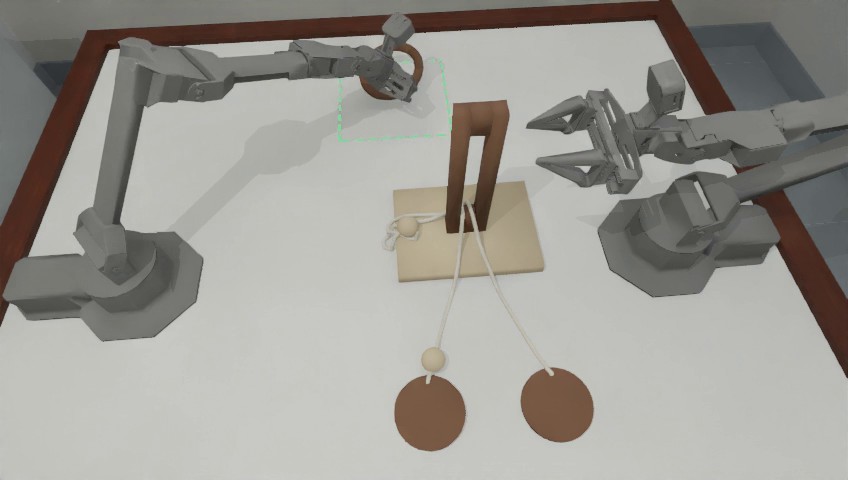} \\
  $13$\,s, first grasp lifts the ring & $18$\,s, the ring falls back & $32$\,s, a $45^\circ$ grasp holds it & $54$\,s, solved \\
  \end{tabular}
  \captionsetup{font=small}
  \caption{\textbf{Three more runs in the \std regime.} Frames from the replay of each run's recorded actions, at the simulated time below each frame, main camera only.
  (a) Argon solves the steps before the key step, then makes $12$ agent calls on the key step, fitting the two pieces together, and none reaches its sub-goal.
  (b) Sol's second step, snapping the shade onto the base, takes $89\,\%$ of the run time, and all $28$ attempts at it fail.
  (c) Argon's first grasp of the ring slips (labelled Execution), and a grasp at $45^\circ$ solves the step.}
  \label{fig:suppl_cases}
\end{figure}

\Cref{fig:suppl_cases} shows three more runs in the \std regime. Runs (a) and (b) fail and stay on one step until they end: (a) at the key step, (b) on a step that takes most of the run time. Run (c) is solved after a failed attempt.

\begingroup
\titlespacing*{\paragraph}{0pt}{1.6ex plus .5ex minus .2ex}{1em}
\subsection{The agents' own pose and extent estimates}
\label{sec:suppl_estimates}

In the \std regime an agent receives calibrated cameras with depth and no object pose (\cref{sec:suppl_modes}), so it
measures every pose itself. The depth images of this regime carry no noise. We compare the positions and boxes that the
seven coding agents write down for task parts with the simulator's values at the same moment (\cref{fig:estimates},
\cref{tab:estimates}).

\paragraph{Estimates.}
An estimate is a line of an agent's tool output or message that gives the centre or the bounding box of a task part in metres.
The line names the part, names it by its colour or material, or reports a measured point set.
We find these lines with fixed text rules and leave out lines about the robot, motion targets,
relative quantities and the goal marker.
The $44$ rigid puzzle tasks have $916$ scored \std runs of the seven agents, and we read the transcripts of $825$ of
them, drawn in random order. Of these, $314$ ($38\,\%$) state at least one estimate that we can place in time and
compare: $86$ of the $210$ solved runs and $228$ of the $615$ failed runs. The other $511$ runs have none. In $272$ of
them the only estimates are of parts that have moved, in runs that we did not replay. $91$ state no labelled vector,
$90$ have no timestamps (only $6$ of the $106$ runs of DeepSeek v4.1 Flash are usable), $26$ lie beyond the length of
the replay, $13$ hold only bare points, $10$ are clock-assembling runs, for which the rules have no vocabulary, $8$ name
a part that is not in the task and $1$ has no transcript.

\paragraph{Truth.}
The \std service reports every pose in the robot frame: the origin is the robot's root link, and the axes are those of the simulator's world.
An estimate of a fixed part, or of a movable part before any control step has run, is compared with the frozen start
state. For a later estimate we replay the recorded actions of the run up to the service call that precedes the estimate
and read the pose of every part there.
We replay $76$ runs in this way, each for at most $4{,}000$ control steps.
The error of a stated centre is its distance to the part's centre or to the part's body, whichever is smaller, so a
point that lies inside the part has no error.
The error of a stated box is the distance between its centre and the centre of the part's true box.
Where a name fits several parts, or the tool call that prints an estimate spans several service calls, the smallest
error counts.
The best estimate of a run is the one with the smallest error.

\begin{table}[!htbp]
\centering
\captionsetup{font=small}
\caption{\textbf{Position estimates per agent.} Runs read: \std runs on the $44$ rigid puzzle tasks whose transcripts were read. With an estimate: runs that state at least one position estimate of a task part that can be placed in time and compared with the simulator. Error of the best estimate: distance of the run's best estimate to the part's centre or to the part's body, whichever is smaller, so an estimate that lies inside the part has no error.}
\label{tab:estimates}
\footnotesize
\setlength{\tabcolsep}{5pt}
\begin{fittab}\begin{tabular}{@{}l r rr rr r rr@{}}
\toprule
 & & \multicolumn{2}{c}{With an estimate} & \multicolumn{2}{c}{Of these} & \multicolumn{3}{c}{Error of the best estimate} \\
\cmidrule(lr){3-4} \cmidrule(lr){5-6} \cmidrule(l){7-9}
Agent & Runs read & Runs & \% & Solved & Failed & Median (mm) & $>3$\,mm (\%) & $>10$\,mm (\%) \\
\midrule
GPT-6 Astra & $123$ & $29$ & $24$ & $16$ & $13$ & $3.7$ & $52$ & $24$ \\
GPT-6.1 Sol & $122$ & $42$ & $34$ & $21$ & $21$ & $2.1$ & $38$ & $5$ \\
Claude Opus 5.5 & $119$ & $45$ & $38$ & $17$ & $28$ & $2.9$ & $49$ & $20$ \\
Gemini 4 Argon & $121$ & $80$ & $66$ & $26$ & $54$ & $0.0$ & $24$ & $12$ \\
\addlinespace[2pt]
Kimi 3.1 Pro & $119$ & $47$ & $40$ & $3$ & $44$ & $0.0$ & $26$ & $13$ \\
DeepSeek v4.1 Flash & $106$ & $6$ & $6$ & $0$ & $6$ & $18.2$ & $67$ & $50$ \\
GLM 5.3 Flash & $115$ & $65$ & $56$ & $3$ & $62$ & $2.6$ & $45$ & $28$ \\
\midrule
All & $825$ & $314$ & $38$ & $86$ & $228$ & $1.5$ & $37$ & $18$ \\
\bottomrule
\end{tabular}\end{fittab}
\end{table}

\paragraph{Position.}
The best estimate of a run is off by a median of $1.5$\,mm.
It is off by more than $3$\,mm, the clearance of the hard interlocked-claws tasks (\cref{sec:suppl_construction}), in $37\,\%$ of the runs ($117$ of $314$), by more than $10$\,mm in $18\,\%$ and by more than $30$\,mm in $7\,\%$.
Measured to the part's centre alone, the median is $7.4$\,mm, and $77\,\%$, $42\,\%$ and $15\,\%$ of the runs exceed the three thresholds.
Solved and failed runs are alike: the best estimate is off by more than $3$\,mm in $41\,\%$ of the $86$ solved runs and in $36\,\%$ of the $228$ failed runs.
Of the runs whose best estimate is within $3$\,mm, $26\,\%$ are solved ($51$ of $197$), and of the runs beyond $3$\,mm, $30\,\%$ ($35$ of $117$, Fisher's exact test, $p=0.44$).
Eight variants of the measure, which differ in the extraction rules they admit, in whether fixed parts count and in the distance they use, give $p$ between $0.22$ and $0.81$.
At a threshold of $10$\,mm the two rates are $30\,\%$ ($77$ of $259$) and $16\,\%$ ($9$ of $55$, $p=0.046$ without correction for the number of tests), and they are $20.2\,\%$ and $20.5\,\%$ when only estimates that name the part count.
Within a run the estimates change little: in the $132$ runs with at least six estimates, the median error is $14.4$\,mm in the first third and $10.7$\,mm in the last third.

\paragraph{Extent.}
$208$ runs state the bounding box of a part.
The best box of a run differs from the part's true box by a median of $13.9$\,mm on its worst face, and by more than $10$\,mm in $60\,\%$ of these runs ($124$ of $208$).
The median is $13.8$\,mm in the $57$ solved and $13.9$\,mm in the $151$ failed runs.
A stated box follows what the cameras show of a part: where the best box differs from the true box by more than $3$\,mm in size, it is smaller than the part in $100$ runs and larger in $82$.

\paragraph{Check of the extraction.}
In a hand-read sample of $60$ scored estimates, $29$ are estimates of the matched part (median
error $4.1$\,mm), $21$ are estimates of a sub-part or of a point set that includes other bodies, and $10$ are not
estimates. Of the $22$ runs whose best estimate is more than $30$\,mm off, $2$ hold a wrong estimate when read by hand.
The transcripts hold $202$ axis estimates in $19$ runs.

\paragraph{Reading.}
The agents locate what the cameras show to within a few millimetres, in solved and in failed runs, and they know the
extent of a part only to within centimetres. The position error of an agent's estimates does not separate solved from
failed runs. This agrees with the ablation of \cref{sec:ladder}: giving GPT-6 Astra the object poses adds $4.4$ points,
giving it the part geometry adds $13.3$, and giving both adds $35.6$.

\endgroup

\section{Limitations}
\label{sec:suppl_limits}
\name runs in simulation, which gives every task a frozen scene, a physics-based grader and a replay that is identical
bit for bit, with assisted grasps of rigid parts and cloth (\cref{sec:suppl_suite}). It ships one reference solution
per task, built and verified for its own scene, so policies are evaluated as released, and its real-robot experiment
(\cref{sec:suppl_real}) covers one task as a check that the results carry over to hardware.

\end{document}